\documentclass[10pt,journal,compsoc]{IEEEtran}
\ifCLASSOPTIONcompsoc
  \usepackage[nocompress]{cite}
\else
  \usepackage{cite}
\fi
\ifCLASSINFOpdf
\else
\fi
\usepackage{graphicx}
\usepackage{float}
\usepackage{cite}
\usepackage{amsmath}
\usepackage{setspace}
\usepackage{tabularx,ragged2e}
\usepackage{color}
\usepackage{forest}
\usepackage[T1]{fontenc}
\usepackage[utf8]{inputenc}
\usepackage{tikz}
\usepackage{calc}  
\usepackage{lipsum}
\usepackage{pdflscape}
\usepackage{environ}
\usepackage{multirow}
\usepackage{pifont}
\usepackage{rotating}
\usepackage{array}
\usepackage{amssymb}
\usepackage{hyperref}
\usepackage[tableposition = top, skip=0pt]{caption}
\usepackage[export]{adjustbox}
\usepackage{changes}
\usepackage{scalerel}
\usepackage{cleveref}
\usepackage{subcaption} 

\crefformat{section}{\S#2#1#3} 
\crefformat{subsection}{\S#2#1#3}
\crefformat{subsubsection}{\S#2#1#3}

\usetikzlibrary{svg.path}
\definecolor{orcidlogocol}{HTML}{A6CE39}
\tikzset{
  orcidlogo/.pic={
    \fill[orcidlogocol] svg{M256,128c0,70.7-57.3,128-128,128C57.3,256,0,198.7,0,128C0,57.3,57.3,0,128,0C198.7,0,256,57.3,256,128z};
    \fill[white] svg{M86.3,186.2H70.9V79.1h15.4v48.4V186.2z}
                 svg{M108.9,79.1h41.6c39.6,0,57,28.3,57,53.6c0,27.5-21.5,53.6-56.8,53.6h-41.8V79.1z M124.3,172.4h24.5c34.9,0,42.9-26.5,42.9-39.7c0-21.5-13.7-39.7-43.7-39.7h-23.7V172.4z}
                 svg{M88.7,56.8c0,5.5-4.5,10.1-10.1,10.1c-5.6,0-10.1-4.6-10.1-10.1c0-5.6,4.5-10.1,10.1-10.1C84.2,46.7,88.7,51.3,88.7,56.8z};
  }
}
\newcommand\orcidicon[1]{\href{https://orcid.org/#1}{\mbox{\scalerel*{
\begin{tikzpicture}[yscale=-1,transform shape]
\pic{orcidlogo};
\end{tikzpicture}
}{|}}}}
\usepackage{hyperref}

\definechangesauthor[color=purple]{EA}

\usetikzlibrary{mindmap, trees, shadows}

\newcommand*{\info}[5][16.3]{%
  \node (#5) [annotation, #3, scale=0.65, text width = #1em,
          inner sep = 2mm ] at (#2) {%
  \list{$\bullet$}{\topsep=0pt\itemsep=0pt\parsep=0pt
    \parskip=0pt\labelwidth=8pt\leftmargin=8pt
    \itemindent=0pt\labelsep=2pt}%
    #4
  \endlist
  };
}

\definecolor{bcolor}{RGB}{50, 125, 50}

\newcommand{\vurgula}{\textit}
\newcommand{\cmark}{\ding{51}}%
\newcommand{\xmark}{\ding{55}}%

\begin{document}
%
\title{Imbalance Problems in Object Detection: A Review}
%
%
%
%

\author{Kemal~Oksuz$^\dagger$ \orcidicon{0000-0002-0066-1517}\ ,
        Baris~Can~Cam \orcidicon{0000-0001-8480-4636}\ ,
        Sinan~Kalkan$^\ddagger$ \orcidicon{0000-0003-0915-5917}\ , and
        Emre~Akbas$^\ddagger$ \orcidicon{0000-0002-3760-6722}\ 
\thanks{All authors are at the Dept. of Computer Engineering, Middle East Technical University (METU), Ankara, Turkey. E-mail:\{kemal.oksuz@metu.edu.tr, can.cam@metu.edu.tr, skalkan@metu.edu.tr, emre@ceng.metu.edu.tr\}\protect}
\thanks{$^\dagger$ Corresponding author.}
\thanks{$^\ddagger$ Equal contribution for senior authorship.}
}

%
%

\markboth{Oksuz \MakeLowercase{\textit{et al.}}: Imbalance Problems in Object Detection: A Review}%
{Oksuz \MakeLowercase{\textit{et al.}}: Imbalance Problems in Object Detection: A Review}
%



\IEEEtitleabstractindextext{%
\begin{abstract}
In this paper, we present a comprehensive review of the imbalance problems in object detection. To analyze the problems in a systematic manner, we introduce a problem-based taxonomy. Following this taxonomy, we discuss each problem in depth and present a unifying yet critical perspective on the solutions in the literature. In addition, we identify major open issues regarding the existing imbalance problems as well as imbalance problems that have not been discussed before. Moreover, in order to keep our review up to date, we provide an accompanying webpage which catalogs papers addressing imbalance problems, according to our problem-based taxonomy. Researchers can track newer studies on this webpage available at:  \url{https://github.com/kemaloksuz/ObjectDetectionImbalance}.
\end{abstract}

}

\maketitle

\IEEEdisplaynontitleabstractindextext

%
\IEEEpeerreviewmaketitle
\section{Introduction}
\label{sec:b1}
\label{sec:introduction}

Object detection is the simultaneous estimation of categories and locations of object instances in a given image. It is a fundamental problem in computer vision with many important applications in e.g.  surveillance \cite{surveillanceod2,surveillanceod1}, autonomous driving \cite{autonomousdrod2,autonomousdrod1}, medical decision making \cite{medicalod11, medicalod12}, and many problems in robotics \cite{bb8, ssd6d, rtseamless6d, bopeccv, roboticmanipulation1,bozcan2019cosmo}.

Since the time when object detection (OD) was cast as a machine learning problem, the first generation OD methods relied on hand-crafted features and linear, max-margin classifiers. The most successful and representative method in this generation was the Deformable Parts Model (DPM) \cite{DPM}. After the extremely influential work by Krizhevsky et al. in 2012 \cite{AlexNet}, deep learning (or deep neural networks) has started to dominate various problems in computer vision and OD was no exception. The current generation OD methods are all based on deep learning where both the hand-crafted features and linear classifiers of the first generation methods have been replaced by deep neural networks. This replacement has brought significant improvements in performance: On a widely used OD benchmark dataset (PASCAL VOC), while the DPM \cite{DPM} achieved 0.34 mean average-precision (mAP), current deep learning based OD models achieve around 0.80 mAP \cite{YOLObetter}.

In the last five years, although the major driving force of progress in OD has been the incorporation of deep neural networks \cite{RCNN,FastRCNN,RFCN,SSD,YOLO,FasterRCNN,FocalLoss,CornerNet}, imbalance problems in OD at several levels have also received significant attention \cite{OHEM,FgClassImbalance,FeaturePyramidNetwork,Snip,Sniper,LibraRCNN,PrimeSample}. An  \textbf{imbalance problem} with respect to an input property occurs when the distribution regarding that property affects the performance. When not addressed, an imbalance problem has adverse effects on the final detection performance. For example, the most commonly known imbalance problem in OD is the foreground-to-background imbalance which manifests itself in the extreme inequality between the number of positive examples versus the number of negatives. In a given image, while there are typically a few positive examples, one can extract millions of negative examples. If not addressed, this imbalance greatly impairs detection accuracy.  

\begin{table*}
\caption{Imbalance problems reviewed in this paper. We state that an imbalance problem with respect to an input property occurs when the distribution regarding that property affects the performance. The first column shows the major imbalance categories. For each Imbalance problem given in the middle column, the last column shows the associated input property concerning the definition of the imbalance problem.}
\centering
\renewcommand{\arraystretch}{0.6}
\begin{tabular}{|c|c|p{7cm}|}
    \hline
    \textbf{Type} & \textbf{Imbalance Problem} & \multicolumn{1}{|c|}{\textbf{Related Input Property}} \\ \hline    
    \multirow{2}{*}{Class}&Foreground-Background Class Imbalance (\cref{subsec:fgbgClassImb})& \multirow{2}{*}{ \begin{minipage}[l]{7cm} The numbers of input bounding boxes pertaining to different classes \end{minipage}} \\  \cline{2-2}
    &Foreground-Foreground  Class  Imbalance (\cref{subsec:fgClassImb})& \\
    \hline
    \multirow{2}{*}{Scale}&Object/box-level Scale Imbalance (\cref{subsec:objboxScale})& The scales of input and ground-truth bounding boxes \\  \cline{2-3}
    &Feature-level Imbalance (\cref{subsec:FPNimb})&Contribution of the feature layer from different abstraction levels of the backbone network (i.e. high and low level) \\
    \hline        
    \multirow{3}{*}{Spatial}&Imbalance in Regression Loss (\cref{subsec:Regression})& Contribution of the individual examples to the regression loss\\  \cline{2-3}
    &IoU Distribution Imbalance (\cref{subsec:IoUImb})&IoU distribution of positive input bounding boxes \\
     \cline{2-3}
    &Object Location Imbalance  (\cref{subsec:objLocImb})&Locations of the objects throughout the image\\
     \cline{2-3}     
    \hline    
    Objective&Objective Imbalance (\cref{sec:ObjImb})&Contribution of different tasks (i.e. classification, regression) to the overall loss\\  
    \hline      
\end{tabular}
\label{tab:imbDefinitions}
\end{table*}



In this paper, we review the deep-learning-era object detection literature and identify eight different imbalance problems. We group these problems in a taxonomy with four main types: class imbalance, scale imbalance, spatial imbalance and objective imbalance (Table \ref{tab:imbDefinitions}). \textbf{Class imbalance} occurs when there is significant inequality among the number of examples pertaining to different classes. While the classical example of this is the foreground-to-background imbalance, there is also imbalance among the foreground (positive) classes. \textbf{Scale imbalance} occurs when the objects have various scales and different numbers of examples pertaining to different scales. \textbf{Spatial imbalance} refers to a set of factors related to spatial properties of the bounding boxes such as regression penalty, location and IoU. Finally, \textbf{objective imbalance} occurs when there are multiple loss functions to minimize, as is often the case in OD (e.g. classification and regression losses).

\subsection{Scope and Aim}
\label{sec:scope_aim}
Imbalance problems in general have a large scope in machine learning, computer vision and pattern recognition. We limit the focus of this paper to imbalance problems in object detection. Since the current state-of-the-art is shaped by deep learning based approaches, the problems and approaches that we discuss in this paper are related to deep object detectors. Although we restrict our attention to object detection in still images, we provide brief discussions on similarities and differences of imbalance problems in other domains. We believe that these discussions would provide insights on future research directions for object detection researchers. 

Presenting a comprehensive background for object detection is not among the goals of this paper; however,  some background knowledge on object detection is required to make the most out of this paper. For a thorough background on the subject, we refer the readers to the recent, comprehensive object detection surveys \cite{ObjectDetectionIJCV, ObjectDetectionPAMI, ObjectDetectionArXiv}. We provide only a brief background on state-of-the-art object detection in Section \ref{sec:object_detection}. 

Our main aim in this paper is to present and discuss imbalance problems in object detection comprehensively.  In order to do that,
\begin{enumerate}
    \item We identify and define imbalance problems and propose a taxonomy for studying the problems and their solutions.
    \item We present a critical literature review for the existing studies with a motivation to unify them in a systematic manner. The general outline of our review includes a definition of the problems, a summary of the main approaches, an in-depth coverage of the specific solutions, and comparative summaries of the solutions. 
    \item We present and discuss open issues at the problem-level and in general.
    \item We also reserved a section for imbalance problems found in domains other than object detection. This section is generated with meticulous examination of methods  considering their adaptability to the object detection pipeline.
    \item Finally, we provide an accompanying webpage\footnote{\url{https://github.com/kemaloksuz/ObjectDetectionImbalance}} as a living repository of papers addressing imbalance problems, organized based on our problem-based taxonomy. This webpage will be continuously updated with new studies.
\end{enumerate}

\subsection{Comparison with Previous Reviews}


Recent object detection surveys \cite{ObjectDetectionIJCV, ObjectDetectionPAMI,ObjectDetectionArXiv} aim to present advances in deep learning based generic object detection. To this end, these surveys propose a taxonomy for object detection methods, and present a detailed analysis of some cornerstone methods that have had high impact. They also provide discussions on popular datasets and evaluation metrics. From the imbalance point of view, these surveys only consider the class imbalance problem with a limited provision. Additionally, Zou et al. \cite{ObjectDetectionPAMI} provide a review for methods that handle  scale imbalance. Unlike these surveys, here we focus on a classification of imbalance problems related to object detection and present a comprehensive review of methods that handle these imbalance problems. 

There are also surveys on category specific object detection (e.g. pedestrian detection, vehicle detection, face detection) \cite{vehicle_detection_pami,pedestrian_pami, face_detection_cviu, text_survey_pami}. Although Zehang Sun et al. \cite{vehicle_detection_pami} and Dollar et al. \cite{pedestrian_pami} cover the  methods proposed before the current deep learning era, they are beneficial from the imbalance point of view since they present a comprehensive analysis of feature extraction methods that handle scale imbalance. Zafeiriou et al. \cite{face_detection_cviu} and Yin et al.  \cite{text_survey_ip} propose comparative analyses of non-deep and deep methods. Litjens et al. \cite{medicalod} discuss applications of various deep neural network based methods \textit{i.e. classification, detection, segmentation} to medical image analysis. They present challenges with their possible solutions which include a limited exploration of the class imbalance problem. These category specific object detector reviews  focus on a single class and do not consider the imbalance problems in a comprehensive manner from the generic object detection perspective. 

Another set of relevant work includes the studies specifically for imbalance problems in machine learning \cite{ImbalanceSurvey3,ImbalanceSurvey2, ImbalanceBook,ImbalanceSurvey1}. These studies are limited to the foreground class imbalance problem in our context (i.e. there is no background class). Generally, they cover dataset-level methods such as undersampling and oversampling, and algorithm-level methods including feature selection, kernel modifications and weighted approaches. We identify three main differences of our work compared to such studies. Firstly, the main scope of such work is the classification problem, which is still relevant for object detection; however, object detection also has a ``search'' aspect, in addition to the recognition aspect, which brings in the background (i.e. negative) class into the picture. Secondly, except Johnson et al. \cite{ImbalanceSurvey1}, they  consider machine learning approaches in general without any special focus on deep learning based methods. Finally, and more importantly, these works only consider foreground class imbalance problem, which is only one of eight different imbalance problems that we present and discuss here (Table \ref{tab:imbDefinitions}). 


\subsection{A Guide to Reading This Review}

The paper is organized as follows. Section \ref{sec:DefNot} provides a brief background on object detection, and the list of frequently-used terms and notation used throughout the paper. Section \ref{sec:ImbalanceProblems} presents our taxonomy of imbalance problems. Sections \ref{sec:classImb}-\ref{sec:ObjImb} then cover each imbalance problem in detail, with a critical review of the proposed solutions and include  open issues for each imbalance problem. Each section dedicated to a specific imbalance problem is designed to be self-readable, containing definitions and a review of the proposed methods. In order to provide a more general perspective, in Section \ref{sec:OtherTasksImb}, we present the solutions addressing imbalance in other but closely related domains. Section \ref{sec:OpenProblems} discusses open issues that are relevant to all imbalance problems. Finally, Section \ref{sec:Conclusion} concludes the paper. 
%

Readers who are familiar with the current state-of-the-art object detection methods can directly jump to Section \ref{sec:ImbalanceProblems} and use Figure \ref{tab:imbDefinitions} to navigate both the imbalance problems and the sections dedicated to the different problems according to the taxonomy. For readers who lack a background in state-of-the-art object detection, we recommend starting with Section \ref{sec:object_detection}, and if this brief background is not sufficient, we refer the reader to the more in-depth reviews mentioned in Section \ref{sec:scope_aim}. 

\section{Background, Definitions and Notation}
\label{sec:DefNot}

In the following, we first provide a brief background on state-of-the-art object detection methods, and then present the definitions and notations used throughout the paper. 

\subsection{State of the Art in Object Detection}
\label{sec:object_detection}


Today there are two major approaches to object detection: top-down and bottom-up. Although both the top-down and bottom-up approaches were popular prior to the deep learning era, today the majority of the object detection methods follow the top-down approach; the bottom-up methods have been proposed relatively recently. The main difference between the top-down and bottom-up approaches is that, in the top-down approach, holistic object hypotheses (i.e., anchors, regions-of-interests/proposals) are generated and evaluated early in the detection pipeline, whereas in the bottom-up approach, holistic objects emerge by grouping sub-object entities like keypoints or parts, later in the processing pipeline.




Methods following the top-down approach are categorized into two: two-stage and one-stage methods. Two-stage methods \cite{RCNN,FastRCNN,RFCN,FasterRCNN} aim to decrease the large number of negative examples resulting from the predefined, dense sliding windows, called anchors, to a manageable size by using a proposal mechanism \cite{SelectiveSearch, Edgeboxes, FasterRCNN} which determines the regions where the objects most likely appear, called Region of Interests (RoIs). These RoIs are further processed by a detection network which outputs the object detection results in the form of bounding boxes and associated object-category probabilities. Finally, the non-maxima suppression (NMS) method is applied on the object detection results to eliminate duplicate or highly-overlapping results. NMS is a universal post-processing step used by all state-of-the-art object detectors. 

\begin{figure*}
\centering
\includegraphics[width=0.9\textwidth]{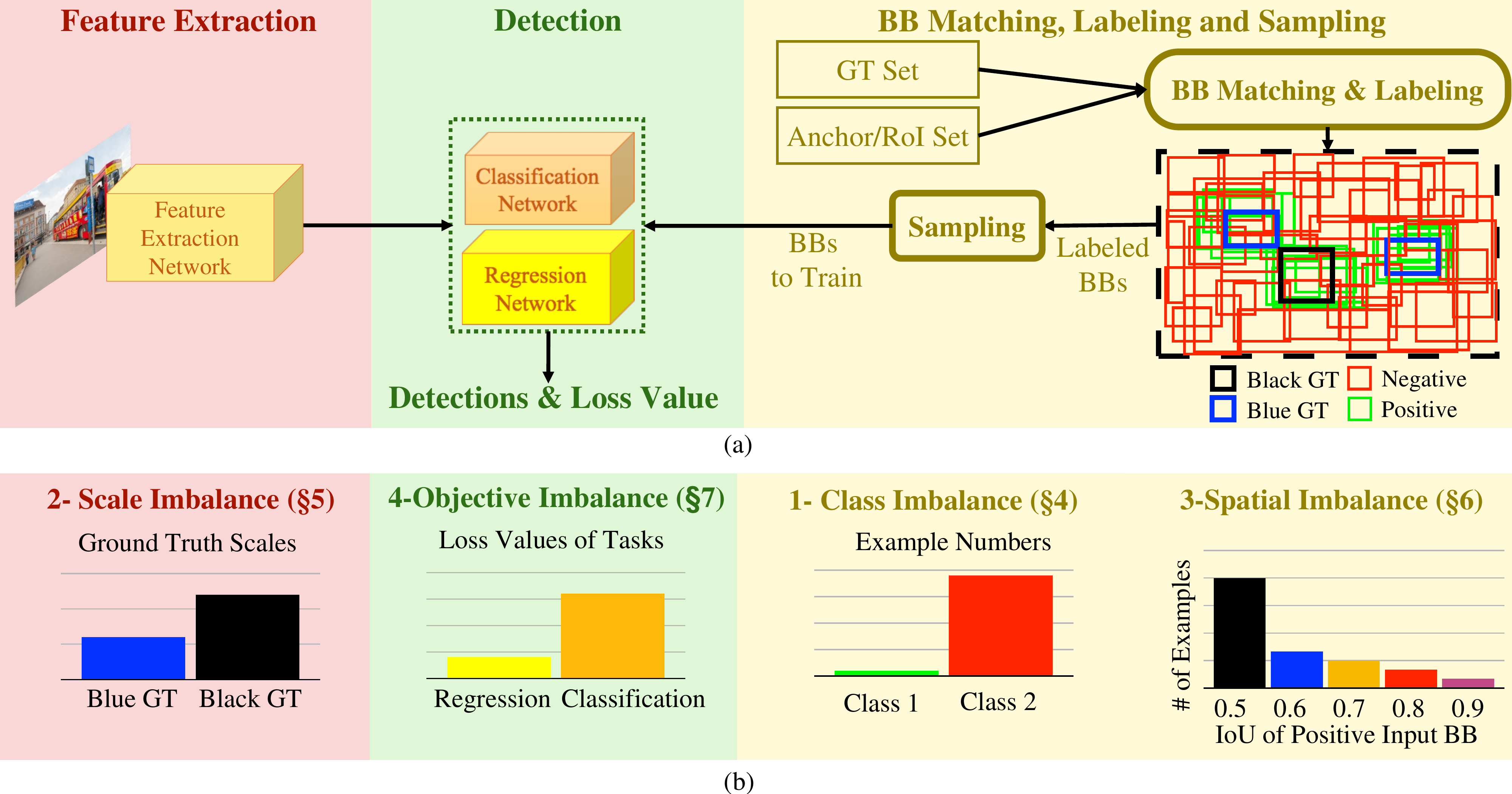}
\caption{\textbf{(a)} The common training pipeline of a generic detection network. The pipeline has 3 phases (i.e. feature extraction, detection and BB matching, labeling and sampling) represented by different background colors. \textbf{(b)} Illustration of an example imbalance problem from each category for object detection through the training pipeline. Background colors specify at which phase an imbalance problem occurs.}
\label{fig:ImbalanceProblems}
\end{figure*}

One-stage top-down methods, including SSD Variants \cite{SSD,DSSD}, YOLO variants \cite{YOLO,YOLObetter,YOLOv3} and RetinaNet \cite{FocalLoss}, are designed to predict the detection results directly from anchors -- without any proposal elimination stage -- after extracting the features from the input image. 
We present a typical one-stage object detection pipeline in Figure \ref{fig:ImbalanceProblems}(a). The pipeline starts with feeding the input image to the feature extraction network, which is usually a deep convolutional neural network. A dense set of object hypotheses (called anchors) are produced, which are then sampled and labeled by matching them to ground-truth boxes. Finally, labeled anchors (whose features are obtained from the output of the feature extraction network) are fed to the classification and regression networks for training. In a two-stage method, object proposals (or regions-of-interest) are first generated using anchors by a separate network (hence, the two stages). 

On the other hand, bottom-up object detection methods \cite{CornerNet,CenterNet,BottomUp} first predict important key-points (e.g. corners, centers, etc.) on objects and then group them  to form whole object instances by using a grouping method such as associative embedding \cite{AssocEmbed} and brute force search \cite{BottomUp}. 


\subsection{Frequently Used Terms and Notation}
\label{sec:terms-and-notation}

Table \ref{tab:notation} presents the notation used throughout the paper, and below is a list of frequently used terms.

\begin{table}
\small
\caption{Frequently used notations in the paper. \label{tab:notation}}
\renewcommand{\arraystretch}{0.6}\footnotesize
    \begin{tabular}{|>{\centering\arraybackslash}m{.15\linewidth}|>{\centering\arraybackslash}m{.2\linewidth}|>{\centering\arraybackslash}m{.5\linewidth}|}
         \hline
         \textbf{Symbol} &\textbf{Domain}&\textbf{Denotes} \\ \hline
         $B$&See.Def.&A Bounding box\\ \hline
         $C$&A set of integers&Set of Class labels in a dataset\\ \hline         
         $|C_i|$&$|C_i| \in \mathbb{Z}^+$&Number of examples for the $i$th class in a dataset\\ \hline         
         $Ci$&$i \in \mathbb{Z}^+$&Backbone feature layer at depth $i$\\ \hline         
         $\mathbb{I}(P)$&$\mathbb{I}(P) \in \{0,1\}$& Indicator function. $1$ if predicate $P$ is true, else $0$
          \\ \hline   
         $Pi$&$i \in \mathbb{Z}^+$&Pyramidal feature layer corresponding to $i$th backbone feature layer\\ \hline
         $p_i$&$p_i \in [0,1]$&Confidence score of $i$th class (i.e. output of classifier)\\ \hline
         $p_s$&$p_s \in [0,1]$&Confidence score of the ground truth class\\ \hline
         $p_0$&$p_0 \in [0,1]$&Confidence score of background class\\ \hline     
         $u$&$u \in C$&Class label of a ground truth\\ \hline
         $\hat{x}$&$\hat{x} \in \mathbb{R}$&Input of the regression loss\\ \hline
    \end{tabular}
\end{table}
\normalsize
\noindent \textbf{Feature Extraction Network/Backbone:} This is the part of the object detection pipeline from the input image until the detection network. 

\noindent \textbf{Classification Network/Classifier:} This is the part of the object detection pipeline from the features extracted by the backbone to the classification result, which is indicated by a confidence score.

\noindent \textbf{Regression Network/Regressor:} This is the part of the object detection pipeline from the features extracted by the backbone to the regression output, which is indicated by two bounding box coordinates each of which consisting of an x-axis and y-axis values.

\noindent \textbf{Detection Network/Detector:} It is the part of the object detection pipeline including both classifier and regressor.

\noindent \textbf{Region Proposal Network (RPN):} It is the part of the two stage object detection pipeline from the features extracted by the backbone to the generated proposals, which also have confidence scores and bounding box coordinates.

\noindent \textbf{Bounding Box:} A rectangle on the image limiting certain features. Formally, $[x_1,y_1,x_2,y_2]$ determine a bounding box with top-left corner $(x_1,y_1)$ and bottom-right corner $(x_2,y_2)$ satisfying $x_2>x_1$ and $y_2>y_1$. 

\noindent \textbf{Anchor:} The set of pre defined bounding boxes on which the RPN in two stage object detectors and detection network in one stage detectors are applied.

\noindent \textbf{Region of Interest (RoI)/Proposal:} The set of bounding boxes generated by a proposal mechanism such as RPN on which the detection network is applied on two state object detectors.

\noindent \textbf{Input Bounding Box:} Sampled anchor or RoI the detection network or RPN is trained with.

\noindent \textbf{Ground Truth:} It is tuple $(B,u)$ such that $B$ is the bounding box and $u$ is the \textbf{class label} where $u \in C$ and $C$ is the enumeration of the classes in the dataset.

\noindent \textbf{Detection:} It is a tuple $(\Bar{B},p)$ such that $\Bar{B}$ is the bounding box and $p$ is the vector over the confidence scores for each class\footnote{We use class and category interchangeably in this paper.} and bounding box. 

\noindent \textbf{Intersection Over Union:} For a ground truth box $B$ and a detection box $\Bar{B}$, we can formally define Intersection over Union(IoU) \cite{PASCAL,ImageNet}, denoted by $\mathrm{IoU}(B,\Bar{B})$, as
\begin{align}
\label{eq:IoU}
\mathrm{IoU}(B,\Bar{B})&=\frac{A(B \cap \Bar{B})}{A(B \cup \Bar{B})},
\end{align}
 such that $A(B)$ is the area of a bounding box $B$.
 
\noindent \textbf{Under-represented Class:} The class which has less samples in a dataset or mini batch during training in the context of class imbalance.

\noindent \textbf{Over-represented Class:} The class which has more samples in a dataset or mini batch during training in the context of class imbalance.

\noindent \textbf{Backbone Features:} The set of features obtained during the application of the backbone network. 

\noindent \textbf{Pyramidal Features/Feature Pyramid:} The set of features obtained by applying some transformations to the backbone features. 



\noindent \textbf{Regression Objective Input:} Some methods make predictions in the log domain by applying some transformation which can also differ from method to method (compare transformation in Fast R-CNN \cite{FastRCNN} and in KL loss \cite{KLLoss} for Smooth L1 Loss), while some methods directly predict the bounding box coordinates \cite{CornerNet}. For the sake of clarity, we use $\hat{x}$ to denote the regression loss input for any method.


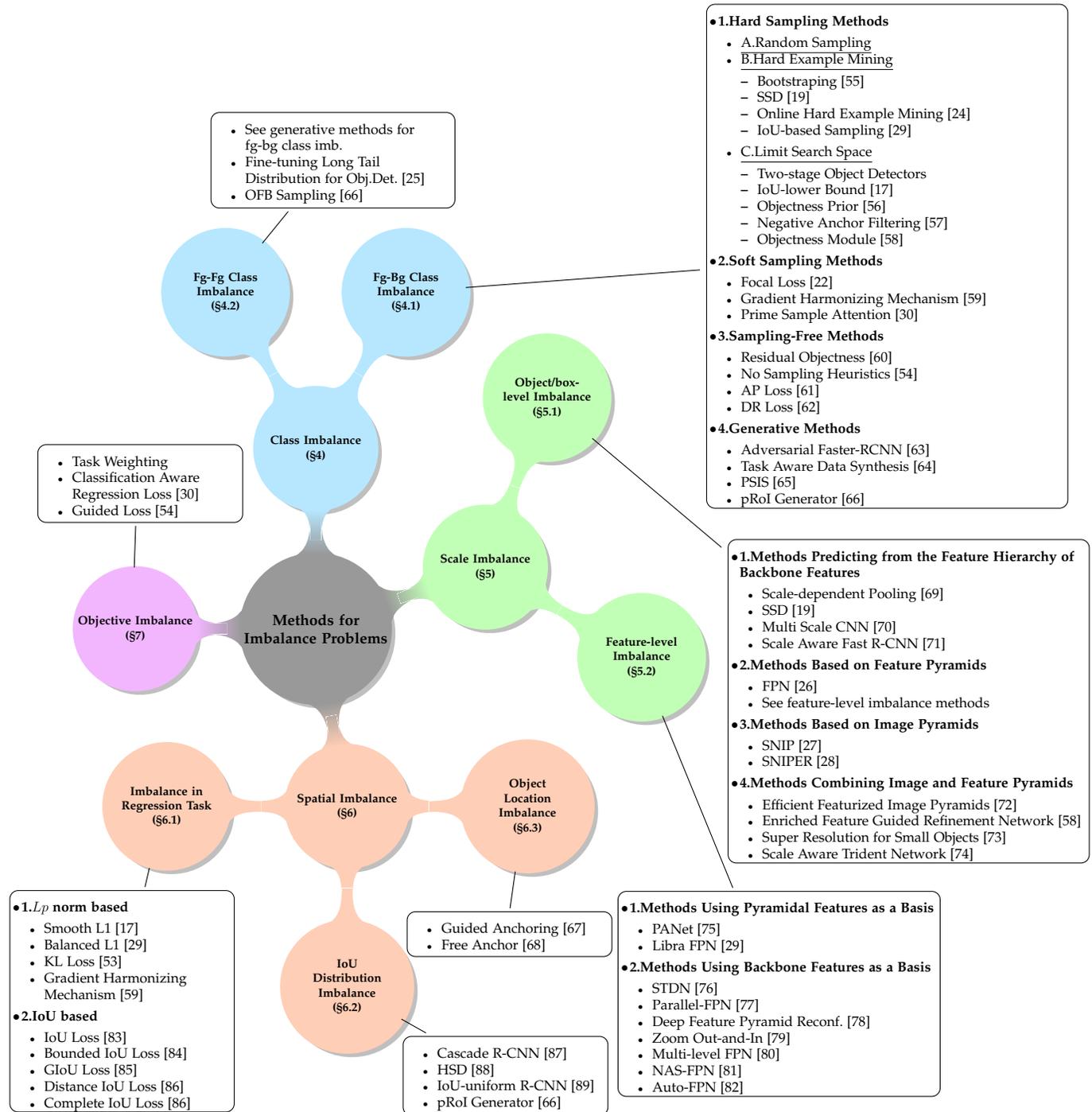
\begin{figure*}
\centering
\hypersetup{hidelinks}
\definecolor{violet}{RGB}{242,182,255}
\definecolor{bleuaefe}{RGB}{182, 231, 255}
\definecolor{matorange}{RGB}{255,206,182}
\definecolor{lightgreen}{RGB}{194,255,182}
\begin{tikzpicture}[every annotation/.style = {draw, fill = white, font = \LARGE, scale = 0.6}, scale = 0.6]
  \path[mindmap,concept color=black!40,text=black,
    every node/.style={concept, drop shadow, scale=0.6},
    root/.style    = {concept color=black!40, font=\large\bfseries,text width=12em},
    level 1 concept/.append style={font=\bfseries, sibling angle=90,text width=10em, level distance=15em,inner sep=0pt},
    level 2 concept/.append style={font=\bfseries,level distance=15em, text width=10em},
    level 3 concept/.append style={font=\bfseries,level distance=15em, text width=9em},
    level 4 concept/.append style={font=\bfseries,level distance=15em, text width=9em},
    level 5 concept/.append style={font=\bfseries,level distance=15em, text width=9em},
  ]
  node[root] {Methods for Imbalance Problems} [clockwise from=0]
    child[concept color=matorange, grow=280]{ node(BB Imbalance) {Spatial Imbalance\\(\cref{sec:spatialImb})} [clockwise from=90]
        child[grow=180] {
          node (regressionImbalance) {Imbalance in Regression Task \\ (\cref{subsec:Regression})} [clockwise from=90]
        }    
        child[grow=270] {
          node (iouImbalance) {IoU \\Distribution\\Imbalance \\ (\cref{subsec:IoUImb})} [clockwise from=90]
        }
        child[grow = 360] {
          node (featureImbalance) [concept] {{Object \\ Location \\ Imbalance \\ (\cref{subsec:objLocImb})}}
            [clockwise from=30]
        }
    }
    child[concept color=lightgreen, grow=20] {
      node[concept] (scaleImbalance) {Scale Imbalance\\(\cref{sec:Scaleimb})}
        [clockwise from=310]
      child [grow=70] { node (objboxImbalance)[concept] {Object/box-level Imbalance  \\ (\cref{subsec:objboxScale})}}
      child [grow=330] { node (fpnfeatureImbalance)[concept] {Feature-level Imbalance \\ (\cref{subsec:FeatImb})}}
    }
    child[concept color=violet] { node (multiTask) [concept] {Objective Imbalance \\ (\cref{sec:ObjImb})} [clockwise from=270]
    }
    child[concept color=bleuaefe] { node[concept] {Class Imbalance \\(\cref{sec:classImb})} [clockwise from=139]
        child[grow=60] { node[concept] (fgbgclassImbalance) {Fg-Bg Class Imbalance \\ (\cref{subsec:fgbgClassImb})}
        }
        child[grow=120]  { node[concept] (datasetLevel) {Fg-Fg Class Imbalance \\ (\cref{subsec:fgClassImb})}
        }
    };
    
    
   \info[25]{multiTask.north east}{anchor=north west, xshift = -12em, yshift= 12em}{%
       \item[] \begin{itemize}
             \item Task Weighting
             \item Classification Aware Regression Loss \cite{PrimeSample}
             \item Guided Loss \cite{SamplingHeuristics}
       \end{itemize}
     }{multitaskbox}
    

     \info[45]{fgbgclassImbalance.north}{anchor=north, xshift = 40em, yshift=19.0em}{%
         \item \textbf{1.Hard Sampling Methods}
            \begin{itemize}
                \item \underline{A.Random Sampling}
                \item \underline{B.Hard Example Mining}
                \begin{itemize}
                    \item Bootstraping \cite{BootstrappingFace1}
                    \item SSD \cite{SSD}
                    \item Online Hard Example Mining \cite{OHEM}
                    \item IoU-based Sampling \cite{LibraRCNN}
                \end{itemize}
                \item \underline{C.Limit Search Space}
                \begin{itemize}
                    \item Two-stage Object Detectors
                    \item IoU-lower Bound \cite{FastRCNN}
                    \item Objectness Prior \cite{RON}
                    \item Negative Anchor Filtering \cite{AnchorRefine}
                    \item Objectness Module \cite{EnrichedFeatureGuided}
                \end{itemize}
          \end{itemize}
          \item \textbf{2.Soft Sampling Methods}
                \begin{itemize}
                    \item Focal Loss \cite{FocalLoss}
                    \item Gradient Harmonizing Mechanism \cite{gradientharmonizing}
                    \item Prime Sample Attention \cite{PrimeSample}
                \end{itemize}
            \item \textbf{3.Sampling-Free Methods}
              \begin{itemize}
                    \item Residual Objectness \cite{ResidualObjectness}
                    \item No Sampling Heuristics \cite{SamplingHeuristics}
                    \item AP Loss \cite{APLoss}
                    \item DR Loss \cite{DRLoss}
              \end{itemize}                
            \item \textbf{4.Generative Methods}
              \begin{itemize}
                  \item Adversarial Faster-RCNN \cite{AFastRCNN}
                  \item Task Aware Data Synthesis \cite{GenerateSynthetic}
                  \item PSIS \cite{PSIS}
                  \item pRoI Generator \cite{BBSampling}
              \end{itemize}
    }{fgbgclassimbalancebox}
    
     \info[30]{datasetLevel.north}{anchor=north west,xshift = -1.5em, yshift = 10.0em}{%
       \item[] \begin{itemize}
           \item See generative methods for fg-bg class imb.
           \item Fine-tuning Long Tail Distribution for Obj.Det. \cite{FgClassImbalance}
           \item OFB Sampling \cite{BBSampling}
       \end{itemize}
     }{datasetlevelbox}
    
    \info[25]{featureImbalance.south}{anchor=north,xshift = -1.5em, yshift = -3em}{%
    \item[]
    \begin{itemize}
        \item Guided Anchoring \cite{GuidedAnchoring}
        \item Free Anchor \cite{FreeAnchor}
    \end{itemize}{}
    }{featureimbalancebox}
    
     \info[45]{objboxImbalance.west}{anchor=south west, xshift = 20em, yshift=-39.0em}{%
     \item \textbf{1.Methods Predicting from the Feature Hierarchy of Backbone Features}
      \begin{itemize}
        \item Scale-dependent Pooling \cite{ScaleDpdPool}
        \item SSD \cite{SSD}
        \item Multi Scale CNN \cite{MSCNN}
        \item Scale Aware Fast R-CNN  \cite{ScaleAwarefastRCNN}      
        \end{itemize}
        \item \textbf{2.Methods Based on Feature Pyramids}   
        \begin{itemize}
        \item FPN \cite{FeaturePyramidNetwork}
        \item See feature-level imbalance methods
        \end{itemize}
        \item \textbf{3.Methods Based on Image Pyramids}     
        \begin{itemize}
        \item SNIP  \cite{Snip}
        \item SNIPER \cite{Sniper}
        \end{itemize}
        \item \textbf{4.Methods Combining Image and Feature Pyramids}
        \begin{itemize}
        \item Efficient Featurized Image Pyramids \cite{EffFeaturizedImgPyramid}
        \item Enriched Feature Guided Refinement Network \cite{EnrichedFeatureGuided}
        \item Super Resolution for Small Objects \cite{BettertoFollow}
        \item Scale Aware Trident Network  \cite{TridentNet}
      \end{itemize}
    }{scaleimbalancebox}
    
    \info[40]{fpnfeatureImbalance.east}{anchor=west, xshift = -7.5em, yshift= -28em}{%
      \item \textbf{1.Methods Using Pyramidal Features as a Basis}
      \begin{itemize}
        \item PANet \cite{PANet}
        \item Libra FPN \cite{LibraRCNN}
      \end{itemize}
    \item \textbf{2.Methods Using Backbone Features as a Basis}      
      \begin{itemize}
        \item STDN \cite{ScaleFPN}
        \item Parallel-FPN \cite{ParallelFPN}
        \item Deep Feature Pyramid Reconf. \cite{PyramidReconfiguration}
        \item Zoom Out-and-In \cite{ZoomOutIn}
        \item Multi-level FPN \cite{MultiLevelFPN}
        \item NAS-FPN \cite{NASFPN}
        \item Auto-FPN \cite{AutoFPN}    
      \end{itemize}
    }{fpnfeatureimbalancebox}
    
    
      \info[27.5]{regressionImbalance.south west}{anchor=north,xshift = 0.0em, yshift= -3em}{%
       \item \textbf{1.$Lp$ norm based}
       \begin{itemize}
                 \item Smooth L1 \cite{FastRCNN}
                 \item Balanced L1 \cite{LibraRCNN}
                 \item KL Loss \cite{KLLoss}
                 \item Gradient Harmonizing Mechanism \cite{gradientharmonizing}
       \end{itemize}      
      \item \textbf{2.IoU based}
       \begin{itemize}
                 \item IoU Loss \cite{UnitBox}
                 \item Bounded IoU Loss \cite{BoundedIoU}
                 \item GIoU Loss \cite{GIoULoss}
                 \item Distance IoU Loss \cite{DIoULoss}
                 \item Complete IoU Loss \cite{DIoULoss}
       \end{itemize}
     }{regressiontaskbox}

    \info[25]{iouImbalance.east}{anchor=south east, xshift = 16.5em, yshift=-11.0em}{%
        \item[] \begin{itemize}
        \item Cascade R-CNN \cite{CascadeRCNN}
        \item HSD \cite{HierarchicalShotDet}
        \item IoU-uniform R-CNN \cite{IoUUniform}
        \item pRoI Generator \cite{BBSampling}
        \end{itemize}
    }{iouimbalancebox}
    

    
    \draw[transform canvas={xshift=-1.5pt}] (fgbgclassImbalance) -- (fgbgclassimbalancebox);
    \draw[transform canvas={xshift=-1.5pt}] (datasetLevel) -- (datasetlevelbox);
    \draw[transform canvas={xshift=-1.5pt}] (featureImbalance) -- (featureimbalancebox);
    \draw[transform canvas={xshift=-1.5pt}] (fpnfeatureImbalance) -- (fpnfeatureimbalancebox);
    \draw[transform canvas={xshift=-1.5pt}] (objboxImbalance) -- (scaleimbalancebox);
    \draw[transform canvas={xshift=-1.5pt}] (regressionImbalance) -- (regressiontaskbox);
    \draw[transform canvas={xshift=-1.5pt}] (iouImbalance) -- (iouimbalancebox);
    \draw[transform canvas={xshift= -1.5pt}] (multiTask) -- (multitaskbox);
    
    
\end{tikzpicture}
\caption{Problem based categorization of the methods used for imbalance problems. Note that a work may appear at multiple locations if it addresses multiple imbalance problems -- e.g. Libra R-CNN \cite{LibraRCNN}}
\label{fig:ProblemBasedCat}
\end{figure*}

\section{A Taxonomy of the Imbalance Problems and their Solutions in Object Detection}
\label{sec:ImbalanceProblems}
\label{sec:Taxonomy}

In Section \ref{sec:introduction}, we defined the problem of imbalance  as the occurrence of a distributional bias regarding an input property in the object detection training pipeline. 
Several different types of such imbalance can be observed at various stages of the common object detection pipeline  (Figure \ref{fig:ImbalanceProblems}). To study these problems in a systematic manner, we propose a taxonomy based on the related input property.

We identify eight different imbalance problems, which we group into four main categories: class imbalance, scale imbalance, spatial imbalance and objective imbalance. Table \ref{tab:imbDefinitions}  presents the complete taxonomy along with a brief definition for each problem. In Figure \ref{fig:ProblemBasedCat}, we present the same taxonomy along with a list of proposed solutions for each problem. Finally, in Figure \ref{fig:ImbalanceProblems}, we illustrate a generic object detection pipeline where each phase is  annotated with their typically observed  imbalance problems. In the following, we elaborate on the brief definitions provided earlier, and illustrate the typical phases where each imbalance problem occurs.

\textbf{Class imbalance} (Section \ref{sec:classImb}; blue branch in Figure \ref{fig:ProblemBasedCat}) occurs when a certain class is over-represented. This problem can manifest itself as either ``foreground-background imbalance,'' where the background instances significantly outnumber the positive ones; or  ``foreground-foreground imbalance,''  where  typically a small fraction of classes dominate the dataset (as observed from the long-tail distribution in Figure \ref{fig:DatasetStats}). The class imbalance is usually handled at the ``sampling'' stage in the object detection pipeline (Figure \ref{fig:ImbalanceProblems}). 

\textbf{Scale imbalance} (Section \ref{sec:Scaleimb}; green branch in Figure \ref{fig:ProblemBasedCat}) is observed when object instances have various scales and different number of examples pertaining to different scales. This problem is a natural outcome of the fact that objects have different dimensions in  nature. Scale also could cause imbalance at the feature-level (typically handled in the ``feature extraction'' phase in Figure \ref{fig:ImbalanceProblems}), where contribution from different abstraction layers (i.e. high and low levels) are not balanced. Scale imbalance problem suggests that a single scale of visual processing is not sufficient for detecting objects at different scales. However, as we will see in Section \ref{sec:Scaleimb}, proposed methods fall short in addressing scale imbalance, especially for small objects, even when small objects are surprisingly abundant in a dataset.

\textbf{Spatial imbalance} (Section \ref{sec:spatialImb}; orange branch in Figure \ref{fig:ProblemBasedCat}) refers to a set of factors related to spatial properties of the bounding boxes. Owing to these spatial properties, we identify three sub-types of spatial imbalance: (i) ``imbalance in regression loss'' is about the contributions of individual examples to the regression loss and naturally the problem is related to loss function design (ii) ``IoU distribution imbalance'' is related to the  biases  in the distribution of IoUs (among ground-truth boxes vs. anchors or detected boxes), and typically observed at the bounding-box matching and labeling phase in the object detection pipeline (Figure \ref{fig:ImbalanceProblems}) (iii) ``object location imbalance'' is about the  location distribution of object instances in an image, which is related to both the design of the anchors and the sampled subset to train the detection network.

Finally, \textbf{objective imbalance} (Section \ref{sec:ObjImb}; purple branch in Figure \ref{fig:ProblemBasedCat})  occurs when there are multiple objectives (loss functions) to minimize (each for a specific task, e.g. classification and box-regression). Since different objectives might be incompatible in terms of their ranges as well as their optimum solutions, it is  essential to develop a balanced strategy that can find a solution acceptable in terms of all objectives. 

\begin{figure}
\centering
\includegraphics[width=0.42\textwidth]{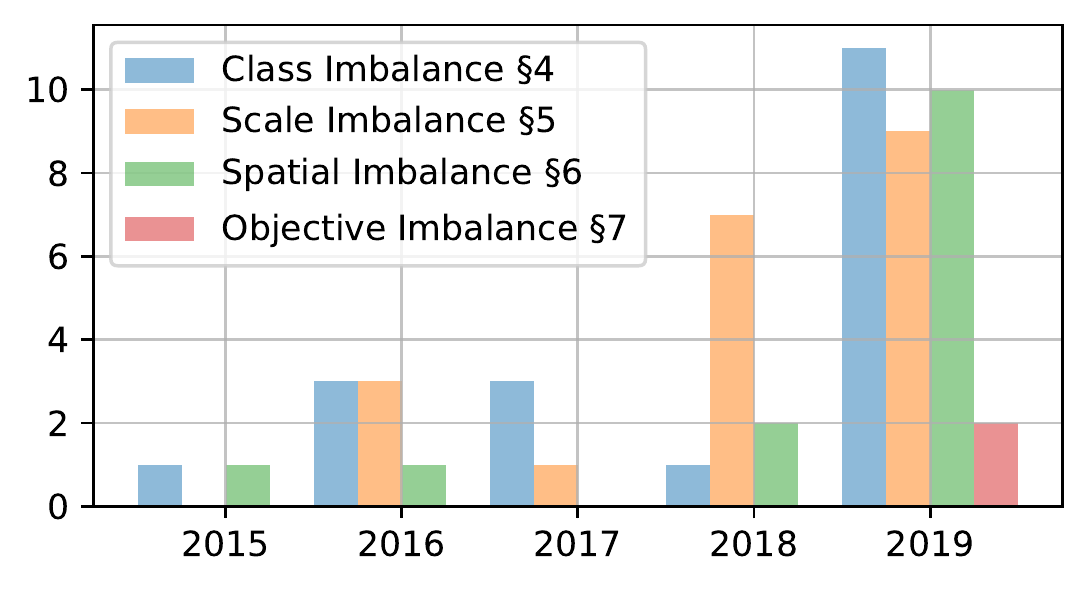}
\caption{Number of papers per imbalance problem category through years.}
\label{fig:AnnualStats}
\end{figure}

Figure \ref{fig:ProblemBasedCat} gives an overall picture of the attention that  different types of imbalance problems have received from the research community. For example, while there are numerous methods devised for the foreground-background class imbalance problem, the objective imbalance and object location imbalance problems are examples of the problems that received relatively little attention. However, recently there have been a rapidly increasing interest (Figure \ref{fig:AnnualStats}) in these imbalance problems as well, which necessitates a structured view and perspective on the problems as well as the solutions, as proposed in this paper.

Note that some imbalance problems are directly owing to the data  whereas some are the by-products of the specific methods used. For example, class imbalance, object location imbalance etc. are the natural outcomes of the distribution of classes in the real world. On the other hand, objective imbalance,  feature-level imbalance and imbalance in regression loss are owing to the selected methods and possibly avoidable with a different set of methods; for example, it is possible to avoid IoU distribution imbalance altogether by following a bottom-up approach where, typically, IoU is not a labeling criterion  (e.g. \cite{CornerNet,BottomUp}).  

\section{Imbalance 1: Class Imbalance}
\label{sec:classImb}
\begin{figure*}
        \captionsetup[subfigure]{}
        \centering
        \begin{subfigure}[b]{0.33\textwidth}
                \includegraphics[width=\textwidth]{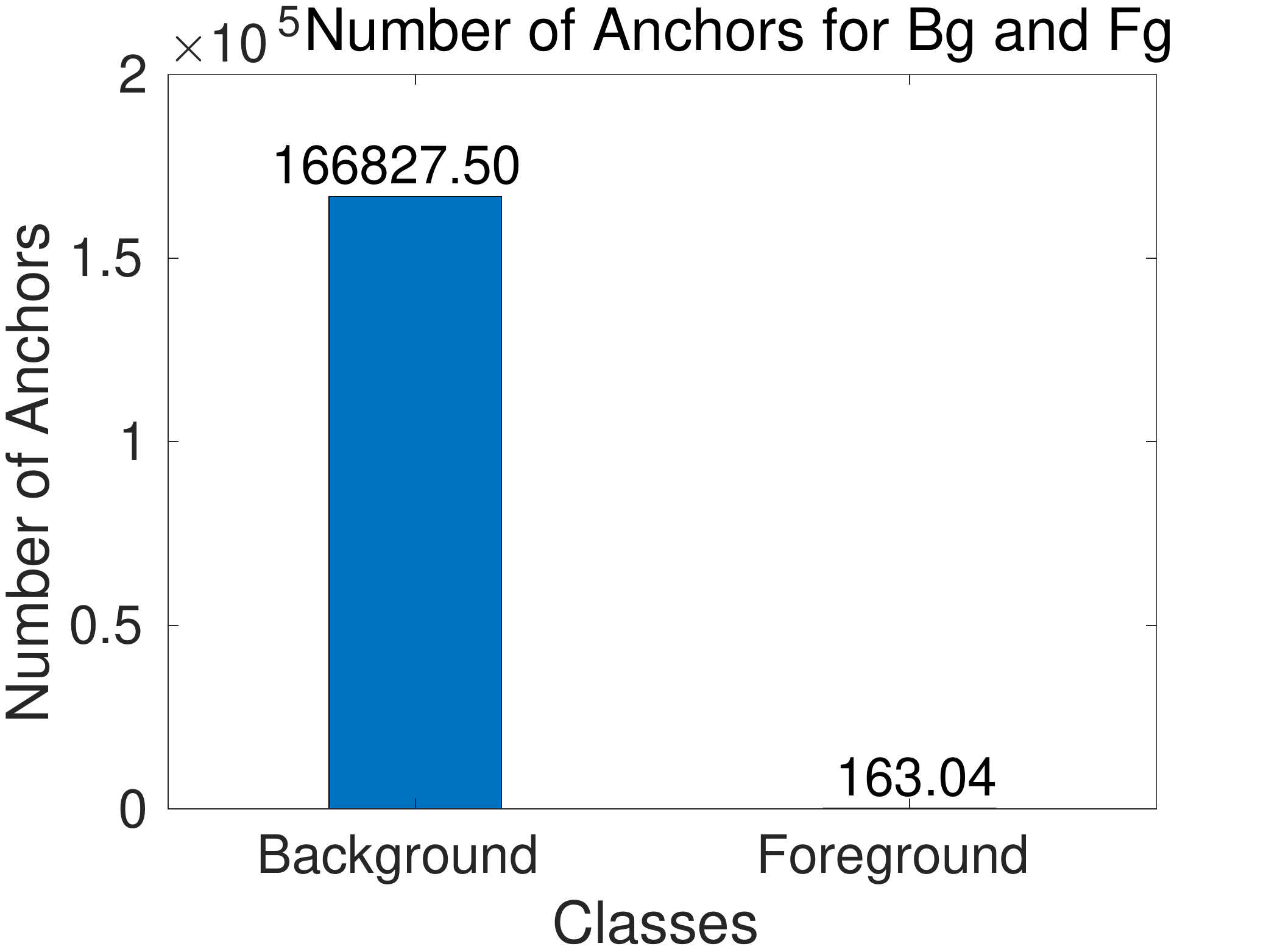}
                \caption{}
        \end{subfigure}       
        \begin{subfigure}[b]{0.55\textwidth}
                \includegraphics[width=\textwidth]{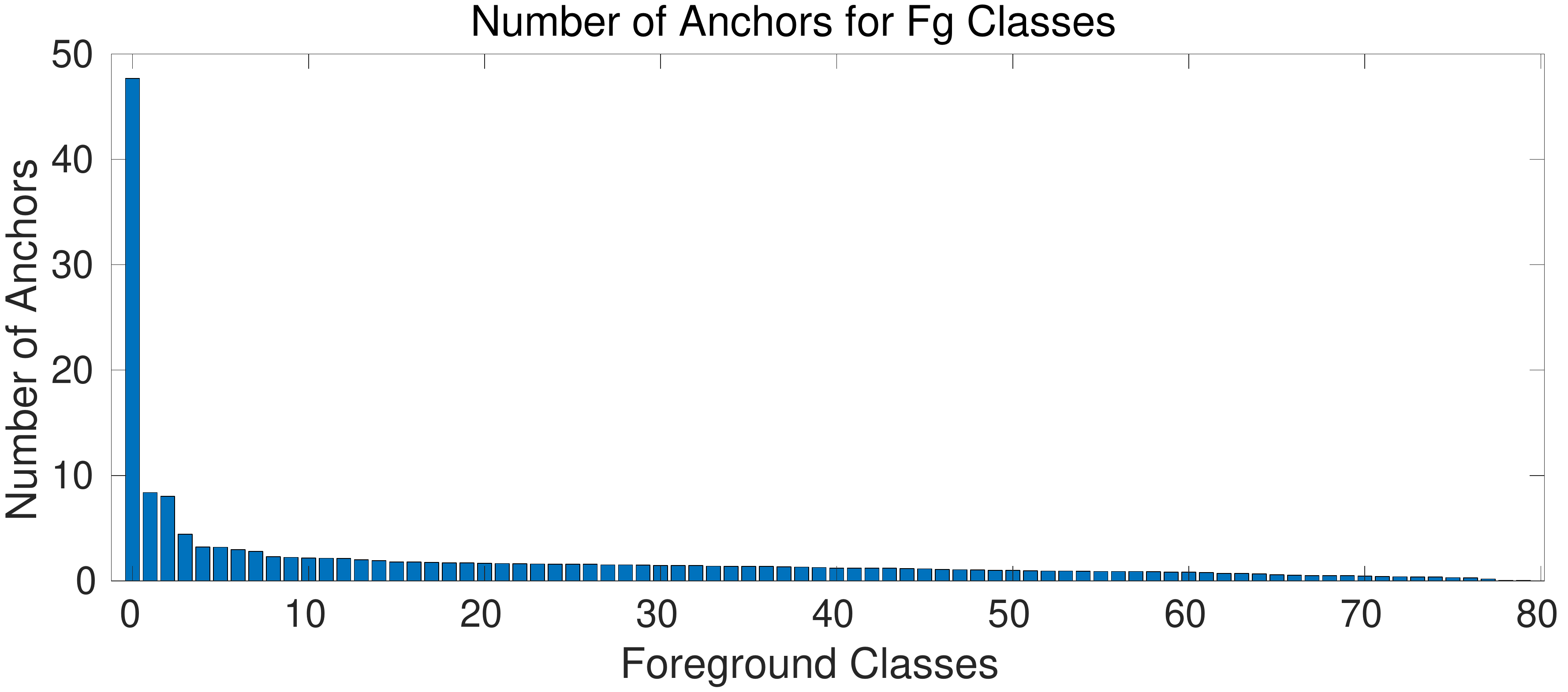}
                \caption{}
        \end{subfigure}
        \caption{Illustration of the class imbalance problems. The numbers of RetinaNet \cite{FocalLoss} anchors on MS-COCO \cite{COCO} are plotted for  foreground-background \textbf{(a)}, and  foreground classes \textbf{(b)}. The values are normalized with the total number of images in the dataset. The figures depict severe imbalance towards some classes.}
        \label{fig:class_imbalance_types}
\end{figure*}

Class imbalance is observed when a class is over-represented, having more examples than others in the dataset. This can occur in two different ways from the object detection perspective: foreground-background imbalance and foreground-foreground imbalance. 

Figure \ref{fig:class_imbalance_types} illustrates the presence of class imbalance. To generate the figure, we apply the default set of anchors from RetinaNet \cite{FocalLoss} on the MS-COCO dataset\footnote{Throughout the text, unless otherwise specified, ``COCO'', ``PASCAL'', and ``Open Images'' respectively correspond to the Pascal VOC 2012 trainval \cite{PASCAL}, MS-COCO 2017 train \cite{COCO}, Open Images v5 training (subset with bounding boxes) \cite{OpenImages} and Objects365 train \cite{Objects365} dataset partitions. And, when unspecified, the backbone is ResNet-50 \cite{ResNet}.} \cite{COCO} and calculated the frequencies for the cases where the IoU of an anchor with a ground truth bounding box exceeds $0.5$ and where it is less than $0.4$ (i.e. it is a background box) following the labeling rule of RetinaNet \cite{FocalLoss}. When an anchor overlaps with a foreground class (with IoU>0.5), we kept a count for each class separately and normalized the resulting frequencies with the number of images in the dataset.

These two types of class imbalance have different characteristics and have been addressed using different types of solutions. Therefore, in the following, we will cover them separately. However, some solutions (e.g. generative modeling) could be employed for both problem types.

\subsection{Foreground-Background Class Imbalance}
\label{subsec:fgbgClassImb}


\noindent\textit{Definition.} In {foreground-background class imbalance}, the over-represented and under-represented classes are background and foreground classes respectively. This type of problem is inevitable because most bounding boxes are labeled as background (i.e. negative) class by the bounding box matching and labeling module as illustrated in Figure \ref{fig:class_imbalance_types}(a). Foreground-background imbalance problem occurs during training and it does not depend on the number of examples per class in the dataset since they do not include any annotation on background. 

\noindent\textit{Solutions.} We can group the solutions for the foreground-background class imbalance into four: (i) hard sampling methods, (ii) soft sampling methods, (iii) sampling-free methods and (iv) generative methods. Each set of methods are explained in detail in the subsections below. 

In sampling methods, the contribution ($w_i$) of a bounding box ($BB_i$) to the loss function is adjusted as follows:
\begin{equation}
    \label{eq:SamplingEq}
    w_i \mathrm{CE}(p_s),
\end{equation}
where $\mathrm{CE}()$ is the cross-entropy loss. Hard and soft sampling approaches differ on the possible values of $w_i$. For the hard sampling approaches, $w_i \in \{0,1\}$, thus a BB is either selected or discarded. For soft sampling approaches, $w_i \in [0,1]$, i.e. the contribution of a sample is adjusted with a weight and each BB is somehow included in training. 

\subsubsection{Hard Sampling Methods}
\label{subsubsec:Hard}

Hard sampling is a commonly-used method for addressing imbalance in object detection. It restricts $w_i$ to be binary; i.e., 0 or 1. In other words, it addresses imbalance by selecting a subset of positive and negative examples (with desired quantities) from a given set of labeled BBs. This selection is performed using heuristic methods and the non-selected examples are ignored for the current iteration. Therefore, each sampled example contributes equally to the loss (i.e. $w_i=1$) and the non-selected examples ($w_i=0$) have no contribution to the training for the current iteration. See Table \ref{tab:FgBgSampling} for a summary of the main approaches.

A straightforward hard-sampling method is \textbf{random sampling}. Despite its simplicity, it is employed in R-CNN family of detectors \cite{RCNN,FasterRCNN} where, for training RPN, 128 positive examples are sampled uniformly at random (out of all positive examples) and 128 negative anchors are sampled in a similar fashion; and 16 positive examples and 48 negative RoIs  are sampled uniformly from each image in the batch at random from within their respective sets, for training the detection network \cite{FastRCNN}. In any case, if the number of positive input bounding boxes is less than the desired values, the mini-batch is padded with randomly sampled negatives. On the other hand, it has been reported that other sampling strategies may perform better when a property of an input box such as its loss value or IoU is taken into account \cite{OHEM,LibraRCNN,PrimeSample}.

The first set of approaches to consider a property of the sampled examples, rather than random sampling, is the \textbf{Hard-example mining methods}\footnote{In this paper, we adopt the boldface font whenever we introduce an approach involving a set of different methods, and the method names themselves are in italic.}. These methods rely on the hypothesis that training a detector more with hard examples (i.e. examples with high losses) leads to better performance. The origins of this hypothesis go back to the \vurgula{bootstrapping} idea in the early works on face detection \cite{BootstrappingFace1,BootstrappingFace2,BootstrappingFace3}, human detection \cite{HOG} and object detection \cite{DPM}. The idea is based on training an initial model using a subset of negative examples, then using the negative examples on which the classifier fails (i.e. hard examples), a new classifier is trained. Multiple classifiers are obtained by applying the same procedure iteratively. Currently deep-learning-based methods also adopt some versions of the hard example mining in order to provide more useful examples by using the loss values of the examples. The first deep object detector to use hard examples in the training was Single-Shot Detector \cite{SSD}, which chooses only the negative examples incurring the highest loss values. A more systematic approach considering the loss values of positive and negative samples is proposed in \vurgula{Online Hard Example Mining} (OHEM) \cite{OHEM}. However, OHEM needs additional memory and causes the training speed to decrease. Considering the efficiency and memory problems of OHEM, \vurgula{IoU-based sampling} \cite{LibraRCNN} was proposed to associate the hardness of the examples with their IoUs and to use a sampling method again for only negative examples rather than computing the loss function for the entire set. In the IoU-based sampling, the IoU interval for the negative samples is divided into $K$ bins and equal number of negative examples are sampled randomly within each bin to promote the samples with higher IoUs, which are expected to have higher loss values.

To improve mining performance, several studies proposed to \textbf{limit the search space} in order to make hard examples easy to mine. \vurgula{Two-stage object detectors} \cite{RFCN,FasterRCNN} are among these methods since they aim to find the most probable bounding boxes (i.e. RoIs) given anchors, and then choose top N RoIs with the highest objectness scores, to which an additional sampling method is applied. Fast R-CNN \cite{FastRCNN} sets the \vurgula{lower bound of IoU} of the negative RoIs to $0.1$ rather than $0$ for promoting hard negatives and then applies random sampling. Kong et al. \cite{RON} proposed a method that learns \vurgula{objectness priors} in an end-to-end setting in order to have a guidance on where to search for the objects. All of the positive examples having an objectness prior larger than a threshold are used during training, while the negative examples are selected such that the desired balance (i.e. $1:3$) is preserved between positive and negative classes. Zhang et al. \cite{AnchorRefine} proposed determining the confidence scores of anchors with the anchor refinement module in a one-stage detection pipeline and again adopted a threshold to eliminate the easy negative anchors. The authors coined their approach as \vurgula{negative anchor filtering}. Nie et al. \cite{EnrichedFeatureGuided} used a cascaded-detection scheme in the SSD pipeline which includes an \vurgula{Objectness Module} before each prediction module. These objectness modules are binary classifiers to filter out the easy anchors.

\subsubsection{Soft Sampling Methods}
\label{subsec:SoftSample}
\newcolumntype{H}{>{\setbox0=\hbox\bgroup}c<{\egroup}@{}}

\begin{table*}
        \centering
        \caption{A toy example depicting the selection methods of common hard and soft sampling methods. One positive and two negative examples are to be chosen from six bounding boxes (drawn at top-right). The properties are the basis for the sampling methods. $p_s$ is the predicted ground truth probability (i.e. positive class probability for positive BBs, and background probability for negative BBs).  If we set a property or hyperparameter for this example, it is shown in the table. For soft sampling methods, the numbers are the weights of each box (i.e. $w_i$). We assumed one foreground class for PISA.}
        \label{tab:FgBgSampling}
        \small
        \setlength\tabcolsep{2pt}
        \resizebox{\textwidth}{!}{
        \begin{tabular}{|c||l|c|c|c|c|c|c|c|c|c|}
            \hline
             & \begin{minipage}{3.6cm} \centering Legend\&Method\end{minipage}  & \begin{minipage}{1.6cm}Considered Property\end{minipage} & \begin{minipage}{1.4cm}Intuition\end{minipage}&\begin{minipage}{1cm} Addit.\\Params. \end{minipage}&\multicolumn{2}{c|}{Positive Examples}&\multicolumn{4}{c|}{Negative Examples}  \\
            \hline
                \rotatebox[origin=c]{90}{Boxes} &
                \begin{minipage}{3.6cm} 
                \vspace*{0.1cm}
                \begin{tikzpicture} 
                    \draw[red] (4,1) -- (5,1) node[right, black]{\small{Negative BB}};
                    \draw[green] (4,0.5) -- (5,0.5) node[right, black]{\small{Positive BB}};
                    \draw[blue] (4,0) -- (5,0) node[right, black]{\small{Ground Truth}};
                \end{tikzpicture} \\
                \hspace*{0.4cm} \cmark \hspace*{0.33cm} Selected BB\\
                \hspace*{0.4cm} \xmark \hspace*{0.33cm} Discarded BB 
                \end{minipage}
                
                &
                N/A
                &
                N/A
                &
                N/A
                &
                
                \begin{minipage}{2cm}
                \centering
                \vspace{0.5cm}
                \begin{tikzpicture} 
                    \draw[line width=0.4mm, green] (1,1) rectangle (2,2);
                    \draw [line width=0.4mm, blue] (1.1, 0.9) rectangle (2.1, 1.9);
                \end{tikzpicture}
                
                $p_s$=0.8 \\
                Loss = 0.22 \\
                IoU = 0.95
                \end{minipage}
                
                &
                \begin{minipage}{2cm}
                \centering
                \vspace{0.5cm}
                \begin{tikzpicture} 
                    \draw[line width=0.4mm, green] (1,1.3) rectangle (2,2.3);
                    \draw [line width=0.4mm, blue] (1, 0.9) rectangle (2, 1.9);
                \end{tikzpicture}
                
                    $p_s$=0.2 \\
                    Loss = 1.61 \\
                    IoU = 0.55 
                \end{minipage}
                
                &
                \begin{minipage}{2cm}
                \centering
                \vspace{0.5cm}    
                \begin{tikzpicture} 
                    \draw[line width=0.4mm, red] (1,1.3) rectangle (2,2.3);
                    \draw [line width=0.4mm, blue] (1, 0.9) rectangle (2, 1.9);
                \end{tikzpicture}
                
                    $p_s$=0.1 \\
                    Loss = 2.30 \\ 
                    IoU = 0.35 
                
                \end{minipage}
                &
                \begin{minipage}{2cm}
                \centering
                \vspace{0.5cm}
                \begin{tikzpicture} 
                    \draw[line width=0.4mm, red] (1,1.6) rectangle (2,2.6);
                    \draw [line width=0.4mm, blue] (1, 0.9) rectangle (2, 1.9);
                \end{tikzpicture}
                
                    $p_s$=0.3 \\
                    Loss = 1.20 \\
                    IoU = 0.15
                
                \end{minipage}
                &
                \begin{minipage}{2cm}
                \centering
                \vspace{0.5cm}
                \begin{tikzpicture} 
                    \draw[line width=0.4mm, red] (1.5,1.6) rectangle (2.5,2.6);
                    \draw [line width=0.4mm, blue] (1, 0.9) rectangle (2, 1.9);
                \end{tikzpicture}

                    $p_s$=0.6 \\
                    Loss = 0.51 \\
                    IoU = 0.06

                \end{minipage}
                &
                \begin{minipage}{2cm}
                \centering
                \vspace{0.4cm}
                \begin{tikzpicture} 
                    \draw[line width=0.4mm, red] (1,2) rectangle (2,3);
                    \draw [line width=0.4mm, blue] (1, 0.9) rectangle (2, 1.9);
                \end{tikzpicture}
                
                    $p_s$=1.0 \\
                    Loss = 0.00 \\
                    IoU = 0.00
                
                \vspace{0.1cm}
                \end{minipage}
                \\\hline
                \hline
                \multirow{7}{*}{\rotatebox[origin=c]{90}{Hard Sampling \quad \quad \quad \quad \quad \quad}} & Random Sampling & $-$ &\begin{minipage}[c][0.7cm][c]{0.2\textwidth}\centering Samples uniformly among boxes\end{minipage} & $-$ & \multicolumn{2}{c|}{Random Sample$\times$1} & \multicolumn{4}{c|}{Random Sample$\times$2} \\\cline{2-11}
                                                                & \begin{minipage}{3.6cm}SSD \cite{SSD}\end{minipage} & Loss &\begin{minipage}[c][0.7cm][c]{0.2\textwidth}\centering Chooses neg. with max loss\end{minipage} & $-$&\multicolumn{2}{c|}{Random Sample$\times$1} & \cmark & \cmark & \xmark & \xmark \\\cline{2-11}
                                                                & \begin{minipage}{3.6cm}OHEM \cite{OHEM}\end{minipage}& Loss &\begin{minipage}[c][0.7cm][c]{0.2\textwidth}\centering Chooses pos.\&neg. with max loss\end{minipage}& $-$ & \xmark & \cmark & \cmark & \cmark & \xmark & \xmark \\\cline{2-11}
                                                                & \begin{minipage}{3.6cm}IoU-Based Sampling \cite{LibraRCNN}\end{minipage}& IoU &\begin{minipage}[c][1.2cm][c]{0.2\textwidth}\centering Samples equal number of negatives from IoU bins\end{minipage}& K=2 & \multicolumn{2}{c|}{\begin{minipage}{3.6cm}Does not specify a criterion for positives\end{minipage}} 
                                                                & \begin{minipage}{1.4cm}\centering \scriptsize{Random Sample$\times$ 1}\end{minipage} & \multicolumn{3}{c|}{Random Sample$\times$ 1}\\\cline{2-11}
                                                                & \begin{minipage}{3.6cm}IoU Lower Bound \cite{FastRCNN}\end{minipage}& IoU=0.05 & \begin{minipage}[c][1.0cm][c]{0.2\textwidth}\centering Discards neg. having IoU less than lower bound\end{minipage}&$-$& \multicolumn{2}{c|}{\begin{minipage}{3.6cm}Does not specify a criterion for positives\end{minipage}} & \multicolumn{3}{c|}{Random Sample $\times$ 2} & \xmark \\\cline{2-11}
                                                                & 
                                        \begin{minipage}{3.6cm}Objectness Prior \cite{RON}\end{minipage}
                                        & $-$ &\begin{minipage}[c][0.7cm][c]{0.2\textwidth}\centering Learns to predict objectness priors for pos.\end{minipage} &$-$& \multicolumn{2}{c|}{\begin{minipage}{3.6cm}Chooses all with prior larger than threshold.\end{minipage}} & \multicolumn{4}{c|}{\begin{minipage}{8cm}The negatives are sampled randomly by preserving 1:3 ratio with positives. So, fixed batch size is not considered.\end{minipage}} \\\cline{2-11}
                                                                &
                                        \begin{minipage}{3.6cm}Negative Anchor Filtering \cite{AnchorRefine}\end{minipage}
                                        & $p_s=0.9$ &\begin{minipage}[c][0.9cm][c]{0.2\textwidth}\centering Discards negs. less than threshold\end{minipage}& $-$ & \multicolumn{2}{c|}{\begin{minipage}{3.6cm}Does not specify a criterion for positives\end{minipage}} & \multicolumn{3}{c|}{\begin{minipage}{6cm} Uses OHEM for the remaining negatives. In this case, the first two boxes are selected \end{minipage}} & \xmark\\\hline
                                                              \hline
                \multirow{3}{*}{\rotatebox[origin=rc]{90}{Soft Sampling \quad }} & 
                                    \begin{minipage}{3.6cm}Focal Loss \cite{FocalLoss}\end{minipage}&$p_s$&\begin{minipage}[c][0.7cm][c]{0.2\textwidth}\centering Promotes examples with larger loss values\end{minipage}& $\gamma=1$ & 0.2 & 0.8 & 0.9 & 0.7 & 0.4 & 0.0 \\\cline{2-11}
                                                                &
                                    \begin{minipage}{3.6cm}Gradient Harmonizing Mechanism \cite{gradientharmonizing}\end{minipage}
                                    & $p_s$ &\begin{minipage}[c][1.2cm][c]{0.2\textwidth}\centering Promotes hard examples by suppressing effects of outliers\end{minipage} &$\epsilon=0.4$ &0.66&0.66&0.60&0.66&1.2&0.60 \\\cline{2-11}
                                                                & 
                                    \begin{minipage}{3.6cm}Prime Sample Attention  \cite{PrimeSample}\end{minipage}& IoU, $p$ &\begin{minipage}[c][1.5cm][c]{0.2\textwidth}\centering Promotes examples with larger IoUs for positives and larger fg scores for negatives (i.e. 1-$p_s$ here) \end{minipage}& \begin{minipage}{1.1cm}\centering $\gamma=1$\\$\beta=0$\end{minipage} &1.00&0.75&1.00&0.75&0.50&0.25 \\
                                    \cline{2-11} 
                                                                                    
                                    \hline

        \end{tabular}}
    \end{table*}

Soft sampling adjusts the contribution ($w_i$) of each example by its relative importance to the training process. This way, unlike hard sampling, no sample is discarded and the whole dataset is utilized for updating the parameters. See Table \ref{tab:FgBgSampling} for a summary of the main approaches.

A straightforward approach is to use constant coefficients for both the foreground and background classes. YOLO \cite{YOLO}, having less number of anchors compared to other one-stage methods such as SSD \cite{SSD} and RetinaNet \cite{FocalLoss}, is a straightforward example for soft sampling in which the loss values of the examples from the background class are halved (i.e. $w_i=0.5$).

\vurgula{Focal Loss} \cite{FocalLoss} is the pioneer example that dynamically assigns more weight to the hard examples as follows:
\begin{equation}
    \label{eq:FocalLossWeight}
    w_i=(1-p_s)^\gamma,
\end{equation}
\noindent where $p_s$ is the estimated probability for the ground-truth class. Since lower $p_s$ implies a larger error, Equation \eqref{eq:FocalLossWeight} promotes harder examples. Note that when $\gamma=0$, focal loss degenerates to vanilla cross entropy loss and Lin et al. \cite{FocalLoss} showed that  $\gamma=2$ ensures a good trade-off between hard and easy examples for their architecture.

Similar to focal loss \cite{FocalLoss}, \vurgula{Gradient Harmonizing Mechanism} (GHM) \cite{gradientharmonizing}  suppresses gradients originating from easy positives and negatives. The authors first observed that there are too many samples with small gradient norm, only limited number of samples with medium gradient norm and significantly large amount of samples with large gradient norm. Unlike focal loss, GHM is a counting-based approach which counts the number of examples with similar gradient norm and penalizes the loss of a sample if there are many samples with similar gradients as follows:
\begin{align}
    w_i= \frac{1}{G(BB_i)/m},
\end{align}
where $G(BB_i)$ is the count of samples whose gradient norm is close to the gradient norm of $BB_i$; and $m$ is the number of input bounding boxes in the batch. In this sense, the GHM method implicitly assumes that easy examples are those with too many similar gradients.
%
Different from other methods, GHM is shown to be useful not only for classification task but also for the regression task. In addition, since the purpose is to balance the gradients within each task, this method is also relevant to the ``imbalance in regression loss'' discussed in Section \ref{subsec:Regression}.

Different from the latter soft sampling methods, \vurgula{PrIme Sample Attention} (PISA) \cite{PrimeSample} assigns weights to positive and negative examples based on different criteria. While the positive ones with higher IoUs are favored, the negatives with larger foreground classification scores are promoted. More specifically, PISA first ranks the examples for each class based on their desired property (IoU or classification score) and calculates a normalized rank, $\Bar{r}_i$, for each example $i$ as follows:
\begin{align}
    \label{eq:PrimeSamplesImp}
    \Bar{r}_i=\frac{N_{max}-r_i}{N_{max}},
\end{align}
where $r_i$ ($0\leq r_i\leq N_j-1$) is the rank of the $i$th example and $N_{max}$ is the maximum number of examples over all classes in the batch. Based on the normalized rank, the weight of each example is defined as:
\begin{align}
    \label{eq:PrimeSamplesWeight}
    w_i= ((1-\beta) \Bar{r}_i + \beta)^\gamma,
\end{align}
where $\beta$ adjusts the contribution of the normalized rank and hence, the minimum sample weight; and $\gamma$ is the modulating factor again. These parameters are validated for positives and negatives independently. Note that the balancing strategy in Equations \eqref{eq:PrimeSamplesImp} and \eqref{eq:PrimeSamplesWeight} increases the contribution of the samples with high IoUs for positives and high scores for negatives to the loss.



\subsubsection{Sampling-Free Methods}
Recently, alternative methods emerged in order to avoid the aforementioned hand-crafted sampling heuristics and therefore, decrease the number of hyperparameters during training. For this purpose, Chen et al. \cite{ResidualObjectness} added an objectness branch to the detection network in order to predict \vurgula{Residual Objectness} scores. While this new branch tackles the foreground-background imbalance, the classification branch handles only the positive classes.  During inference, classification scores are obtained by multiplying classification and objectness branch outputs. The authors showed that such a cascaded pipeline improves  the performance. This architecture is trained using vanilla  cross entropy loss. 

Another recent approach \cite{SamplingHeuristics} suggests that if the hyperparameters are set appropriately, not only the detector can be trained without any sampling mechanism but also the performance can be improved. Accordingly, the authors propose methods for setting the initialization bias, loss weighting (see Section \ref{sec:ObjImb}) and class-adaptive threshold, and in such a way they trained the network using vanilla cross entropy loss achieving better performance.

Finally, an alternative method is to directly model the final performance measure and weigh examples based on this model. This approach is adopted by \vurgula{AP Loss} \cite{APLoss} which formulates the classification part of the loss as a ranking task (see also \vurgula{DR Loss} \cite{DRLoss} which also uses a ranking method to define a classification loss based on Hinge Loss) and uses average precision (AP) as the loss function for this task.  
In this method, the confidence scores are first transformed as $x_{ij}=-(p^i-p^j)$ such that $p^i$ is the confidence score of the $i$th bounding box. Then, based on the transformed values, the primary terms of the AP loss are computed as (presented here in a simplified form):
\begin{align}
    U_{ij}= \frac{\mathbb{I}(x_{ij}>0)}{1+ \sum\limits_{k \in \mathcal{P} \cup\mathcal{N}} \mathbb{I}(x_{ik}>0)},
\end{align}
where $\mathcal{P},\mathcal{N}$ are the set of positive and negative labeled examples respectively. Note that $U_{ij}$ is zero if $p_i < p_j$ and a positive value less than one otherwise. With this quantity, AP loss is defined as follows:
\begin{align}
    \label{eq:APLoss}
    L_{AP}= \frac{1}{|\mathcal{P}|} \sum_{i,j} U_{ij} y_{ij},
\end{align}
where $y_{ij}$ is the ranking label, set to $1$ only if the $i$th box is a foreground bounding box and $j$th box is a background bounding box. 

To make this differentiable, the authors propose a novel error-driven update rule, which performs better than the previous works aiming to include average precision as a training objective \cite{APLossAugmentedInference,MAPTrain}. However, this optimization algorithm is beyond the scope of our work.

\subsubsection{Generative Methods}
\label{subsec:generative}

Unlike sampling-based and sampling-free methods, generative methods address imbalance by directly producing and injecting artificial samples into the training dataset. Table \ref{tab:GenerativeModels} presents a summary of the main approaches.

One approach is to use generative adversarial networks (GANs). A merit of GANs is that they adapt themselves to generate harder examples during training since the loss values of these networks are directly based on the classification accuracy of the generated examples in the final detection. An example is the \vurgula{Adversarial-Fast-RCNN} model \cite{AFastRCNN}, which generates hard examples with occlusion and various deformations. In this method, the generative manipulation is directly performed at the feature-level, by taking the fixed size feature maps after RoI standardization layers (i.e. RoI pooling \cite{FastRCNN}). For this purpose, they proposed two networks: (i) adversarial spatial dropout network for occluded feature map generation, and (ii) adversarial spatial transformer network for deformed (transformed) feature map generation. These two networks are placed sequentially in the network design in order to provide harder examples and they are integrated into the conventional object training pipeline in an end-to-end manner. 

Alternatively, artificial images can be produced to augment the dataset \cite{CutPasteLearn,ModelingVisualcontext,GenerateSynthetic,PSIS} by generating composite images in which multiple crops and/or images are blended. A straightforward approach is to randomly place cropped objects onto images as done by Dwibedi et al. \cite{CutPasteLearn}. However, the produced images may look unrealistic. This problem is alleviated by determining where to paste and the size of the pasted region according to the visual context \cite{ModelingVisualcontext}. In a similar vein, the objects can be swapped between images: \vurgula{Progressive and Selective Instance-Switching} (PSIS) \cite{PSIS} swaps single objects belonging to the same class between a pair of images considering also the scales and shapes of the candidate instances. Producing images by swapping objects of low-performing classes improves detection quality. For this reason, they use the performance ranking of the classes while determining the objects to swap and the number of images to generate. 

A more prominent approach is to use GANs to generate images rather than copying existing objects: an example is \vurgula{Task Aware Data Synthesis} \cite{GenerateSynthetic} which uses three competing networks to generate hard examples: a synthesizer, a discriminator and a target network where the synthesizer is expected to fool both the discriminator and the target network by yielding high quality synthetic hard images. Given an image and a foreground object mask, the synthesizer aims to place the foreground object mask onto the image to produce realistic hard examples. The discriminator is adopted in order to enforce the synthesizer towards realistic composite images. The target network is an object detector, initially pretrained to have a baseline performance.  

Instead of generating images, the \vurgula{Positive RoI (pRoI) Generator} \cite{BBSampling} generates a set of positive RoIs with given IoU, BB relative spatial and foreground class distributions. The approach relies on a bounding box generator that is able to generate bounding boxes (i.e. positive example) with the desired IoU with a given bounding box (i.e. ground truth). Noting that the IoU of an input BB is related to its hardness \cite{LibraRCNN}, the pRoI generator is a basis for simulating, thereby analyzing, hard sampling methods (Section \ref{subsubsec:Hard}).

\begin{table*}
    \centering    
    \caption{Comparison of major generative methods addressing class imbalance.}
    \begin{tabular}{|p{5cm}|p{11cm}|}
         \hline
         \multicolumn{1}{|c|}{\textbf{Generative Method}}& \multicolumn{1}{|c|}{\textbf{Generates}} \\ \hline
         Adversarial-Fast-RCNN \cite{AFastRCNN} &Occluded and spatially transformed features during RoI pooling in order to make the examples harder \\ \hline 
         Task Aware Data Synthesis \cite{GenerateSynthetic} &Images with hard examples in a GAN setting in which given foreground mask is to be placed onto the given image by the generator.\\ \hline 
         PSIS \cite{PSIS} &Images by switching the instances between existing images considering the performance of the class during training\\ \hline
         pRoI Generator \cite{BBSampling} &Positive RoIs (i.e. BBs) following desired IoU, spatial and foreground class distributions\\ \hline 
    \end{tabular}
    \label{tab:GenerativeModels}
\end{table*}
\subsection{Foreground-Foreground Class Imbalance}
\label{subsec:fgClassImb}

\begin{figure*}
        \captionsetup[subfigure]{}
        \centering
        \begin{subfigure}[b]{0.37\textwidth}
                \includegraphics[width=\textwidth]{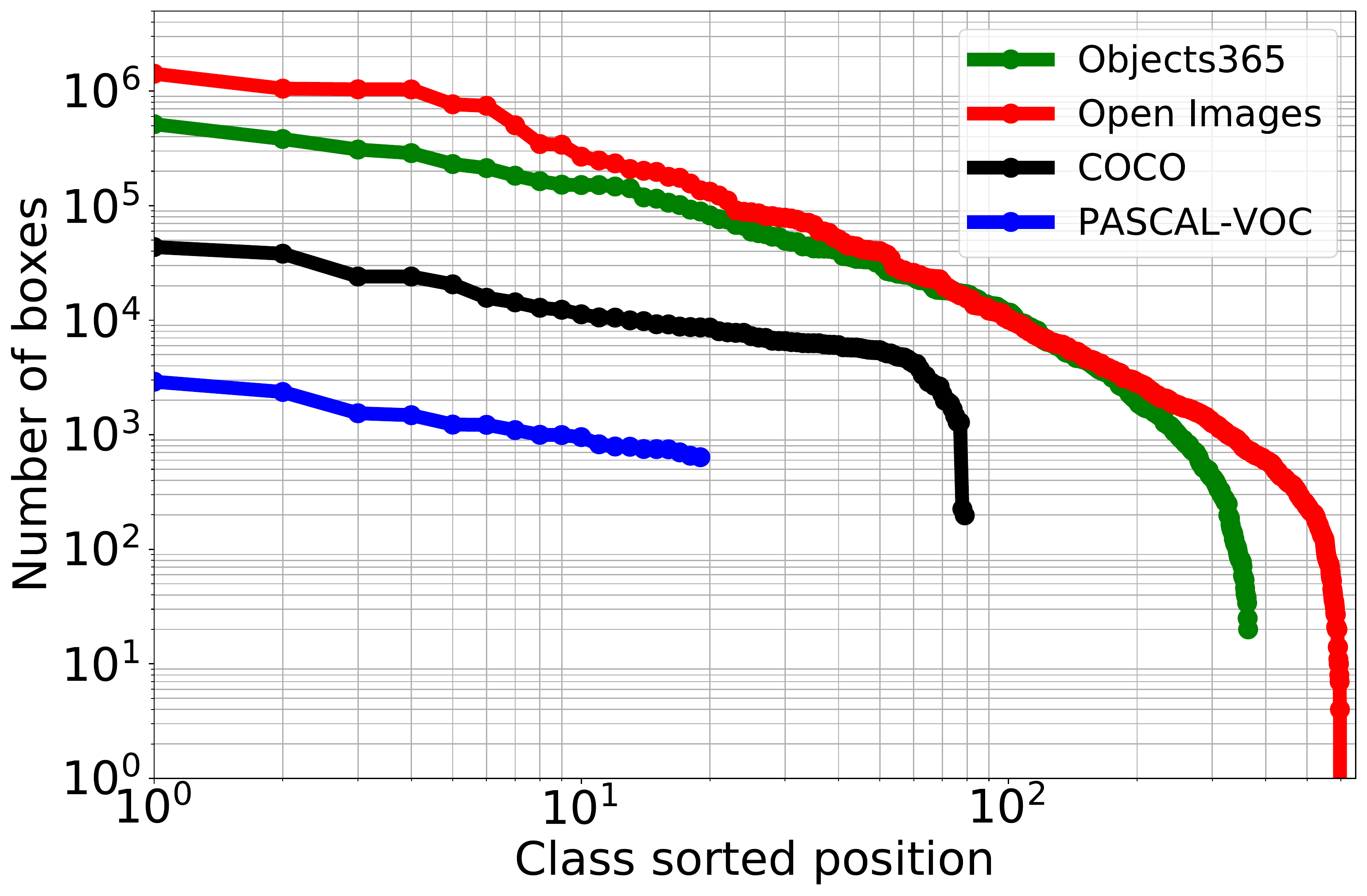}
                \caption{}
        \end{subfigure}       
        \begin{subfigure}[b]{0.26\textwidth}
                \includegraphics[width=\textwidth]{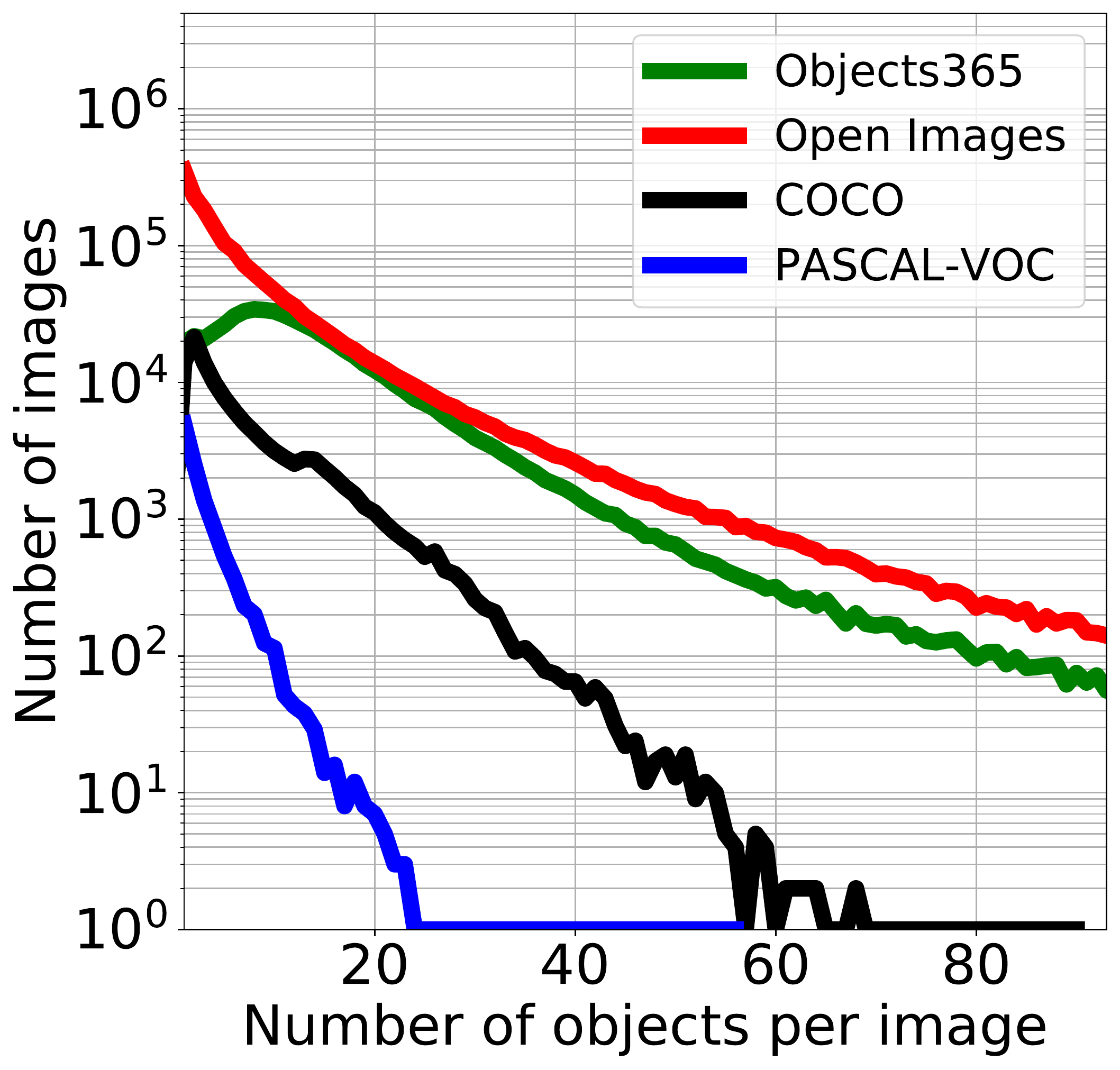}
                \caption{}
        \end{subfigure}
        \begin{subfigure}[b]{0.275\textwidth}
                \includegraphics[width=\textwidth]{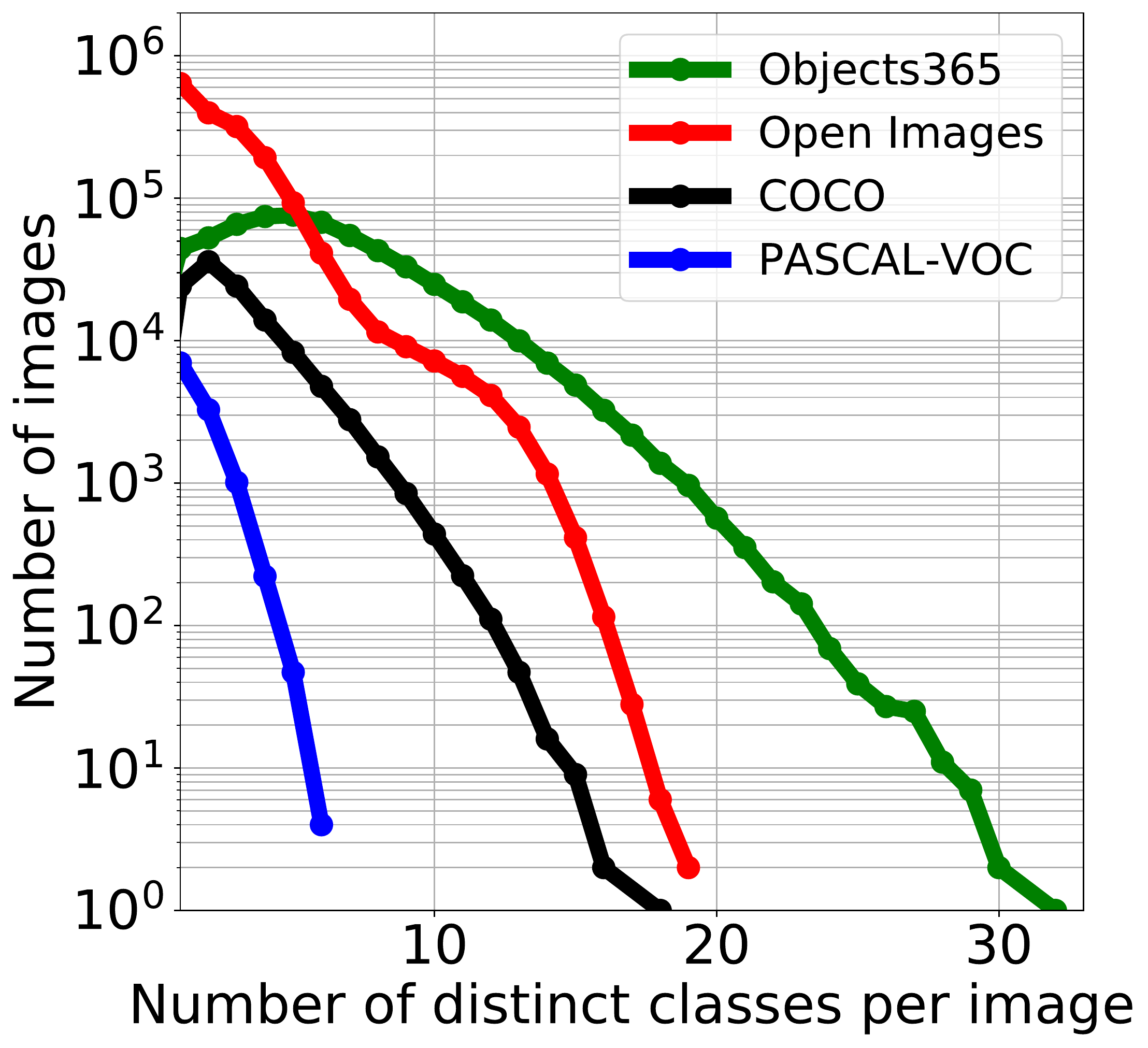}
                \caption{}
        \end{subfigure}        
        \caption{Some statistics of common datasets (training sets). For readability, the $y$ axes are in logarithmic scale. \textbf{(a)} The total number of examples from each class.
        \textbf{(b)} The number of images vs. the number of examples.
        \textbf{(c)} The number of images vs. the number of classes. 
        (This figure is inspired from Kuznetsova et al. \cite{OpenImages} but calculated and drawn from scratch).} 
        \label{fig:DatasetStats}
\end{figure*}



\noindent\textit{Definition.} In {foreground-foreground class imbalance}, the over-represented and the under-represented classes are both foreground classes. This imbalance among the foreground classes has not attracted as much interest as foreground-background imbalance. We discuss this problem in two categories according to their origin: (i) dataset-level, and (ii) batch-level. 


\subsubsection{Foreground-Foreground Imbalance Owing to the Dataset}
\label{subsubsec:DatasetLevel}

\noindent\textit{Definition.} Objects exist at different frequencies in nature, and therefore, there is naturally an imbalance among the object classes in datasets -- see Figure \ref{fig:DatasetStats}(a) which shows that the datasets suffer from significant gap in class examples. For this reason, overfitting in favor of the over-represented classes may be inevitable for naive approaches on such datasets.

\noindent\textit{Solutions.} Owing to the fact that some of the generative methods are able to generate new images or bounding boxes (see Section \ref{subsec:generative}) that cannot be obtained by the conventional training pipeline, these methods can also be adopted to alleviate the foreground-foreground class imbalance problem. 

Another method addressing this imbalance problem from the object detection perspective is proposed by Ouyang et al. \cite{FgClassImbalance}. Their method of \vurgula{finetuning long-tail distribution for object detection} (here, ``long tail'' corresponds to under-represented classes) provides an analysis on the effects of this level to the training process and uses clustering based on visual similarity.  In their analysis, two factors affecting the training are identified: (i) the accuracy of the prediction, and (ii) the number of examples. Based on this observation, they handcrafted a similarity measure among classes based on the inner product of the features of the last layer of the pretrained backbone network (i.e. GoogLe Net \cite{GoogLeNet}), and  grouped the classes hierarchically in order to compensate for dataset-level foreground class imbalance. For each node in the designed hierarchy tree, a classifier is learned based on the confidence scores of the classifier. The leaves of the tree are basically an SVM classifier that determines the final detection for the given input BB.  

\subsubsection{Foreground-Foreground Imbalance Owing to the Batch}
\label{subsubsec:MinibatchLevel}

\noindent\textit{Definition.} The distribution of classes in a batch may be uneven, introducing a bias in learning. To illustrate the batch-level foreground imbalance, Figure \ref{fig:class_imbalance_types}(b) provides the mean number of anchors per class on the MS COCO dataset \cite{COCO}. A random sampling approach is expected to allocate an unbalanced number of positive examples in favor of the one with more anchors, which may lead the model to be biased in favor of the over-represented class during training. Also see Figure \ref{fig:DatasetStats}(b) and (c), which display that the numbers of objects and classes in an image vary significantly.

\noindent\textit{Solutions.} A solution for this problem, \vurgula{Online Foreground Balanced (OFB) sampling} by Oksuz et a. \cite{BBSampling}, shows that the foreground-foreground class imbalance problem can be alleviated at the batch level by assigning probabilities to each bounding box to be sampled, so that the distribution of different classes within a batch is uniform. In other words, the approach aims to promote the classes with lower number of positive examples during sampling. While the method is efficient, the performance improvement is not significant. Moreover, the authors have not provided an analysis on whether or not batch-level imbalance causes a bias in learning, therefore, we discuss this as an open problem in Section \ref{subsec:OpenProblemsClImb}. 

\subsection{Comparative Summary}

In early deep object detectors,  hard sampling methods were commonly used to ensure balanced training.  OHEM \cite{OHEM} was one of the most effective methods with  a relative improvement of $ 14.7\%$ over the Fast R-CNN baseline (with VGG-16 backbone) on MS COCO 2015. On the other hand, recent methods aim to use all examples either by soft sampling or without sampling. Among the seven papers investigated under these categories, six of them have been published during the last year. The pioneering Focal Loss method \cite{FocalLoss} achieves $9.3 \%$ relative improvement compared to the baseline alpha-balanced cross entropy on Retina-Net (i.e. from $31.1$ to $34.0$ mAP\footnote{Unless otherwise stated, MS COCO 2017 minival results are reported using the COCO-style mAP.}). AP Loss \cite{APLoss}, with a relative improvement of $3.9 \%$ over Focal Loss (i.e. from $33.9$ to $35.0$ mAP), is also promising. 

Sampling methods for positive examples are not many. One approach, prime sample attention (PISA) \cite{PrimeSample}, reported a relative improvement of $4.4 \%$ (from $36.4$ to $38.0$ mAP when applied to  positives only) and showed that sampling also matters for positives. We notice several crucial points regarding prime sampling. First of all, contrary to the popular belief that hard examples are more preferable over easy examples, PISA shows that, if balanced properly, the positive examples with higher IoUs, which normally incur smaller loss values, are more useful for training compared to OHEM \cite{OHEM} applied to positives. Moreover, the results suggest that the major improvement of the method is on localization since there is no performance improvement in $AP@0.50$, and  there is significant improvement for APs with higher IoUs (i.e. up to $2.6\%$ in $AP@0.75$). As a result, the improvement can be due to the changing nature of the IoU distribution rather than presenting more descriptive samples to the classifier since the classifier performs worse but the regressor improves (see the discussion on IoU distribution imbalance in Section \ref{subsec:IoUImb}). 

\subsection{Open Issues}
\label{subsec:OpenProblemsClImb}

As we have highlighted before, class imbalance problem can be analyzed in two main categories: foreground-background class imbalance and foreground-foreground class imbalance. In the following, we identify issues to be addressed with a more focus on foreground-foreground imbalance since it has received less attention.

\subsubsection{Sampling More Useful Examples}

\noindent\textit{Open Issue.} Many criteria to identify useful examples (e.g. hard example, example with higher IoUs etc.) for both positive and negative classes have been proposed. However, recent studies point out interesting phenomena that need further investigation: (i) For the foreground-background class imbalance, soft sampling methods have become more prominent and optimal weighting of examples needs further investigation. (ii) Hard example mining \cite{OHEM} for the positive examples is challenged by the idea that favors examples with higher IoUs, i.e. the easier ones \cite{PrimeSample}. (iii) The usefulness of the examples are not considered in terms of foreground-foreground class imbalance.

\noindent\textit{Explanation\footnote{Issues that may not be immediately clear include a detailing explanation section}.} In terms of the usefulness of the examples, there are two criteria to be identified: (i) The usefulness of the background examples, and (ii) the usefulness of the foreground examples. 

The existing approaches mostly concentrated around the first criterion using different properties (i.e. IoU, loss value, ground truth confidence score) to sample a useful example for the background. However, the debate is still open after Li et al. \cite{gradientharmonizing} showed that there are large number of outliers during sampling using these properties, which will result in higher loss values and lower confidence scores. Moreover, the methods preferring a weighting over all the examples \cite{FocalLoss,gradientharmonizing} rather than discarding a large portion of samples have proven to yield more performance improvement. For this reason, currently, soft sampling approaches that assign weights to the examples are more popular, but which negative examples are more useful needs more investigation.

For the foreground examples, Cao et al. \cite{PrimeSample} apply a specific sampling methodology to the positives based on the IoU, which has proven useful. Note that, this idea conflicts with the hard example mining approach since while OHEM \cite{OHEM} offers to pick the difficult samples, prime samples concentrate on the positive samples with higher IoUs with the ground truth, namely the easier ones. For this reason, currently it seems that the usefulness of the examples is different for the positives and negatives. To sum up, further investigation is required for identifying the best set of examples, or figuring out how to weigh the positive examples during training.

Moreover, sampling methods have  proven to be useful for foreground-background class imbalance, however, their effectiveness for the foreground-foreground class imbalance problem needs to be investigated. 

\subsubsection{Foreground-Foreground Class Imbalance Problem}

\noindent\textit{Open Issue.} Foreground-foreground class imbalance has not been addressed as thoroughly as foreground-background class imbalance. For instance, it is still not known why a class performs worse than others; a recent work \cite{PSIS} discredits the correlation between the number of examples in a class and its detection performance. Moreover, despite its differences, the rich literature for addressing imbalance in image classification has not been utilized in object detection.

\noindent\textit{Explanation.} One important finding of a recent study \cite{PSIS} is that a class with the fewest examples in the dataset can yield one of the best detection performances and thus the total number of instances in the dataset is not the only issue to balance the performance of foreground classes. Such discrepancies presents the necessity of an in-depth  analysis  to identify the root cause and investigate for better sampling mechanisms to employ while balancing a dataset. 

Moreover, we identify a similarity and a difference between the class imbalance from image classification perspective and the foreground-foreground class imbalance problem. The similarity is that neither has a background class. On the other hand, the input BBs are labeled and sampled in an online fashion in object detection, which makes the data that the detector is trained with not static. 

Class imbalance problem is addressed in image classification from a larger scope by using not only classical over-sampling and under-sampling methods but also (i) transfer learning to transfer the information from over-represented to under-represented classes, (ii) data redundancy methods to be useful for under-sampling the over-represented classes, (iii) weak supervision in order to be used in favor of under-represented class and (iv) a specific loss function for balancing foreground classes. Such approaches can be adopted for addressing foreground-foreground imbalance in object detection as well.

\subsubsection{Foreground-Foreground Imbalance Owing to the Batch}
\label{subsubsec:OpenProblemsBatchLevel}

\noindent\textit{Open Issue.} The distribution of the foreground classes in a batch might be very different than that of the overall dataset, especially when the batch size is small. An important question is whether this may have an influence on the overall performance.

\begin{figure}
        \captionsetup[subfigure]{}
        \centering
        \begin{subfigure}[b]{0.25\textwidth}
                \includegraphics[width=\textwidth]{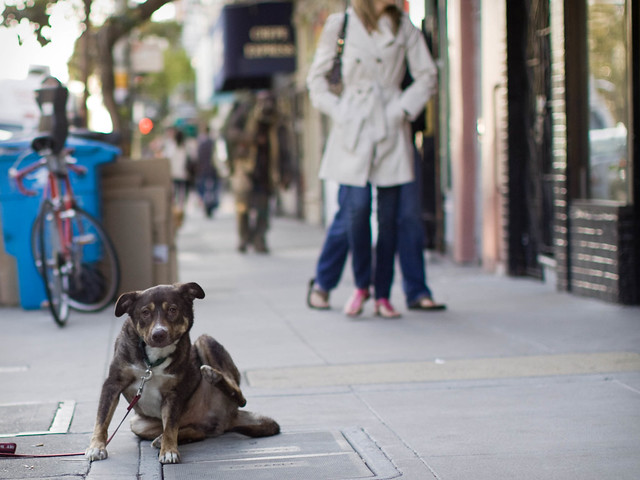}
        \end{subfigure}       
        \begin{subfigure}[b]{0.14\textwidth}
                \includegraphics[width=\textwidth]{figures/unusual.jpg}
        \end{subfigure}
        \vskip\baselineskip
        \begin{subfigure}[b]{0.18\textwidth}
                \includegraphics[width=\textwidth]{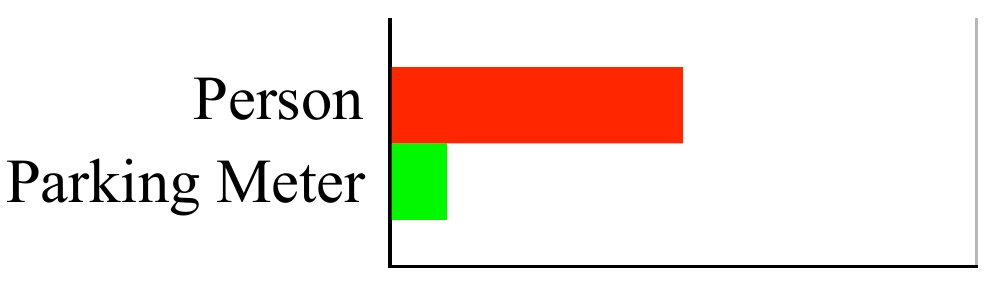}
                \caption{}
        \end{subfigure}   
        \begin{subfigure}[b]{0.18\textwidth}
                \includegraphics[width=\textwidth]{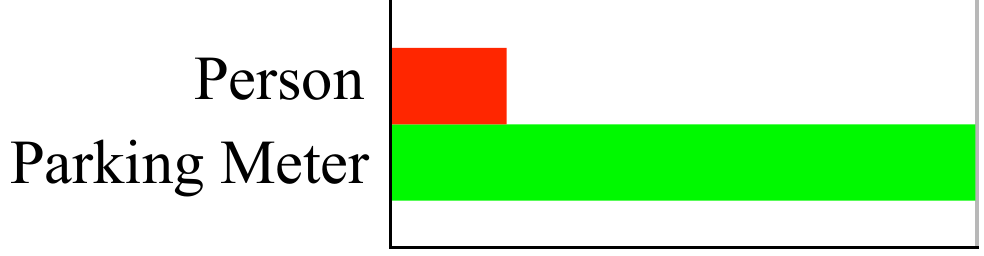}
                \caption{}
        \end{subfigure}             
        \caption{Illustration of batch-level class imbalance.  \textbf{(a)} An example that is consistent with the overall dataset (person class has more instances than parking meter). \textbf{(b)} An example that has a different distribution from the dataset. Images are from the MS COCO dataset.}
        \label{fig:DatasetEx}
\end{figure}

\noindent\textit{Explanation.} A specific example is provided in Figure \ref{fig:DatasetEx} from the MS COCO dataset \cite{COCO}: An over-represented class (`person' class in this example) across the dataset may be under-represented in a batch or vice versa. Similar to the foreground-background class imbalance problem, over-representing a class in a batch will increase the probability of the corresponding class to dominate more.

 Even when a batch has uniform foreground-foreground class distribution, an imbalance may occur during  sampling due to the fact that sampling algorithms either select a subset or apply a weighting to the large number of boxes. If the sampling mechanism tends to choose the samples from specific classes in the image,  balancing the dataset (or the batch) may not be sufficient to address the foreground-foreground class imbalance entirely. 
 
\subsubsection{Ranking-Based Loss Functions}
\noindent\textit{Open Issue.} AP Loss \cite{APLoss} sorts the confidence scores pertaining to all classes together to make a ranking between the detection boxes. However, this conflicts with the observation that the optimal confidence scores vary from class to class \cite{LRP}.

\noindent\textit{Explanation.} Chen et al. \cite{APLoss} use all the confidence scores all together without paying attention to the fact that the optimal threshold per class may vary \cite{LRP}. Therefore, a method that sorts the confidence scores in a class-specific manner might achieve a better performance. Addressing this issue is an open problem.

\section{Imbalance 2: Scale Imbalance} 
\label{sec:Scaleimb}

 \begin{figure*}
        \captionsetup[subfigure]{}
        \centering
        \begin{subfigure}[b]{0.25\textwidth}
                \includegraphics[width=\textwidth]{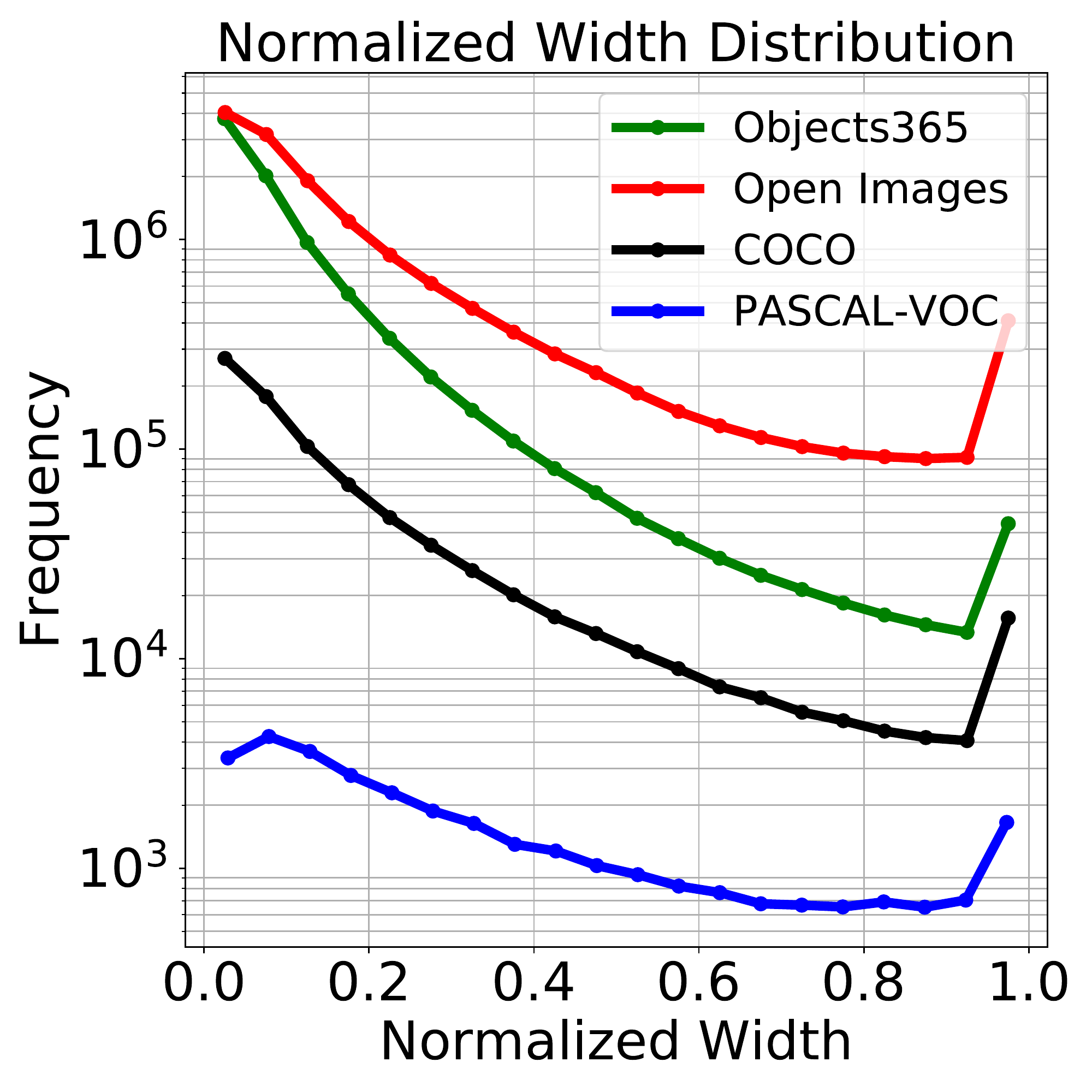}
                \caption{}
        \end{subfigure}       
        \begin{subfigure}[b]{0.25\textwidth}
                \includegraphics[width=\textwidth]{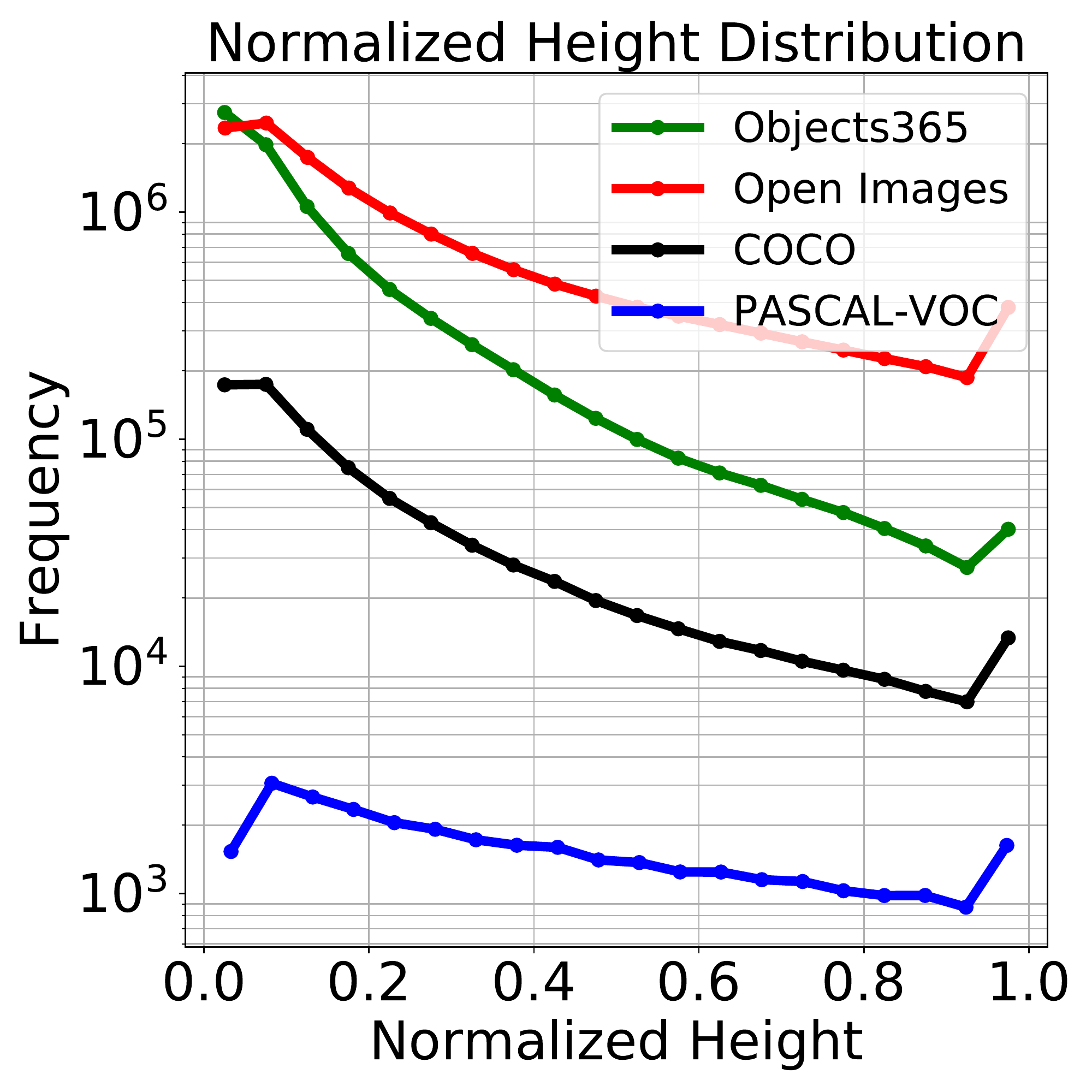}
                \caption{}
        \end{subfigure}
        \begin{subfigure}[b]{0.25\textwidth}
                \includegraphics[width=\textwidth]{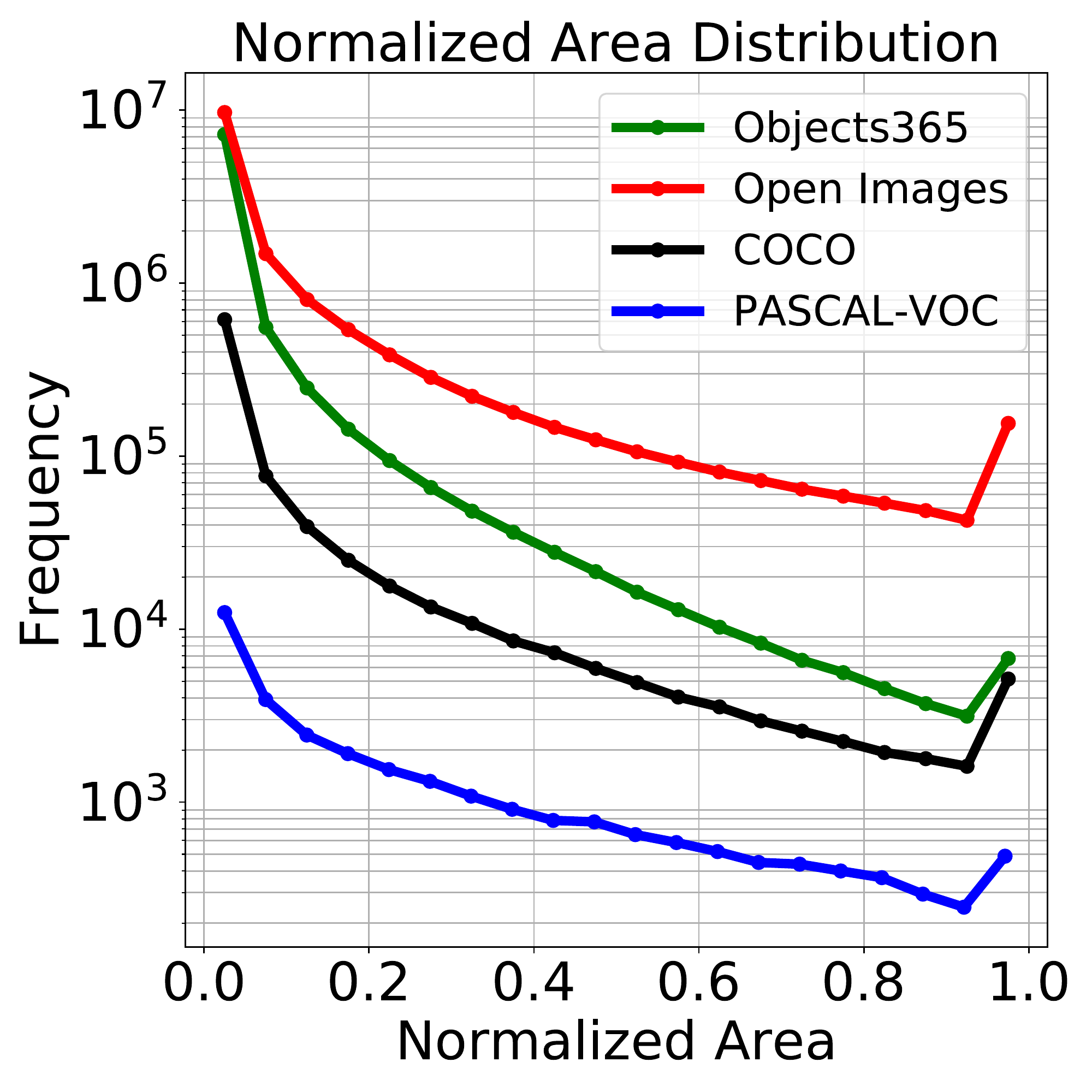}
                \caption{}
        \end{subfigure}   
        \caption{Imbalance in scales of the objects in common datasets: the distributions of BB width \textbf{(a)}, height \textbf{(b)}, and area \textbf{(c)}. Values are with respect to the normalized image (i.e. relative to the image). For readability, the $y$ axes are in log-scale.
        }
        \label{fig:ScaleImb}
\end{figure*}
\begin{figure}
\centering
\includegraphics[width=0.48\textwidth]{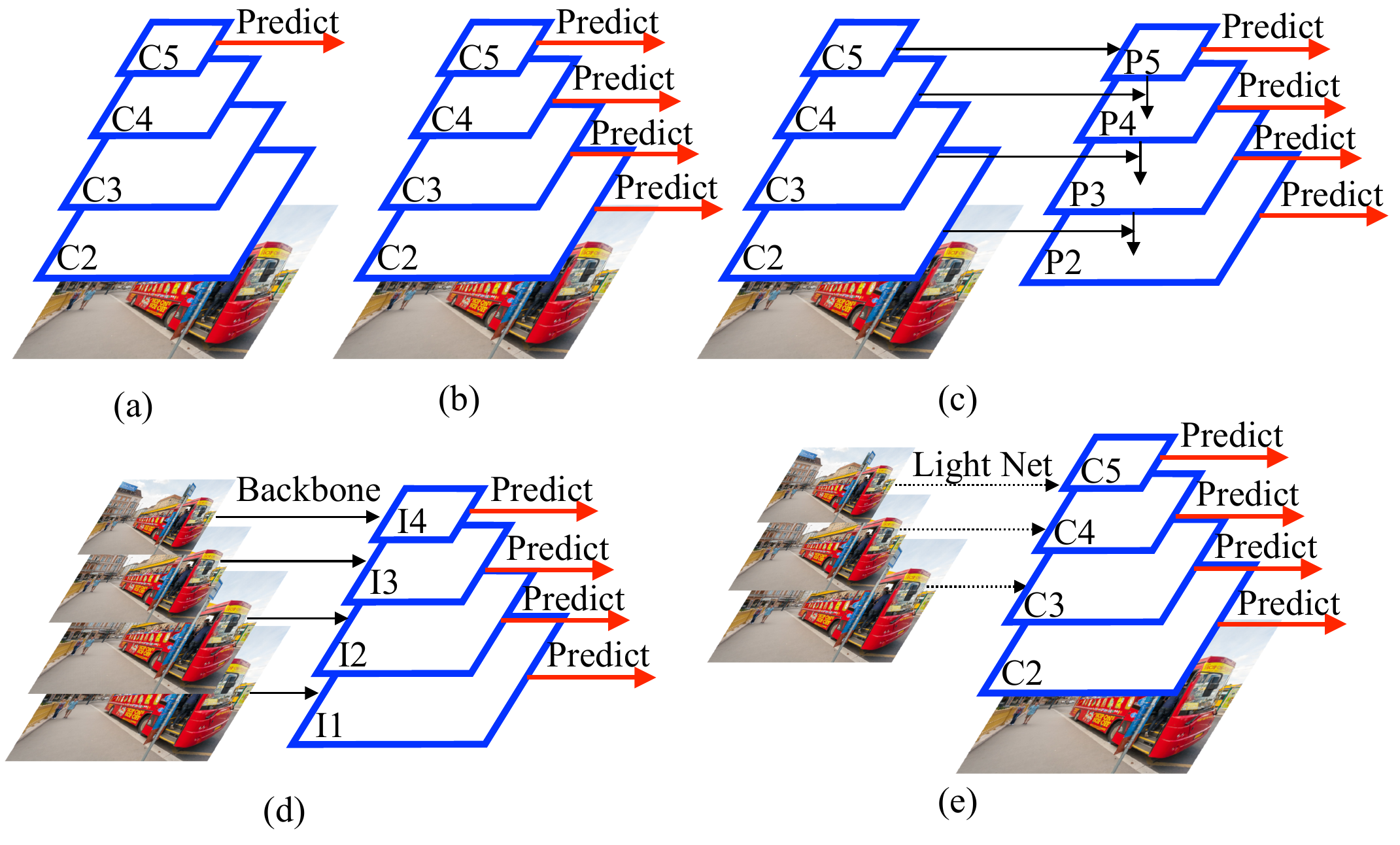} 
\caption{An illustration and comparison of the solutions for scale imbalance. ``Predict'' refers to the prediction performed by a detection network. The layered boxes correspond to convolutional layers. \textbf{(a)} No scale balancing method is employed. \textbf{(b)} Prediction is performed from backbone features at different levels (i.e. scales) (e.g. SSD \cite{SSD}). \textbf{(c)} The intermediate features from different scales are combined before making prediction at multiple scales (e.g. Feature Pyramid Networks \cite{FeaturePyramidNetwork}). \textbf{(d)} The input image is scaled first and then processed. Each I corresponds to an image pyramidal feature (e.g.  Image Pyramids \cite{Snip}). \textbf{(e)} Image and feature pyramids are combined. Rather than applying backbones, light networks are used in order to extract features from smaller images.
}
\label{fig:ScaleImbMethods}
\end{figure}
We discuss scale the imbalance problem in two parts: the first part, Object/Box-Level Scale Imbalance, analyses the problems arising from  the imbalanced distribution of  the scales of the objects and input bounding boxes. The second part,  Feature Imbalance, analyses the problems arising at the feature extraction level and concerns the methods  using pyramidal features.

\subsection{Object/Box-Level Scale Imbalance}
\label{subsec:objboxScale}
\noindent\textit{Definition.} Scale imbalance occurs when certain sizes of the objects or input bounding boxes are over-represented in the dataset. It has been shown that this affects the scales of the estimated RoIs and the overall detection performance \cite{Snip}. Figure \ref{fig:ScaleImb} presents the relative width, height and the area of the objects in the MS-COCO dataset \cite{COCO}; we observe a skewness in the distributions in favor of smaller objects.

Many of the deep object detectors rely on a backbone convolutional neural network (e.g.  \cite{VGG,GoogLeNet,ResNet,HourglassNet,MobileNet}), pretrained on an image classification task, in order to extract visual features from the input image. Li et al. \cite{DetNet} discuss the cons of employing such networks   designed for the classification task and propose a backbone specially designed for the object detection task where they limit the spatial downsampling rate for the high level features. Overall, these networks, also called the backbones, play an important role for the performance of the object detectors, but they alone are insufficient for handling the scale diversity of the input bounding boxes. 

\noindent\textit{Solutions.} First examples of deep object detectors made predictions from the final layer of the backbone network (see \cite{RCNN,FastRCNN} and Figure \ref{fig:ScaleImbMethods}(a)), and therefore, neglected the scale-diversity of BBs. The solutions to addressing scale imbalance can be grouped into four (Figure \ref{fig:ScaleImbMethods}): methods predicting from the hierarchy of the backbone features (Figure \ref{fig:ScaleImbMethods}(b)), methods based on feature pyramids (Figure \ref{fig:ScaleImbMethods}(c)), methods based on image pyramids (Figure \ref{fig:ScaleImbMethods}(d)) and finally methods combining image and feature pyramids (Figure \ref{fig:ScaleImbMethods}(e)). 

\subsubsection{Methods Predicting from the Feature Hierarchy of Backbone Features}
\label{subsec:FeatureHierarchy}

These methods make independent predictions from the features at different levels of the backbone network (Figure \ref{fig:ScaleImbMethods}(b)). This approach naturally considers object detection at multiple scales since the different levels encode information at different scales; e.g., if the input contains a small object, then earlier levels already contain strong indicators about the small object \cite{ScaleDpdPool}. 

An illustratory example for one-stage detectors is the Single Shot Detector (\vurgula{SSD}) \cite{SSD}, which makes predictions from features at different layers.

Two-stage detectors can exploit features at different scales while either estimating the regions (in the first stage) or extracting features from these regions (for the second stage). For example, the \vurgula{Multi Scale CNN} (MSCNN) \cite{MSCNN} uses different layers of the backbone network while estimating the regions in the first stage whereas Yang et al. \cite{ScaleDpdPool} choose an appropriate layer to pool based on the scale of the estimated RoI; called \vurgula{Scale Dependent Pooling} (SDP), the method, e.g., pools features from an earlier layer if the height of the RoI is small. Alternatively, the \vurgula{Scale Aware Fast R-CNN} \cite{ScaleAwarefastRCNN} learns an ensemble of two classifiers, one for the small scale and one for the large scale objects, and combines their predictions.

\subsubsection{Methods Based on Feature Pyramids}
Methods based on feature hierarchies use features from different levels independently without integrating low-level and high-level features. However, the abstractness (semantic content) of information varies among different layers, and thus it is not reliable to make predictions directly from a single layer (especially the lower layers) of the backbone network.


To address this issue, the \vurgula{Feature Pyramid Networks} (FPN) \cite{FeaturePyramidNetwork} combine the features at different scales before making predictions. FPN exploits an additional top-down pathway along which the features from the higher level are supported by the features from a lower level using lateral connections in order to have a balanced mixed of these features (see Figure \ref{fig:ScaleImbMethods}(c)). The top-down pathway involves upsampling to ensure the sizes to be compatible and lateral connections are basically $1 \times 1$ convolutions. Similar to feature hierarchies, a RoI pooling step takes the scale of the RoI into account to choose which level to pool from. These improvements allow the predictor network to be applied at all levels which improves the performance especially for small and medium sized objects. 

Although FPN was originally proposed for object detection, it quickly became popular and has been used for different (but related) tasks such as  shadow detection \cite{ShadowFPN}, instance segmentation \cite{InstanceFPN,LandFPN} and panoptic segmentation \cite{PanopticFPN}. 

Despite its benefits, FPN is known to suffer from a major shortcoming due to the straightforward combination of the features gathered from the backbone network -- the \textit{feature imbalance problem}. We discuss this problem in Section \ref{subsec:FeatImb}. 

\subsubsection{Methods Based on Image Pyramids}
\label{subsec:scalevar}

The idea of using multi-scale image pyramids, presented in Figure \ref{fig:ScaleImbMethods}(d), for the image processing tasks goes back to the early work by Adelson et al. \cite{ImagePyramid1} and was popular before deep learning. In deep learning, these methods are not utilized as much due to their relatively high computational  and memory costs. However, recently, Singh and Davis \cite{Snip} presented a detailed analysis on the effect of the scale imbalance problem with important conclusions and proposed a method to alleviate the memory constraint for image pyramids. They investigated the conventional approach of training object detectors in smaller scales but testing them in larger scales due to memory constraints, and showed that this inconsistency between test and training time scales has an impact on the performance. In their controlled experiments, upsampling the image by two performed better than reducing the stride by two. Upon their analysis, the authors proposed a novel training method coined as \vurgula{SNIP} based on image pyramids rather than feature pyramids. They argue that, while training scale-specific detectors by providing the input to the appropriate detector will lose a significant portion of the data, and that using multi-scale training on a single detector will increase the scale imbalance by preserving the variation in the data. Therefore, SNIP trains multiple proposal and detector networks with images at different sizes, however, for each network only the appropriate input bounding box scales are marked as valid, by which it ensures multi-scale training without any loss in the data. Another challenge, the limitation of the GPU memory, is overcome by an image cropping approach. The image cropping approach is made more efficient in a follow-up method, called \vurgula{SNIPER} \cite{Sniper}.

\subsubsection{Methods Combining Image and Feature Pyramids}
\label{subsubsec:CombineImageFeature}

Image pyramid based methods are generally less efficient than feature pyramid based methods in terms of computational time and memory. However,  image pyramid based methods are expected to perform better, since feature pyramid based methods are efficient approximations of such methods. Therefore, to benefit from advantages of both approaches, they can be combined in a single model. 

Even though there are different architectures, the generic approach is illustrated in Figure \ref{fig:ScaleImbMethods}(e). For example, \vurgula{Efficient Featurized Image Pyramids} \cite{EffFeaturizedImgPyramid} uses five images at different scales, four of which are provided to the light-weight featurized image pyramid network module (i.e. instead of additional backbone networks) and the original input is fed into the backbone network. This light-weight network consists of $4$ consecutive convolutional layers designed specifically for each input. The features gathered from this module are integrated to the backbone network features at appropriate levels according to their sizes, such that the image features are combined with the features extracted from the backbone network using feature attention modules. Furthermore, after attention modules, the gathered features are integrated with the higher levels by means of forward fusion modules before the final predictions are obtained. 

A similar method that is also built on SSD and uses downsampled images is \vurgula{Enriched Feature Guided Refinement Network} \cite{EnrichedFeatureGuided} in which a single downsampled image is provided as an input to the Multi-Scale Contextual Features (MSCF) Module. The MSCF consists of two consecutive convolutional layers followed by three parallel dilated convolutions, which is also similar to the idea in Trident Networks \cite{TridentNet}. The outputs of the dilated convolutions are again combined using $1 \times 1$ convolutions. The authors set the downsampled image size to meet the first prediction layer in the regular SSD architecture (e.g. for $320 \times 320$ input, the size of the downsampled image is $40 \times 40$). While Pang et al. \cite{EffFeaturizedImgPyramid} only provide experiments with the SSD backbone, Nie et al. \cite{EnrichedFeatureGuided} show that their modules can also be used in the ResNet backbone.

An alternative approach is to generate super-resolution feature maps. Noh et al. \cite{BettertoFollow} proposed \vurgula{super-resolution for small object detection} for two-stage object detectors which lack strong representations of small RoIs after RoI standardization layers. Their architecture adds four additional modules to the baseline detector: (i) Given the original image, target extractor outputs the targets for the discriminator by using dilated convolutions. This network also shares parameters with the backbone network. (ii) Given the features obtained from the backbone network using the original image and a $0.5$x downsampled image, the generator network generates super resolution feature maps for small RoIs. (iii) Given the outputs of (i) and (ii), a discriminator is trained in the conventional GAN setting. (iv) Finally, if the RoI is a small object according to a threshold, then the prediction is carried out by the small predictor, or else by the large predictor.

Another approach, \vurgula{Scale Aware Trident Networks} \cite{TridentNet}, combines the advantages of the methods based on feature pyramids and image pyramids without using multiple downsampled images but employing only dilated convolutions. The authors use dilated convolutions \cite{DilatedConvolution} with dilation rates $1, 2$ and $3$ in parallel branches in order to generate scale-specific feature maps, making the approach more accurate compared to feature pyramid based methods. In order to ensure that each branch is specialized for a specific scale, an input bounding box is provided to the appropriate branch according to its size. Their analysis on the effect of receptive field size on objects of different scales shows that larger dilation rates are more appropriate for objects with larger scales. In addition, since using multiple branches is expected to degrade the efficiency due to the increasing number of operations, they proposed a method for approximating these branches with a single parameter-sharing branch, with minimal (insignificant) performance loss.

\subsection{Feature-level Imbalance}
\label{subsec:FeatImb}
\label{subsec:FPNimb}

\begin{figure}
    \centering
    \includegraphics[width=0.35\textwidth]{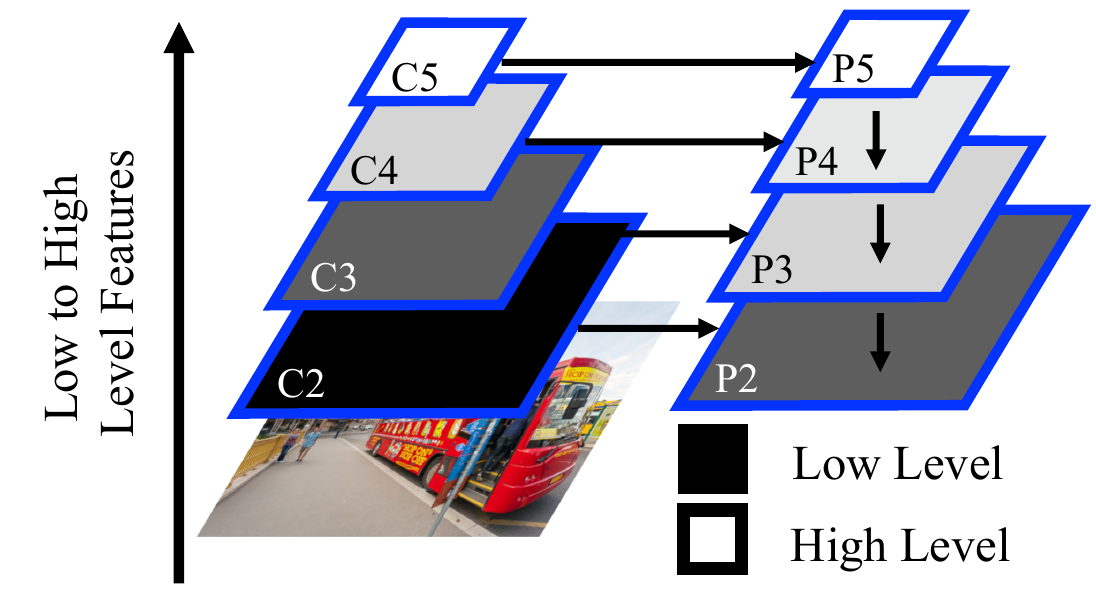}
    \caption{Feature-Level imbalance is illustrated on the FPN architecture. \label{fig:FeatImb}}
\end{figure}

\noindent\textit{Definition.} The integration of the features from the backbone network is expected to be balanced in terms of low- and high-level features so that consistent predictions can follow. To be more specific, if we analyze the conventional FPN architecture in Figure \ref{fig:FeatImb}, we notice that, while there are several layers from the $C2$ layer of the bottom-up pass with low-level features to the $P5$ layer of the feature pyramid, the $C2$ layer is directly integrated to the $P2$ layer, which implies the effect of the high-level and low-level features in the $P2$ and $P5$ layers to be different.

\noindent\textit{Solutions.} There are several methods to address imbalance in the FPN architectures, which range from designing improved top-down pathway connections \cite{RON,AnchorRefine} to completely novel architectures. Here, we consider the methods to alleviate the feature-level imbalance problem using novel architectures, which we group into two according to what they use as a basis,  pyramidal or backbone features. 

\begin{figure*}
\centering
\includegraphics[width=0.83\textwidth]{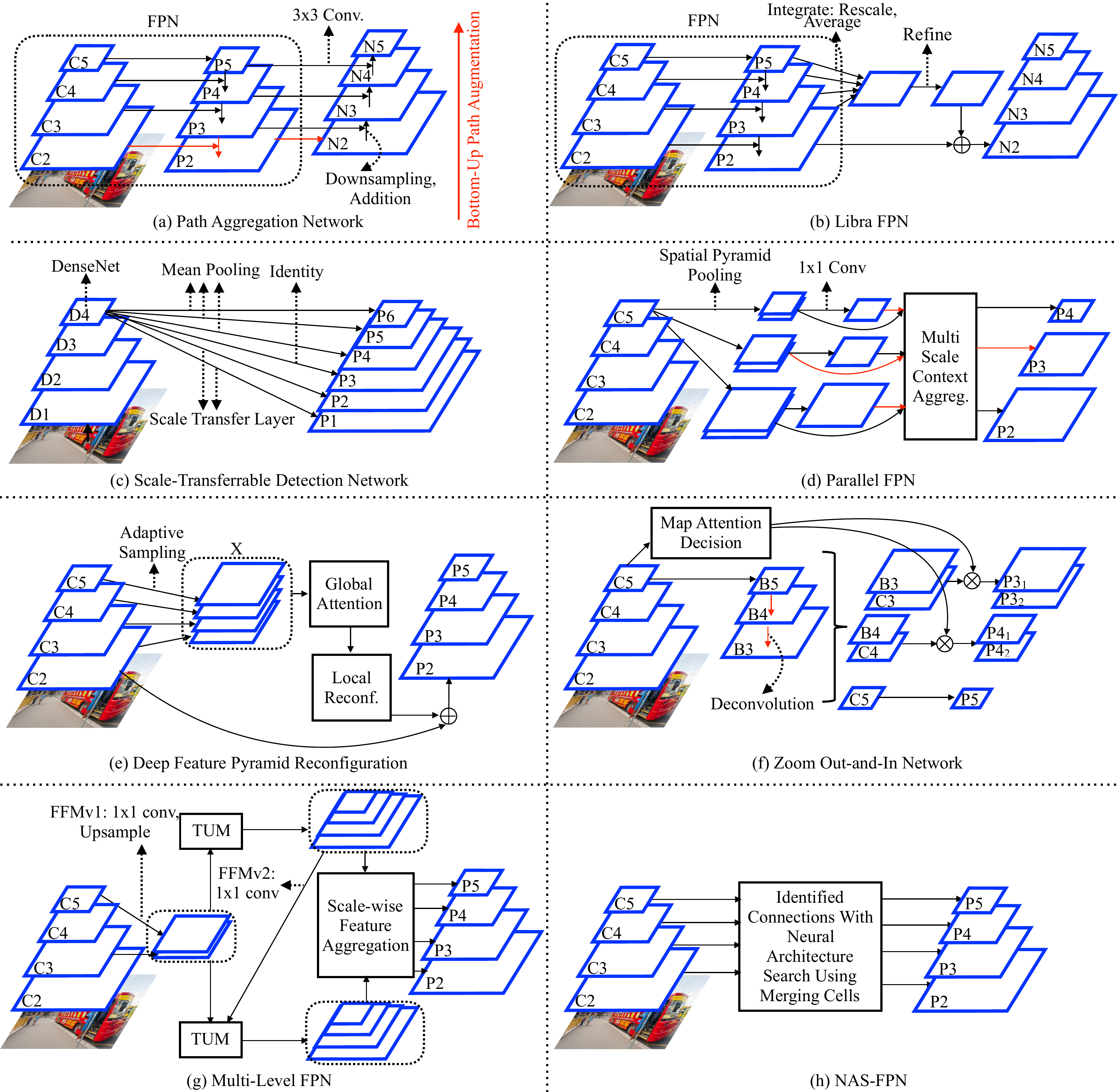} 
\caption{High-level diagrams of the methods designed for feature-level imbalance. \textbf{(a)} \textit{Path Aggregation Network}. FPN is augmented by an additional bottom-up pathway to facilitate a shortcut of the low-level features to the final pyramidal feature maps. Red arrows represent the shortcuts. \textbf{(b)} \textit{Libra FPN}. FPN pyramidal features are integrated and refined to learn a residual feature map. We illustrate the process originating from $P2$ feature map. Remaining feature maps ($P3$-$P5$) follow the same operations. \textbf{(c)}\textit{ Scale Transferrable Detection Network}. Pyramidal features are learned via pooling, identity mapping and scale transfer layers depending on the layer size. \textbf{(d)} \textit{Parallel FPN}. Feature maps with difference scales are followed by spatial pyramid pooling. These feature maps are fed into the multi-scale context aggregation (MSCA) module. We show the input and outputs of MSCA module for $P3$ by red arrows. \textbf{(e)} \textit{Deep Feature Pyramid Reconfiguration}. A set of residual features are learned via global attention and local reconfiguration modules. We illustrate the process originating from $P2$ feature map. Remaining feature maps ($P3$-$P5$) follow the same operations. \textbf{(f)} \textit{Zoom Out-And-In Network}. A zoom-in phase based on deconvolution (shown with red arrows) is adopted before stacking the layers of zoom-out and zoom-in phases. The weighting between them is determined by map attention decision module. \textbf{(g)} \textit{Multi-Level FPN}. Backbone features from two different levels are fed into Thinned U-Shape Module (TUM) recursively to generate a sequence of pyramidal features, which are finally combined into one by acale-wise feature aggregation module. \textbf{(h)} NAS-FPN. The layers between backbone features and pyramidal features are learned via Neural Architecture Search.}
\label{fig:FPNNetworks}
\end{figure*}

\subsubsection{Methods Using Pyramidal Features as a Basis}
\label{subsubsec:FPN}

These methods aim to improve the pyramidal features gathered by FPN using additional operations or steps -- see an overview of these methods in Figure \ref{fig:FPNNetworks}(a,b).

\vurgula{Path Aggregation Network} (PANet) \cite{PANet} is the first to show that the features gathered by an FPN can be further enhanced and an RoI can be mapped to each layer of the pyramid rather than associating it with a single one. The authors suggest that low-level features, such as edges, corners, are useful for localizing objects, however, the FPN architecture does not sufficiently make use of these features. Motivated from this observation, PANet, depicted in Figure \ref{fig:FPNNetworks}(a), improves the FPN architecture with two new contributions:
\begin{enumerate}
    \item \textit{Bottom-up path augmentation} extends the feature pyramid in order to allow the low-level features to arrive at the layers where the predictions occur in shorter steps (see red arrows in Figure \ref{fig:FPNNetworks}(a) within FPN and to the final pyramidal features to see the shortcut). For this reason, in a way a shortcut is created for the features in the initial layers. This is important since these features have rich information about localization thanks to edges or instance parts. 
    \item While in the FPN, each RoI is associated with a single level of feature based on its size, PANet associates each RoI to every level, applies RoI Pooling, fuses using element-wise max or sum operation and the resulting fixed-sized feature grid is propagated to the detector network. This process is called \textit{Adaptive Feature Pooling}.
\end{enumerate}
Despite these contributions, PANet still uses a sequential pathway to extract the features. 

Different from the sequential enhancement pathway of PANet, \vurgula{Libra FPN} \cite{LibraRCNN} aims to learn the residual features by using all of the features from all FPN layers at once (see Figure \ref{fig:FPNNetworks}(b)). Residual feature layer computation is handled in two steps:
\begin{enumerate}
    \item \textit{Integrate:} All feature maps from different layers are reduced to one single feature map by rescaling and averaging. For this reason, this step does not have any learnable parameter.
    \item \textit{Refine:} The integrated feature map is refined by means of convolution layers or non-local neural networks \cite{NonLocalNN}.
\end{enumerate}
Finally the refined features are added to each layer of the pyramidal features. The authors argue that in addition to FPN, their method is complementary to other methods based on pyramidal features as well, such as PANet \cite{PANet}.

\subsubsection{Methods Using Backbone Features as a Basis}

These methods build their architecture on the backbone features and ignore the top-down pathway of FPN by employing different feature integration mechanisms, as displayed in Figure \ref{fig:FPNNetworks}(c-h).

\vurgula{Scale-Transferrable Detection Network} (STDN) \cite{ScaleFPN} generates the pyramidal features from the last layer of the backbone features which are extracted using DenseNet \cite{DenseNet} blocks (Figure \ref{fig:FPNNetworks}(c)). In a DenseNet block, all the lower level features are propagated to every next layer within a block. In Figure \ref{fig:FPNNetworks}(c), the number of DenseNet (dense) blocks is four and the $i$th block is denoted by $D_i$. Motivated by the idea that direct propagation of lower-level layers to the subsequent layers also carries lower-level information, STDN  builds pyramidal features consisting of six layers by using the last block of DenseNet. In order to map these layers to lower sizes, the approach uses mean pooling with different receptive field sizes. For the fourth feature map, an identity mapping is used. For the last two layers which the feature maps of DenseNet are to be mapped to higher dimensions, the authors propose a \textit{scale transfer layer} approach. This layer does not have any learnable parameter and given $r$, the desired enlargement for a feature map, the width and height of the feature map are enlarged by $r$ by decreasing the total number of feature maps (a.k.a. channels). STDN incorporates high- and low-level features with the help of DenseNet blocks and is not easily adaptable to other backbone networks. In addition, no method is adopted to balance the low- and high-level features within the last block of DenseNet. 

Similar to STDN, \vurgula{Parallel FPN} \cite{ParallelFPN} also employs only the last layer of the backbone network and generates multi-scale features by exploiting the spatial pyramid pooling (SPP) \cite{SPPNet} -- Figure \ref{fig:FPNNetworks}(d). Differently, it increases the width of the network by pooling the last $D$ feature maps of the backbone network multiple times with different sizes, such that feature maps with different scales are obtained. Figure \ref{fig:FPNNetworks}(e) shows the case when it is pooled for three times and $D=2$. The number of feature maps is decreased to $1$ by employing $1 \times 1$ convolutions. These feature maps are then fed into the \textit{multi-scale context aggregation (MSCA) module}, which integrates context information from other scales for a corresponding layer. For this reason, MSCA, operating on a scale-based manner, has the following inputs: Spatial pyramid pooled $D$ feature maps and reduced feature maps from other scales. We illustrate the inputs and outputs to the MSCA module for the middle scale feature map by red arrows in Figure \ref{fig:FPNNetworks}(e). MSCA first ensures the sizes of each feature map to be equal and applies $3 \times 3$ convolutions. 

While previous methods based on backbone features only use the last layer of the backbone network, \vurgula{Deep Feature Pyramid Reconfiguration} \cite{PyramidReconfiguration} combines features from different levels of backbone features into a single tensor ($X$ in Figure \ref{fig:FPNNetworks}(e)) and then learns a set of residual features from this tensor. A sequence of two modules are applied to the tensor $X$ in order to learn a residual feature map to be added to each layer of the backbone network. These modules are,
\begin{enumerate}
    \item \textit{Global Attention Module} aims to learn the inter-dependencies among different feature maps for tensor $X$. The authors adopt Squeeze and Excitation Blocks \cite{squeezeExcitation} in which the information is squeezed to lower dimensional features for each feature map initially (i.e. squeeze step), and then a weight is learned for each feature map based on learnable functions including non-linearity (i.e. excitation step).
    \item \textit{Local Configuration Module} aims to improve the features after global attention module by employing convolutional layers. The output of this module presents the residual features to be added for a feature layer from the backbone network.
\end{enumerate}

Similarly, \vurgula{Zoom Out-and-In Network} \cite{ZoomOutIn} also combines low- and high-level features of the backbone network. Additionally, it includes deconvolution-based zoom-in phase in which intermediate step pyramidal features, denoted by $Bi$s in Figure \ref{fig:FPNNetworks}(f), are learned. Note that, unlike FPN \cite{FeaturePyramidNetwork}, there is no lateral connection to the backbone network during the zoom-in phase, which is basically a sequence of deconvolutional layers (see red arrows for the zoom-in phase). Integration of the high- and low-level features are achieved by stacking the same-size feature maps from zoom-out and zoom-in phases after zoom-in phase (i.e. $B3$ and $C3$). On the other hand, these concatenated feature maps are to be balanced especially for the $B3-C3$ and $B4-C4$ blocks since $B3$ and $C3$ (or $B4$ and $C4$) are very far from each other in the feature hierarchy, which makes them have different representations of the data. In order to achieve this, the proposed \textit{map attention decision module} learns a weight distribution on the layers. Note that the idea is similar to the squeeze and excitation modules \cite{squeezeExcitation} employed by Kong et al. \cite{PyramidReconfiguration}, however, it is shown by the authors that their design performs better for their architecture. One drawback of the method is that it is built upon Inception v2 (a.k.a. Inception BN) \cite{inceptionv2} and corresponding inception modules are exploited throughout the method, which may make the method difficult to adopt for other backbone networks.

Different from Kong et al. \cite{PyramidReconfiguration} and Li et al. \cite{ZoomOutIn}, \vurgula{Multi-Level FPN} \cite{MultiLevelFPN} stacks one highest and one lower level feature layers and recursively outputs a set of pyramidal features, which are all finally combined into a single feature pyramid in a scale-wise manner (see Figure \ref{fig:FPNNetworks}(g)). \textit{Feature fusion module (FFM) v1} equals the dimensions of the input feature maps by a sequence of $1 \times 1$ convolution and upsampling operations. Then, the resulting two-layer features are propagated to \textit{thinned U-shape modules (TUM)}. Excluding the initial propagation, each time these two-layer features are integrated to the output of the previous TUM by $1 \times 1$ convolutions in \textit{FFMv2}. Note that the depth of the network is increasing after each application of the TUM and the features are becoming more high level. As a result of this, a similar problem with the FPN feature imbalance arise again. As in the work proposed by \cite{PyramidReconfiguration}, the authors employed squeeze and excitation networks \cite{squeezeExcitation} to combine different pyramidal shape features. 

Rather than using hand-crafted architectures, \vurgula{Neural Architecture Search FPN} (NAS-FPN) \cite{NASFPN} aims to search for the best architecture to generate pyramidal features given the backbone features by using neural architecture search methods \cite{NASRL} -- Figure \ref{fig:FPNNetworks}(h). This idea was also previously applied to the image classification task and showed to perform well \cite{NASImgClassification,NASImgClassification2,NASImgClassification3}. \vurgula{Auto-FPN} is also another example for using NAS while learning the connections from backbone features to pyramidal features and beyond. While NAS-FPN achieves higher performance, Auto-FPN is more efficient and has less memory footprint. The idea is also applied to backbone design for object detection by Chen et al. \cite{DetNAS}, however it is not within the scope of our paper. Considering their performance in other tasks such as EfficientNet \cite{NASImgClassification3} and different definitions of search spaces may lead to better performance in NAS methods, more work is expected for FPN design using NAS.

\subsection{Comparative Summary}
SSD \cite{SSD} provides an analysis on the importance of making predictions from different number of layers with varying scales. Increasing the number of prediction layers leads to a significant performance gain. While predicting from one layer provides $62.4$ mAP@0.5 on Pascal VOC 2007 test set \cite{PASCAL} with SSD-300, it is $70.7$ and $74.6$ when the number of prediction layers is $3$ and $5$ respectively (while keeping the number of predictions almost equal). Therefore, once addressed in the detection pipeline, scale imbalance methods can significantly boost performance.

Feature pyramid based methods increased the performance significantly compared to SSD. Once included in Faster R-CNN with ResNet-101, the pioneering FPN study \cite{FeaturePyramidNetwork} has a relative improvement of $3.7 \%$ on MS COCO testdev (from $34.9$ to $36.2$). Another virtue of the feature pyramids is the efficiency due to having lighter detection networks: e.g., the model with FPN is reported to be more than two times faster than baseline Faster R-CNN with ResNet-50 backbone ($150$ms vs $320$ms per image on a single NVIDIA M40 GPU) \cite{FeaturePyramidNetwork}. Currently, the most promising results are obtained via NAS by learning how to combine the backbone features from different levels. Using the ResNet-50 backbone with $1024 \times 1024$ image size, a $10.2 \%$ relative improvement is obtained on MS COCO testdev (from $40.1$ to $44.2$) by NAS-FPN \cite{NASFPN} but it needs also to be noted that number of parameters in the learned structure is approximately two times higher and inference time is $26 \%$ more ($92.1$ms vs $73.0$ms per image on a single P100 GPU). Inference time is also a major drawback of the image pyramid based methods. While SNIP reports inference at approximately $1$ image/sec, the faster version, SNIPER, improves it to $5$ images/sec on a V100 GPU ($200$ ms/image).

In order to adjust the inference time-performance trade-off in scale imbalance, a common method is to use multi-scale images with one backbone and multiple lighter networks. All methods in Section \ref{subsubsec:CombineImageFeature} are published last year. To illustrate the performance gains: Pang et al. \cite{EffFeaturizedImgPyramid} achieved $19.5 \%$ on MS COCO test-dev relative improvement with $12$ms per image inference time compared to SSD300 with the same size image and backbone network having $10$ms per image inference time (mAPs are $25.1$ vs $30.0$).

\subsection{Open Issues}
\label{subsec:OpenProblemsScImb}

Here we discuss open problems concerning scale imbalance.

\subsubsection{Characteristics of Different Layers of Feature Hierarchies}

In feature-pyramid based methods (Section \ref{subsec:FPNimb}), a prominent and common pattern is to include a top-down pathway in order to integrate higher-layer features with lower-layer ones. Although this approach has yielded promising improvements in performance, an established perspective about what critical aspects of features (or information) are handled differently in those methods is missing. Here, we highlight three such aspects:

{\noindent}\textbf{(i) Abstractness}. Higher layers in a feature hierarchy carry high-level, semantically more meaningful information about the objects or object parts whereas the lower layers represent low-level information in the scene, such as edges, contours, corners etc. In other words, higher-layer features are more abstract.
    
{\noindent}\textbf{(ii) Coarseness}. To reduce dimensions, feature networks gradually reduce the size of the layers towards the top of the hierarchy. Although this is reasonable given the constraints, it has an immediate outcome on the number of neurons that a fixed bounding box at the image level encapsulates at the different feature layers. Namely, the BB will include less neurons when projected to the highest layer. In other words, higher layers are more coarse.
    
{\noindent}\textbf{(iii) Cardinality}. In FPN and many of its variants, prediction is performed for an object from the layer that matches the object's scale. Since the scales of objects are not balanced, this approach has a direct affect on the number of predictions and backpropagation performed through a layer.

We argue that analyzing and addressing these aspects in a more established manner is critical for developing more profound solutions. Although we see that some methods handle imbalance in these aspects (e.g. Libra FPN \cite{LibraRCNN} addresses all three aspects, Path Aggregation Network \cite{PANet} handles abstractness and cardinality, whereas FPN solves only abstractness to a certain extent), these aspects should be quantified and used for comparing different methods.



\subsubsection{Image Pyramids in Deep Object Detectors}
{\noindent}\textit{Open Issue.} It is hard to exploit image pyramids using neural networks due to memory limitations. Therefore, finding solutions alleviating this constraint (e.g., as in SNIP \cite{Snip}) is still an open problem.

{\noindent}\textit{Explanation.} The image pyramids (Figure \ref{fig:ScaleImbMethods}(d)) were commonly adopted by the pre-deep learning era object detectors. However, memory limitations motivated the methods based on pyramidal features which, with less memory, are still able to generate a set of features with different scales allowing predictions to occur at multiple scales. On the other hand, feature-pyramids are actually approximations of the features extracted from image pyramids, and there is still room for improvement given that using image pyramids is not common among deep object detectors.



\section{Imbalance 3: Spatial Imbalance}
\label{sec:BBimb}
\label{sec:spatialImb}

{\noindent}\textit{Definition.} Size, shape, location -- relative to both  the image or  another box -- and IoU are spatial attributes of bounding boxes. Any imbalance in such attributes is likely to affect the training and generalization performance. For example, a slight shift in position may lead to drastic changes in the regression (localization) loss, causing an imbalance in the loss values, if a suitable loss function is not adopted. In this section, we discuss these problems specific to the spatial attributes and regression loss.

\subsection{Imbalance in Regression Loss}
\label{subsec:Regression}

\begin{figure}
\centering
\includegraphics[width=0.40\textwidth]{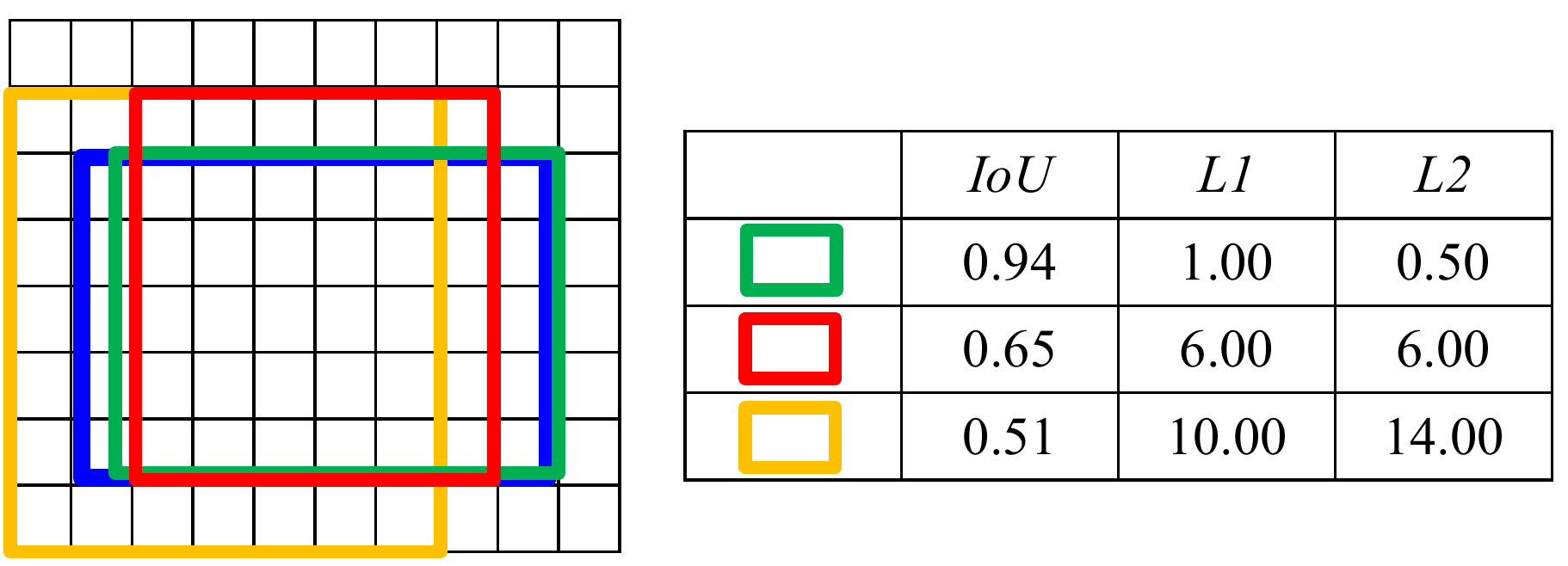}
\caption{An illustration of imbalance in regression loss. Blue denotes the ground truth BB. There are three prediction boxes, marked with green, red and yellow colors. In the table on the right, $L1$ and $L2$ columns show the sum of $L1$ and $L2$ errors between the box corners of the associated prediction box and the ground-truth (blue) box. Note that the contribution of the yellow box to the $L2$ loss is more dominating than its effect on total $L1$ error. Also, the contribution of the green box is less for the $L2$ error.}
\label{fig:ObjImbalanceExLevel}
\end{figure}


\noindent\textit{Definition.} This imbalance problem is concerned with the uneven contributions of different individual examples to the regression loss. Figure \ref{fig:ObjImbalanceExLevel} illustrates the problem using $L1$ and $L2$ losses, where the hard example (i.e. the one with low IoU, the yellow box) is dominating the $L2$ loss, whereas $L1$ loss assigns relatively more balanced errors to all examples.  

\noindent\textit{Solutions.} The regression losses for object detection have evolved under two main streams: The first one is the $Lp$-norm-based (e.g. $L1$, $L2$) loss functions and the second one is the IoU-based loss functions. Table \ref{tab:RegressionLosses} presents a comparison of the widely used regression losses.

\begin{table*}
\caption{A list of widely used loss functions for the BB regression task.}
    \centering    
    \begin{tabular}{|p{5cm}|p{11cm}|}
         \hline
         \multicolumn{1}{|c|}{\textbf{Loss Function}} & \multicolumn{1}{|c|}{\textbf{Explanation}}  \\ \hline
         $L2$ Loss &Employed in earlier deep object detectors \cite{RCNN}. Stable for small errors but penalizes outliers heavily. \\ \hline 
         $L1$ Loss &Not stable for small errors.\\ \hline 
         Smooth $L1$ Loss \cite{FastRCNN}&Baseline regression loss function. More robust to outliers compared to $L1$ Loss. \\ \hline 
         Balanced $L1$ Loss \cite{LibraRCNN} &Increases the contribution of the inliers compared to smooth $L1$ loss.  \\ \hline
         Kullback-Leibler Loss \cite{KLLoss} & Predicts a confidence about the input bounding box based on KL divergence.\\ \hline
         IoU Loss \cite{UnitBox} & Uses an indirect calculation of IoU as the loss function. \\ \hline 
         Bounded IoU Loss \cite{BoundedIoU}& Fixes all parameters of an input box in the IoU definition except the one whose gradient is estimated during backpropagation. \\ \hline 
         GIoU Loss \cite{GIoULoss}& Extends the definition of IoU based on the smallest enclosing rectangle of the inputs to the IoU, then uses directly IoU and the extended IoU, called GIoU, as the loss function. \\ \hline 
         DIoU Loss, CIoU Loss \cite{DIoULoss}& Extends the definition of IoU by adding additional penalty terms concerning aspect ratio difference and center distances between two boxes. \\ \hline          
    \end{tabular}
    \label{tab:RegressionLosses}
\end{table*}

Replacing $L2$ regression loss \cite{Overfeat, FastRCNN}, \vurgula{Smooth $L1$ Loss} \cite{FastRCNN} is the first loss function designed specifically for deep object detectors, and it has been widely adopted (e.g. \cite{SSD,FasterRCNN, FocalLoss}) since it reduces the effect of the outliers (compared to $L2$ loss) and it is more stable for small errors (compared to $L1$ loss). Smooth $L1$ loss, a special case of Huber Loss \cite{HuberLoss}, is defined as:
%
\begin{align}
  \label{eq:smoothl1}
    L1_{smooth}(\hat{x}) = 
    \begin{cases}
       0.5 \hat{x}^2, &\quad\text{if } |\hat{x}| < 1 \\
       |\hat{x}|-0.5, &\quad\text{otherwise.} \\
    \end{cases}
\end{align}
%
where $\hat{x}$ is the difference between the estimated and the target BB coordinates. 

Motivated by the fact that the gradients of the outliers still have a negative effect on learning the inliers with smaller gradients in Smooth $L1$ loss, \vurgula{Balanced $L1$ Loss} \cite{LibraRCNN} increases the gradient contribution of the inliers to the total loss value. To achieve this, the authors first derive the definition of the loss function originating from the desirable balanced gradients across inliers and outliers:
\begin{align}
   \label{eq:balancedl1grad} 
    \frac{ \partial L1_{balanced}} {\partial \hat{x}}= \begin{cases} \alpha \ln(b|\hat{x}|+1), &\quad \text{if } 
    |\hat{x}|<1 \\
    \theta, &\quad \text{otherwise,}
    \end{cases}
\end{align}
where $\alpha$ controls how much the inliers are promoted (small $\alpha$ increases the contribution of inliers); $\theta$ is the upper bound of the error to help balancing between the tasks. Integrating Equation \eqref{eq:balancedl1grad}, $L1_{balanced}$ is derived as follows:
\small
\begin{align}
    \label{eq:balancedl1def}
    L1_{balanced} (\hat{x})= \begin{cases} \frac{\alpha}{b} (b|\hat{x}|+1) \ln (b |\hat{x}|+1) - \alpha |\hat{x}|, & \text{if } |\hat{x}|<1 \\
    \gamma |\hat{x}|+Z, & \text{ otherwise,} \end{cases}
\end{align}
\normalsize
where $b$ ensures $L1_{balanced}(\hat{x}=1)$ is a continuous function, $Z$ is a constant and the association between the hyper-paramaters is:
\begin{align}
    \label{eq:balancedl1ass}
    \alpha \ln(b+1)=\gamma .
\end{align}
Having put more emphasis on inliers, Balanced $L1$ Loss improves the performance  especially for larger IoUs (namely, $AP@0.75$ improves by $\%1.1$).

Another approach, \vurgula{Kullback-Leibler Loss} (KL Loss) \cite{KLLoss}, is driven by the fact that the ground truth boxes can be ambiguous in some cases due to e.g. occlusion, shape of the object or inaccurate labeling. For this reason, the authors aim to predict a probability distribution for each BB coordinate rather than direct BB prediction. The idea is similar to the networks with an additional localization confidence associated prediction branch \cite{IoUNet,PreciseDetection,IoUUniform,GaussianYolo,LearningtoRank}, besides classification and regression, in order to use the predicted confidence during inference. Differently, KL Loss, even without its proposed NMS, has an improvement in localization compared to the baseline. The method assumes that each box coordinate is independent and follows a Gaussian distribution with mean $\hat{x}$ and standard deviation $\sigma$. Therefore, in addition to conventional boxes, a branch is added to the network to predict the standard deviation, that is $\sigma$, and the loss is backpropagated using the KL divergence between the prediction and the ground truth such that the ground truth boxes are modelled by the dirac delta distribution centered at the box coordinates. With these assumptions, KL Loss is proportional to:
\begin{align}
    \label{eq:KLLoss}
    L_{KL}(\hat{x}, \sigma) \propto \frac{\hat{x}^2}{2 \sigma^2}+ \frac{1}{2} \log \sigma^2.
\end{align}
They also employ gradient clipping similar to smooth $L1$ in order to decrease the effect of the outliers. During NMS, a voting scheme is also proposed to combine bounding boxes with different probability distributions based on the certainty of each box; however, this method is out of the scope of our paper. Note that the choice of probability distribution for bounding boxes matters since the loss definition is affected by this choice. For example, Equation \eqref{eq:KLLoss} degenerates to Euclidean distance when $\sigma=1$. 

In addition to the $Lp$-norm-based loss functions, there are also IoU based loss functions which exploit the differentiable nature of IoU. An earlier example is the \vurgula{IoU Loss} \cite{UnitBox}, where an object detector is successfully trained  by directly formulating a loss based on the IoU as:
\begin{equation}
    L_{IoU} = -\ln(IoU).
\end{equation}
Another approach to exploit the metric-nature of $1-IoU$ is the \vurgula{Bounded IoU} loss \cite{BoundedIoU}. This loss warps a modified version of $1-IoU$ to the smooth $L1$ function. The modification involves bounding the IoU by fixing all the parameters except the one to be computed, which implies the computation of the maximum attainable IoU for one parameter:
\begin{align}
    \label{eq:BoundedIoULoss}    
    L_{BIoU}(x,\hat{x})=2L1_{smooth}(1-IoU_B(x,\hat{x})),
\end{align}
where the bounding boxes are represented by center coordinates, width and height as $[c_x,c_y,w,h]$. Here, we define the bounded IoU only for $c_x$ and $w$ since $c_y$ and $h$ have similar definitions. We follow our notation in Section \ref{sec:DefNot} to denote ground truth and detection (i.e. $c_x$ for ground truth and $\bar{c}_x$ for detection). With this notation, $IoU_B$, the bounded IoU, is defined as follows:
\begin{align}
    \label{eq:BIoUFun}    
    IoU_B(c_x,\bar{c}_x)&=\max \left( 0, \frac{w-2|\bar{c}_x-c_x|}{w+2|\bar{c}_x-c_x|} \right), \\ 
    IoU_B(w,\bar{w})&=\min \left( \frac{\bar{w}}{w}, \frac{w}{\bar{w}} \right).
\end{align}
In such a setting, $IoU_B \geq IoU$. Also, an IoU based loss function is warped into the smooth $L1$ function in order to make the ranges of the classification and localization task consistent and to decrease the effect of outliers.

Motivated by the idea that the best loss function is the performance metric itself, in \vurgula{Generalized Intersection over Union} (GIoU) \cite{GIoULoss} showed that IoU can be directly optimized and that IoU and the proposed GIoU can be used as a loss function. GIoU is proposed as both a performance measure and a loss function while amending the major drawback of the IoU (i.e. the plateau when IoU=0) by incorporating an additional smallest enclosing box $E$. In such a way, even when two boxes do not overlap, a GIoU value can be assigned to them and this allows the function to have non-zero gradient throughout the entire input domain rather being limited to IoU>0. Unlike IoU, the $\mathrm{GIoU}(B, \bar{B}) \in [-1, 1]$. Having computed $E$, GIoU is defined as:
\begin{align}
    \label{eq:GIoU}    
    L_{GIoU} (B,\Bar{B})=IoU(B, \Bar{B})-\frac{A(E)-A(B \cup \Bar{B})}{A(E)},
\end{align}
where GIoU is a lower bound for IoU, and it converges to IoU when $A(B \cup \Bar{B})=A(E)$. GIoU preserves the advantages of IoU, and makes it differentiable when IoU=0. On the other hand, since positive labeled BBs have IoU larger than 0.5 by definition, this portion of the function is never visited in practice, yet still, GIoU Loss performs better than using IoU directly as a loss function.

A different idea proposed by Zheng et al. \cite{DIoULoss} is to add penalty terms to the conventional IoU error (i.e. $1-IoU(B, \Bar{B})$) in order to ensure faster and more accurate convergence. To achieve that, in Distance IoU (DIoU) Loss, a penalty term related with the distances of the centers of $B$ and $\Bar{B}$ is added as:
\begin{align}
    L_{DIoU} (B, \Bar{B})=1-IoU(B, \Bar{B})+\frac{d^2(B_C, \Bar{B}_C)}{E_D^2},
\end{align}
where $d()$ is the Euclidean distance, $B_C$ is the center point of box $B$ and $E_D$ is the diagonal length of $E$ (i.e. smallest enclosing box). To further enhance their method, $L_{DIoU}$ is extended with an additional penalty term for inconsistency in aspect ratios of two boxes. The resulting loss function, coined as Complete IoU (CIoU), is defined as:
\begin{align}
    L_{CIoU} (B, \Bar{B})=1-IoU(B, \Bar{B})+\frac{d^2(B_C, \Bar{B}_C)}{E_D^2}+ \alpha v,
\end{align}{}
such that
\begin{align}
    v= \frac{4}{\pi ^ 2} \left( \arctan(\frac{w}{h})-\arctan(\frac{\Bar{w}}{\Bar{h}}) \right)^2,
\end{align}
and
\begin{align}
    \alpha = \frac{v}{(1-IoU(B, \Bar{B}))+v}.
\end{align}
In this formulation, $\alpha$, the trade-off parameter, ensures the importance of the cases with smaller IoUs to be higher. In the paper, one interesting approach is to validate faster and more accurate convergence by designing simulation scenarios since it is not straightforward to analyze IoU-based loss functions (see Section \ref{subsec:OpenProblemsIoUImb}).

\subsection{IoU Distribution Imbalance}
\label{subsec:IoUImb}

\begin{figure*}
        \captionsetup[subfigure]{}
        \centering
        \begin{subfigure}[b]{0.32\textwidth}
                \includegraphics[width=\textwidth]{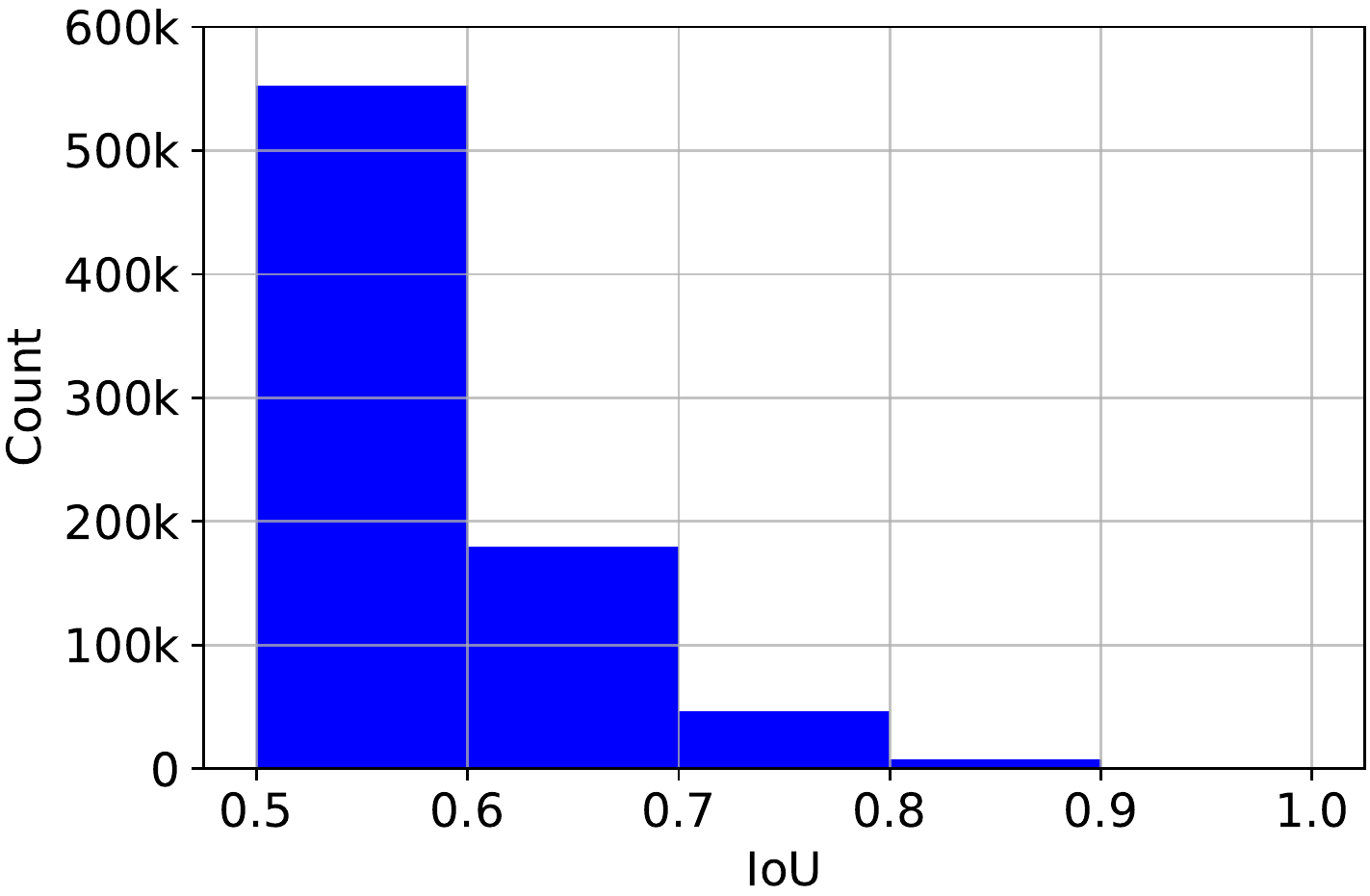}
                \caption{}
        \end{subfigure}       
        \begin{subfigure}[b]{0.32\textwidth}
                \includegraphics[width=\textwidth]{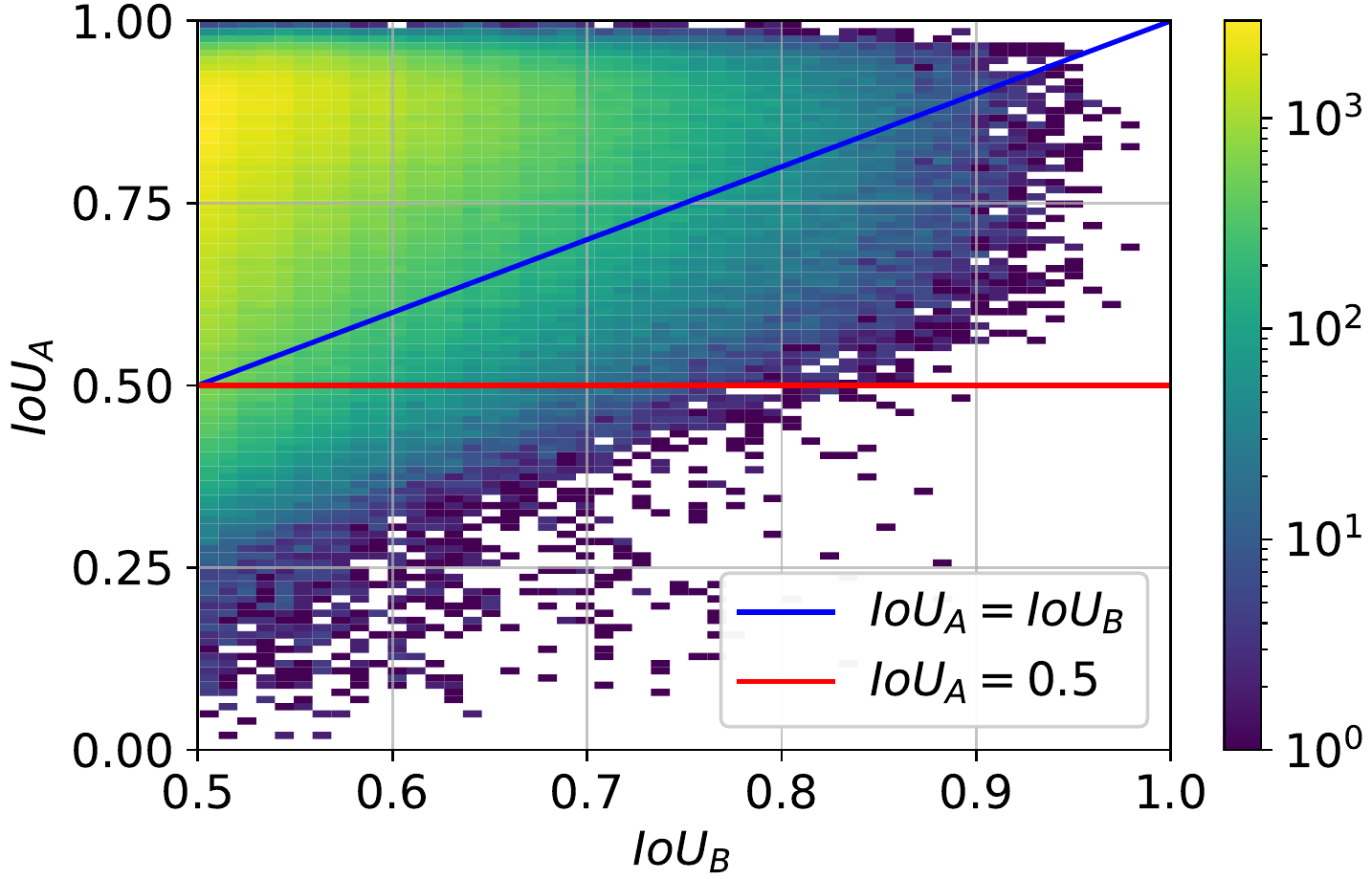}
                \caption{}
        \end{subfigure}
        \begin{subfigure}[b]{0.32\textwidth}
                \includegraphics[width=\textwidth]{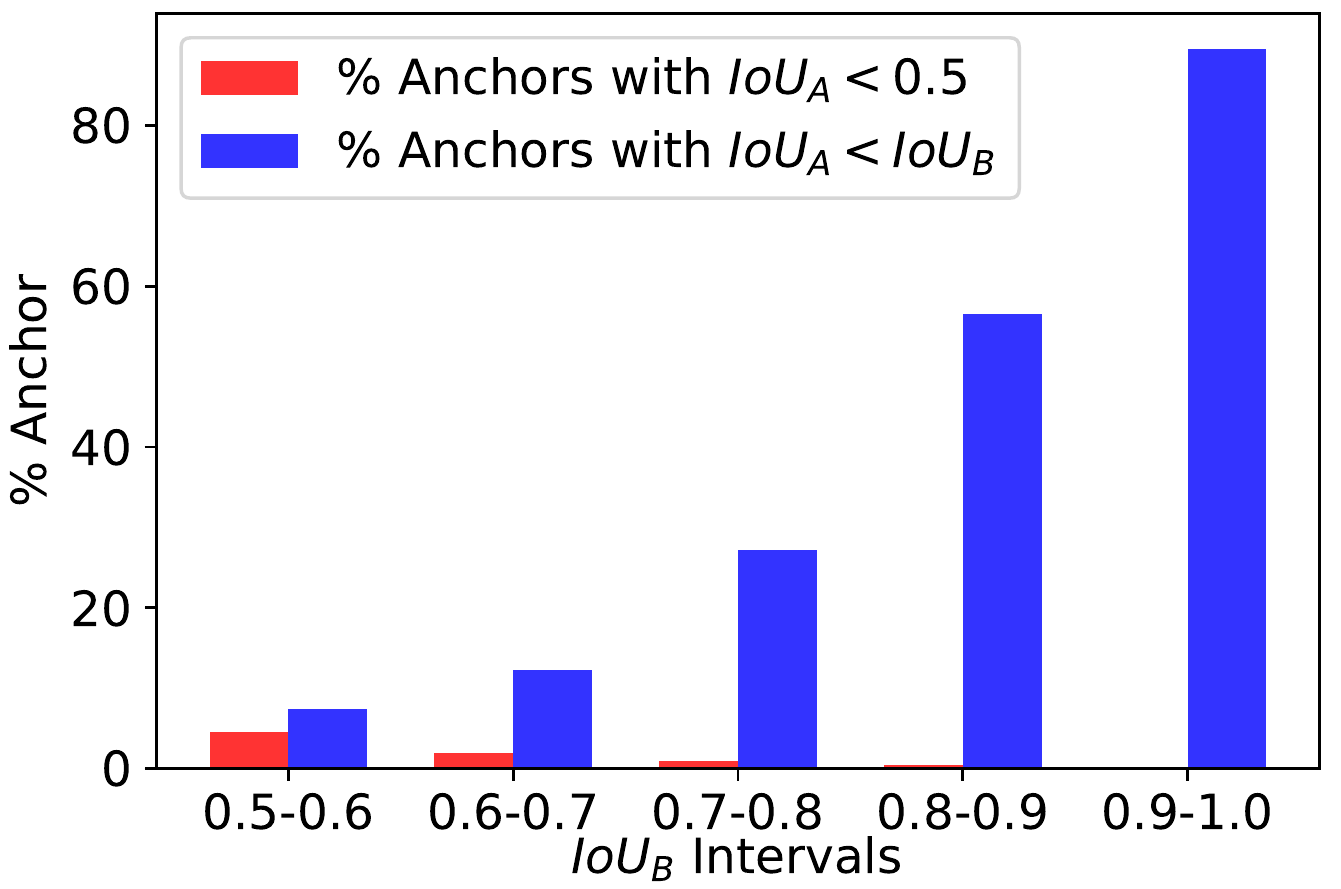}
                \caption{}
        \end{subfigure}        
        \caption{The IoU distribution of the positive anchors for a converged RetinaNet \cite{FocalLoss} with ResNet-50 \cite{ResNet} on MS COCO \cite{COCO} before regression \textbf{(a)}, and \textbf{(b)} the density of how the IoU values are affected by regression ($IoU_B$: before regression, $IoU_A$: after regression). For clarity, the density is plotted in log-scale. \textbf{(c)} A concise summary of how regression changes IoUs of anchors as a function of their starting IoUs ($IoU_B$). Notice that while it is better not to regress the boxes in larger IoUs, regressor causes false positives more in lower IoUs.}
        \label{fig:BBImb}
\end{figure*}

\noindent\textit{Definition.} IoU distribution imbalance is observed when input bounding boxes have a skewed IoU distribution. The problem is illustrated in Figure \ref{fig:BBImb}(a), where the IoU distribution of the anchors in RetinaNet \cite{FocalLoss} are observed to be skewed towards lower IoUs. It has been  previously shown that this imbalance is also observed while training two-stage detectors \cite{CascadeRCNN}. Differently, we present in Figure \ref{fig:BBImb}(b) how each anchor is affected by regression. The rate of degraded anchors after regression (i.e. positive anchors under the blue line) decreases towards the threshold that the regressor is trained upon, which quantitatively confirms the claims of Cai et al. (see Figure \ref{fig:BBImb}(c)). On the other hand, the rate of the anchors that become a false positive (i.e. positive anchors under the red line), is increasing towards the $0.5-0.6$ bin, for which around $5 \%$ of the positive anchors are lost by the regressor (see Figure \ref{fig:BBImb}(c)). Furthermore, comparing the average IoU error of the anchors before and after regression on a converged model, we notice that it is better off without applying regression and use the unregressed anchors for the $IoU_B$ intervals $0.8-0.9$ and $0.9-1.0$. These results suggest that there is still room to improve by observing the effect of the regressor on the IoUs of the input bounding boxes.

\noindent\textit{Solutions.}   The \vurgula{Cascade R-CNN} method \cite{CascadeRCNN} has been the first to address the IoU imbalance. Motivated by the arguments that (i) a single detector can be optimal for a single IoU threshold, and (ii) skewed IoU distributions make the regressor overfit for a single threshold, they show that the IoU distribution of the positive samples has an effect on the regression branch. In order to alleviate the problem, the authors trained three  detectors, in a cascaded pipeline, with IoU thresholds $0.5$, $0.6$ and $0.7$ for positive samples. Each detector in the cascade uses  the boxes from the previous stage rather than sampling them anew. In this way, they show that the skewness of the distribution can be shifted from the left-skewed to approximately uniform and even to the right-skewed, thereby allowing the model to have enough samples for the optimal IoU threshold that it is trained with. The authors show that such a cascaded scheme works better compared to the previous work, such as Multi-Region CNN \cite{IterativeBB} and AttractioNet \cite{IterativeBB2}, that  iteratively apply the same network to the bounding boxes. Another cascade-based structure implicitly addressing the IoU imbalance is \vurgula{Hierarchical Shot Detector} (HSD) \cite{HierarchicalShotDet}. Rather than using a classifier and regressor at different cascade-levels, the method runs its classifier after the boxes are regressed, resulting in a more balanced distribution. 

In another set of studies, randomly generated bounding boxes are utilized  to provide a set of positive input bounding boxes with balanced IoU distribution to the second stage of Faster R-CNN. \vurgula{IoU-uniform R-CNN} \cite{IoUUniform} adds controllable jitters and in such a way provides approximately uniform positive input bounding boxes to the  regressor only (i.e. the classification branch still uses the RPN RoIs). Differently, \vurgula{pRoI Generator} \cite{BBSampling} trains both branches with the generated RoIs but the performance improvement is not that significant probably because the training set  covers a much larger space than the test set. However, one significant contribution of Oksuz et al. \cite{BBSampling} is, rather than adding controllable jitters, they systematically generate bounding boxes using the proposed bounding box generator. Using this pRoI generator, they conducted a set of experiments for different IoU distributions and reported the following: (i) The IoU distribution of the input bounding boxes has an effect not only on the regression but also on the classification performance. (ii) Similar to the finding of Pang et al. \cite{LibraRCNN}, the IoU of the examples is related to their hardness. However, contrary to Cao et al. \cite{PrimeSample}, who argued that OHEM \cite{OHEM} has an adverse effect when applied only to positive examples, Oksuz et al. \cite{BBSampling} showed that the effect of OHEM depends on the IoU distribution of the positive input BBs. When a right-skewed IoU distribution is used with OHEM, a significant performance improvement is observed. (iii) The best performance is achieved when the IoU distribution is uniform.
\begin{figure*}
        \captionsetup[subfigure]{}
        \centering
        \begin{subfigure}[b]{0.23\textwidth}
                \includegraphics[width=\textwidth]{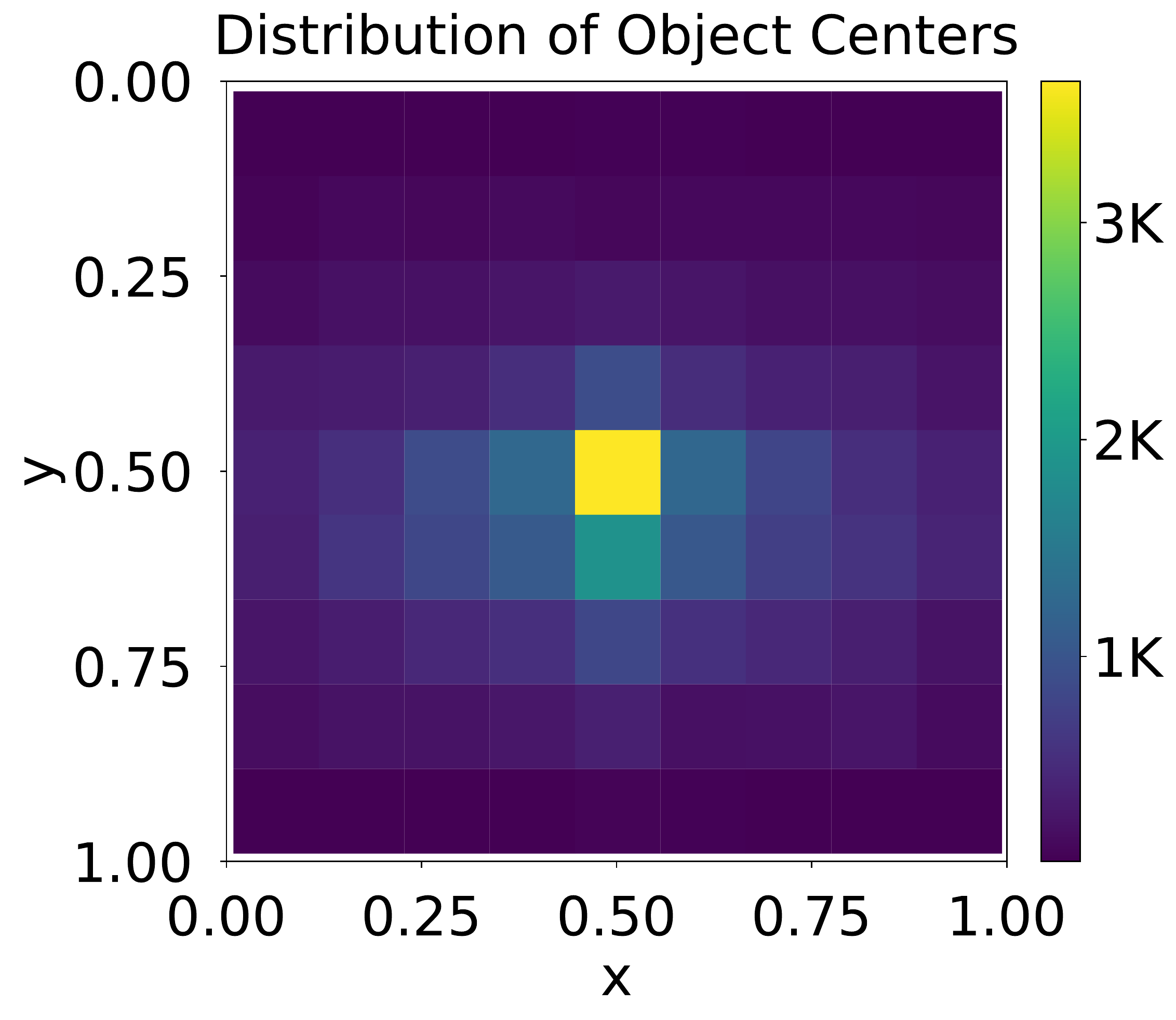}
                \caption{PASCAL-VOC \cite{PASCAL}}
        \end{subfigure}       
        \begin{subfigure}[b]{0.23\textwidth}
                \includegraphics[width=\textwidth]{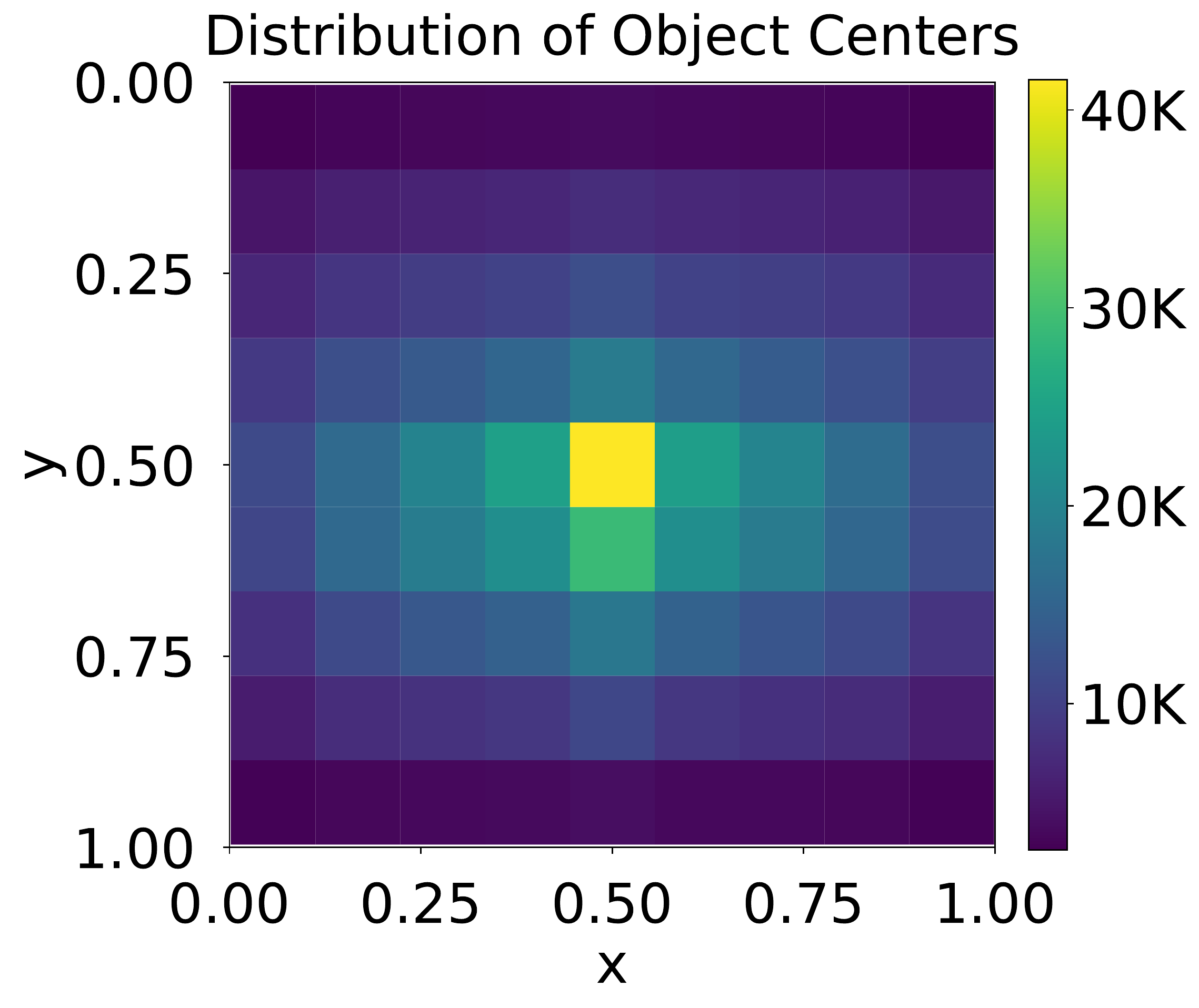}
                \caption{MS-COCO \cite{COCO}}
        \end{subfigure}
        \begin{subfigure}[b]{0.24\textwidth}
                \includegraphics[width=\textwidth]{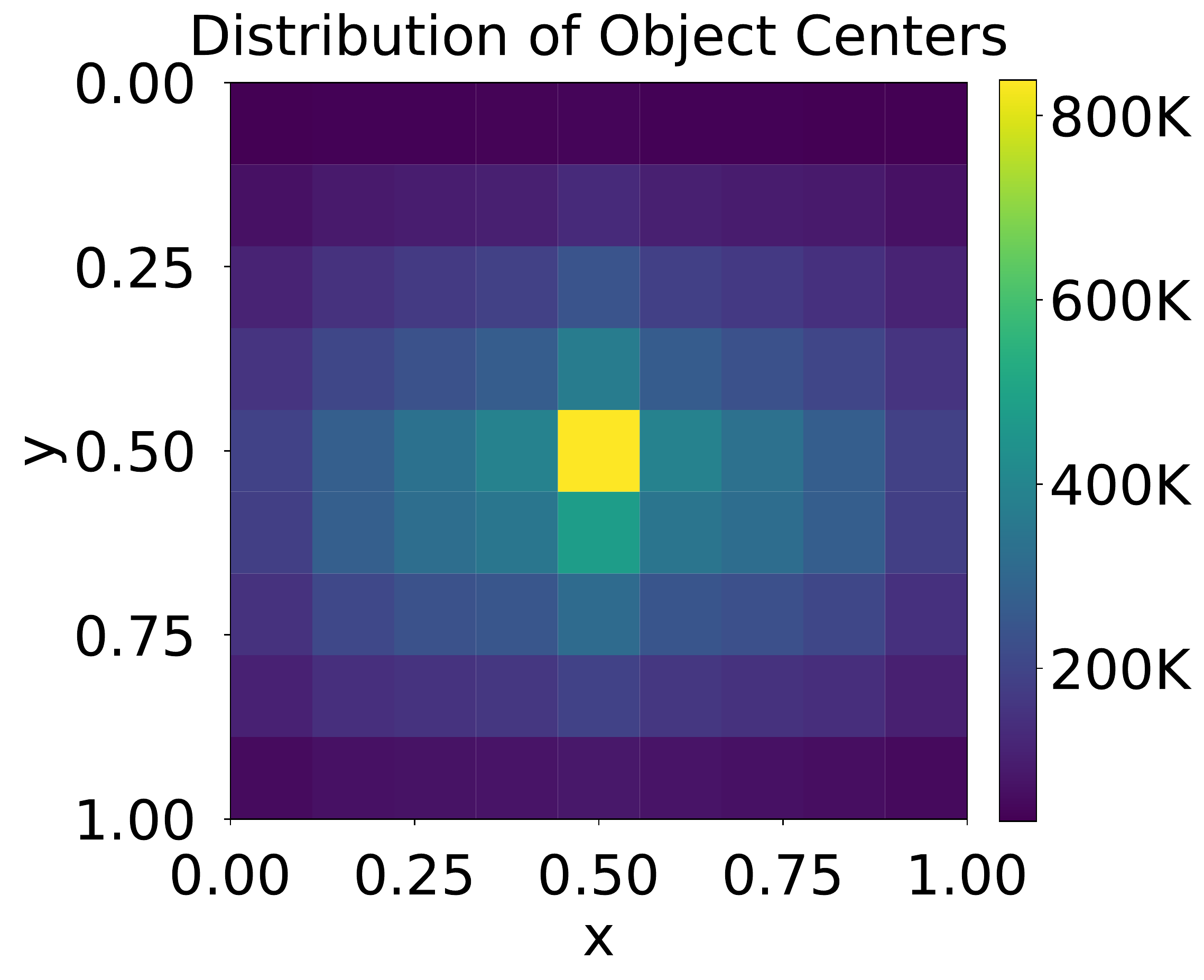}
                \caption{Open Images \cite{OpenImages}}
        \end{subfigure}        
        \begin{subfigure}[b]{0.24\textwidth}
                \includegraphics[width=\textwidth]{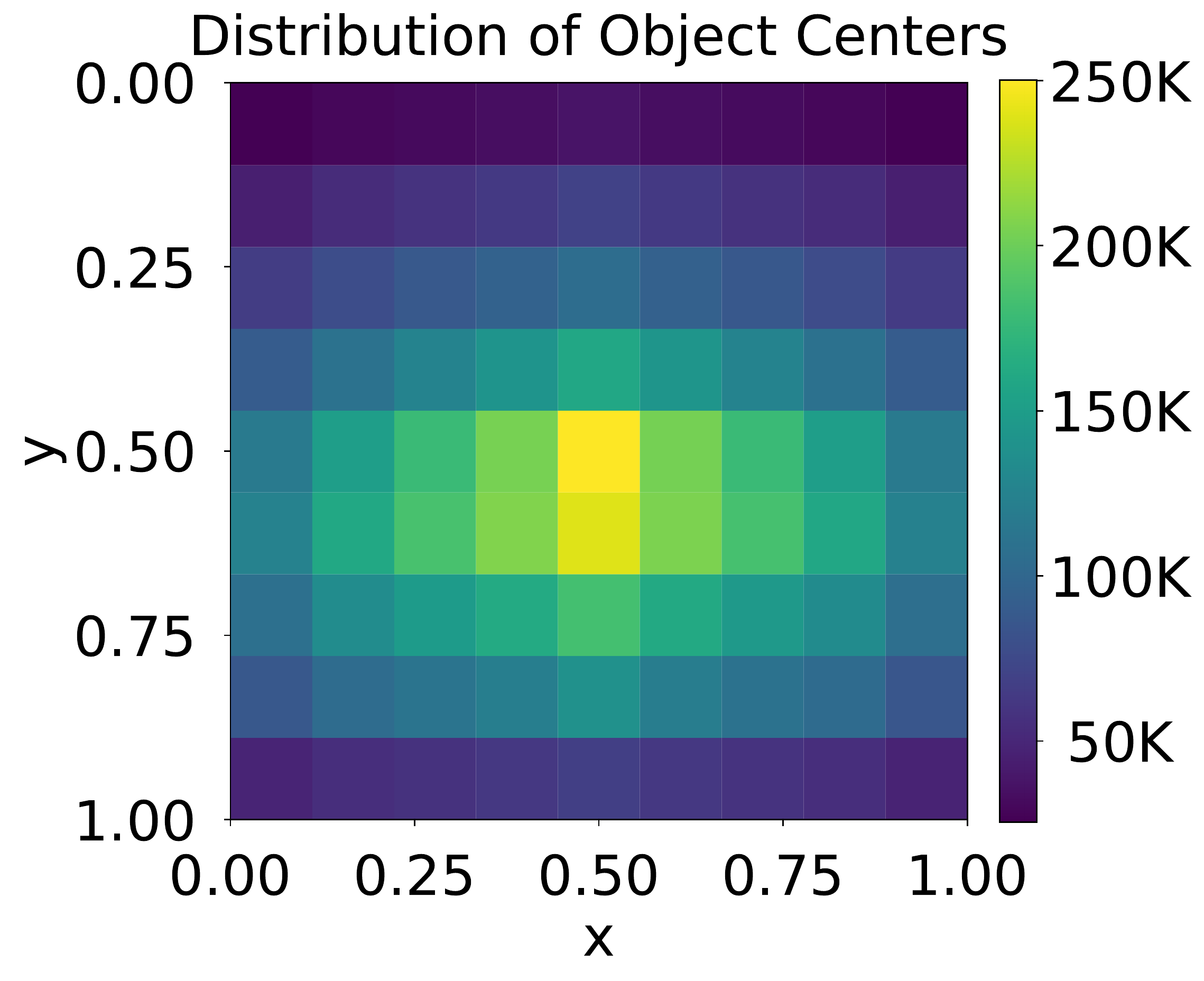}
                \caption{Objects 365 \cite{Objects365}}
        \end{subfigure}             
        \caption{Distribution of the centers of the objects in the common datasets over the normalized image.}
        \label{fig:LocationImb}
\end{figure*}

\subsection{Object Location Imbalance} 
\label{subsec:objLocImb}
\noindent\textit{Definition.} The distribution of the objects throughout the image matters because current deep object detectors employ densely sampled anchors as sliding window classifiers. For most of the methods, the anchors are evenly distributed within the image, so that each part in the image is considered with the same importance level. On the other hand, the objects in an image do not follow a uniform distribution (see Figure \ref{fig:LocationImb}), i.e. there is an imbalance about object locations.

\noindent\textit{Solutions.} Motivated by the fact that the objects are not distributed uniformly over the image, Wang et al. \cite{GuidedAnchoring} aim to learn the location, scale and aspect ratio attributes of the anchors concurrently in order to decrease the number of anchors and improve recall at the same time. Specifically, given the backbone feature maps, a prediction branch is designed for each of these tasks to generate anchors: (i) anchor location prediction branch predicts a probability for each location to determine whether the location contains an object, and a hard thresholding approach is adopted based on the output probabilities to determine the anchors, (ii) anchor shape prediction branch generates the shape of the anchor for each location. Since the anchors vary depending on the image, different from the conventional methods (i.e. one-stage generators and RPN) using a fully convolutional classifier over the feature map, the authors proposed anchor-guided feature adaptation based on deformable convolutions \cite{DeformableRoIPool} in order to have a balanced representation depending on the anchor size. Rather than learning the anchors, \vurgula{free anchor} method \cite{FreeAnchor} loosens the hard constraint of the matching strategy (i.e. a sample being positive if $IoU>0.5$) and in such a way, each anchor is considered a matching candidate for each ground truth. In order to do that, the authors pick a bag of candidate anchors for each ground truth using the sorted IoUs of the anchors with the ground truth. Among this bag of candidates, the proposed loss function aims to enforce the most suitable anchor by considering both the regression and the classification task to be matched with the ground truth. 

\subsection{Comparative Summary}
Addressing spatial imbalance problems has resulted in significant improvements for object detectors. In general, while the methods for imbalance in regression loss and IoU distribution imbalance yield improvement especially in the regressor branch, removing the bias in the anchors in the location imbalance  improves classification, too.

Despite the widespread use of the Smooth $L1$ loss, four new loss functions have been proposed last year. Chen et al. \cite{mmdetection} compared Balanced $L1$, IoU, GIoU, Bounded IoU on Faster R-CNN+FPN against Smooth $L1$, and reported relative improvement of $0.8 \%$, $ 1.1\%$, $1.4$ and $- 0.3\%$ respectively. With the same configuration, DIoU and CIoU losses are reported to have a relative improvement of $0.2 \%$ and $1.7 \%$ against the GIoU baseline (without DIoU-NMS). Compared to the loss functions designed for the foreground-background imbalance problem (i.e. Focal Loss - see Section \ref{subsec:SoftSample}), the relative improvement of the regression losses over Smooth $L1$ are smaller. We also analyzed for which cases the baseline loss function fails in Figure \ref{fig:BBImb}(b), which can be used as a tool to compare the pros/cons of different regression loss functions.

Cascaded structures are proven to be very useful for object detection by regulating the IoU distribution. Cascade R-CNN, being a two-stage detector, has a relative improvement of $18.2 \%$ on MS-COCO testdev compared to  its baseline Faster R-CNN+FPN with ResNet-101 backbone ($36.2$ vs. $42.8$). A one-stage architecture, HSD, also has a relative improvement of $34.7 \%$ compared to the SSD512 with VGG-16 backbone ($28.8$ vs $38.8$). It is difficult to compare Cascade R-CNN and HSD since they are different in nature (one-stage and two-stage pipelines), and their performances were reported for different input resolutions. However, we observed that HSD with a $768 \times 768$ input image runs approximately $1.5 \times$ faster than Cascade R-CNN for a $1333 \times 800$ image  and achieves slightly lower performance on MS-COCO testdev ($42.3$ vs. $42.8$). One observation between these two is that, even though HSD performs worse than Cascade R-CNN at $AP@0.5$, it yields better performance for $AP@0.75$ than Cascade R-CNN, which implies that the regressor of HSD is better trained.

As for the methods that we discussed for location imbalance, guided anchoring \cite{GuidedAnchoring} increases average recall by $9.1\%$ with $90\%$ fewer anchors via learning the parameters of the anchors. Compared to guided anchoring, free anchor \cite{FreeAnchor} reports a lower relative improvement for average recall with $2.4 \%$ against RetinaNet with ResNet50, however it has a $8.4 \%$ relative improvement (from $35.7$ to $38.7$). 

\subsection{Open Issues} 
\label{subsec:OpenProblemsIoUImb}

This section discusses the open issues related to the spatial properties of the input bounding boxes and objects.

\subsubsection{A Regression Loss with Many Aspects}
{\noindent}\textit{Open Issue.} Recent studies have proposed alternative regression loss definitions with different perspectives and aspects. Owing to their benefits, a single regression loss function that can combine these different aspects can be beneficial.

{\noindent}\textit{Explanation.} Recent regression loss functions have different motivations: (i) Balanced $L1$ Loss \cite{LibraRCNN} increases the contribution of the inliers. (ii) KL Loss \cite{KLLoss} is motivated from the ambiguity of the positive samples. (iii) IoU-based loss functions have the motive to use a performance metric as a loss function.  These seemingly mutually exclusive motives can be integrated to a single regression loss function. 

\subsubsection{Analyzing the Loss Functions}
In order to analyze how outliers and inliers affect the regression loss, it is useful to analyze the loss function and its gradient with respect to the inputs. 
To illustrate such an analysis, in Focal Loss \cite{FocalLoss}, the authors plot the loss function with respect to the confidence score of the ground truth class with a comparison to the cross entropy loss, the baseline. Similarly, in Balanced L1 Loss \cite{LibraRCNN}, both the loss function itself and the gradients are depicted with a comparison to Smooth L1 Loss. Such an analysis might be more difficult for the recently proposed more complex loss functions. As an example, AP Loss \cite{APLoss} is computed considering the ranking of the individual examples, which is based on the confidence scores of all BBs. So, the loss depends on the entire set rather than individual examples, which makes it difficult to plot the loss (and its gradient) for a single input as conventionally done. Another example is GIoU Loss \cite{GIoULoss}, which uses the ground truth box and the smallest enclosing box in addition to the detection box. Each box is represented by four parameters (see Section \ref{subsec:Regression}), which creates a total of twelve parameters.  For this reason, it is necessary to develop appropriate analysis methods to observe how these  loss functions penalize the examples.

\subsubsection{Designing Better Anchors}
 Designing an optimal anchor set with high recall has received limited attention. The Meta Anchor \cite{MetaAnchor} method attempts to find an optimal set of aspect ratios and scales for anchors. More recently, Wang et al. \cite{GuidedAnchoring} have improved recall more than 9\% on MS COCO dataset \cite{COCO} while using 90\% less anchors than RPN \cite{FasterRCNN}. Addressing the imbalanced nature of the locations and scales of the objects seems to be an open issue.

\begin{figure}
\centering
\includegraphics[width=0.25\textwidth]{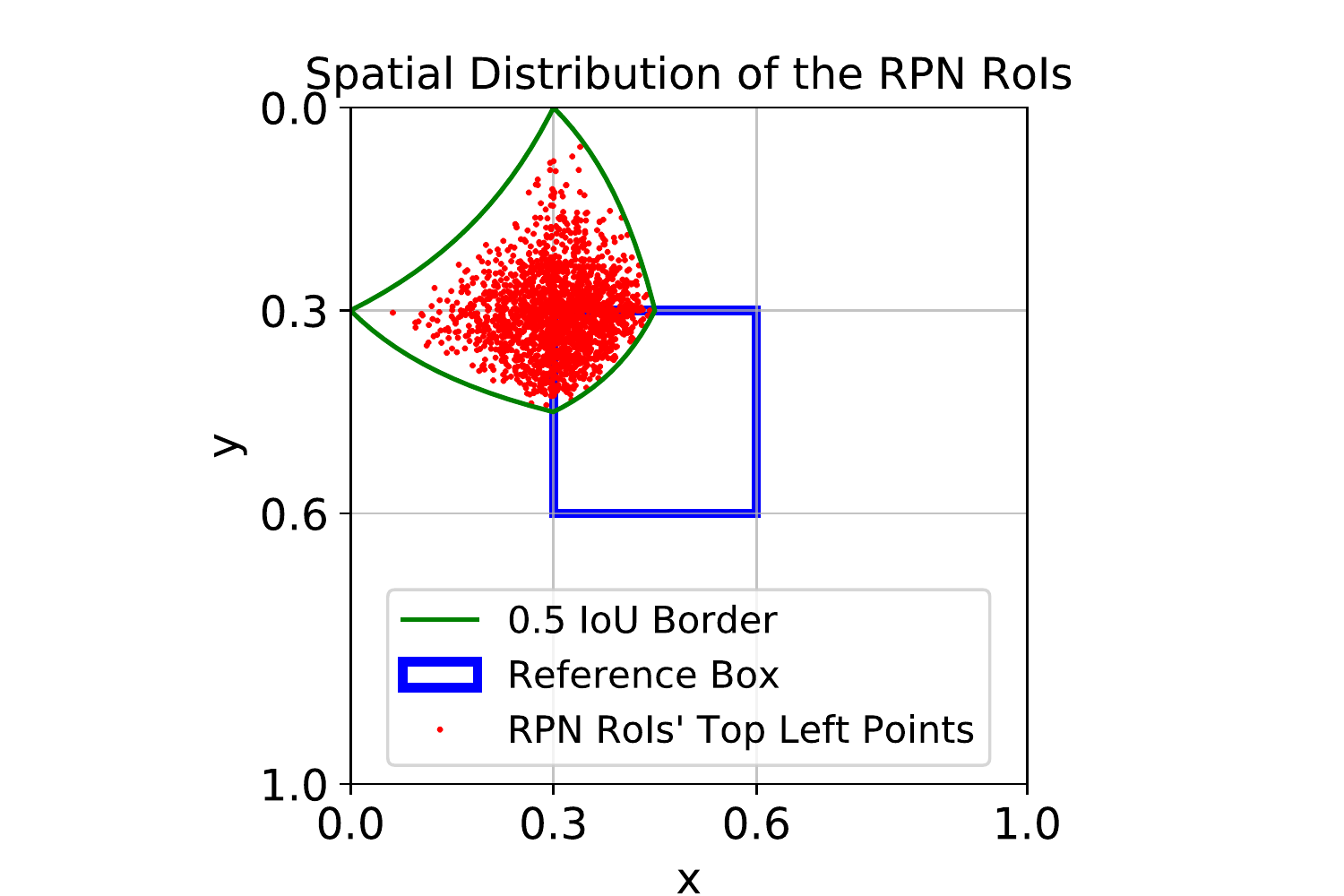}
\caption{The (relative) spatial distributions of $1K$ RoIs with IoU between $0.5$ to $0.6$ from the RPN (of Faster R-CNN with ResNet101 backbone) collected from Pascal \cite{PASCAL} during the last epoch of the training. A bias is observed around the top-left corners of ground truths such that RoIs are densely concentrated at the top-left corner of the normalized ground truth box.}
\label{fig:RelSpImb}
\end{figure}

\subsubsection{Relative Spatial Distribution Imbalance}
\label{subsec:BBSpImb}
{\noindent}\textit{Open Issue.} As we discussed in Section \ref{sec:BBimb}, the distribution of the IoU between the estimated BBs and the ground truth is imbalanced and this has an influence on the performance. A closer inspection \cite{BBSampling} reveals that the locations of estimated BBs relative to the matching ground truths also have an imbalance. Whether this imbalance affects the performance of the object detectors remains to be investigated.

{\noindent}\textit{Explanation.} During the conventional training of the object detectors, input bounding boxes are labeled as positive when their IoU with a ground truth is larger than $0.5$. This is adopted in order to provide more diverse examples to the classifier and the regressor, and to allow good quality predictions at test time from noisy input bounding boxes. The work by Oksuz et al. \cite{BBSampling} is currently the only study that points to an imbalance in the distribution of the relative location of BBs. Exploiting the scale-invariance and shift-invariance properties of the IoU, they plotted the top-left point of the RoIs from the RPN (of Faster R-CNN) with respect to a single reference box representing the ground truth (see Figure \ref{fig:RelSpImb}). They reported that the resulting spatial distribution of the top-left points of the RPN RoIs are skewed towards  the top-left point of the reference ground truth box. We see that the samples are scarce  away from the top-left corner of the reference box.

\begin{figure}
\centering
\includegraphics[width=0.45\textwidth]{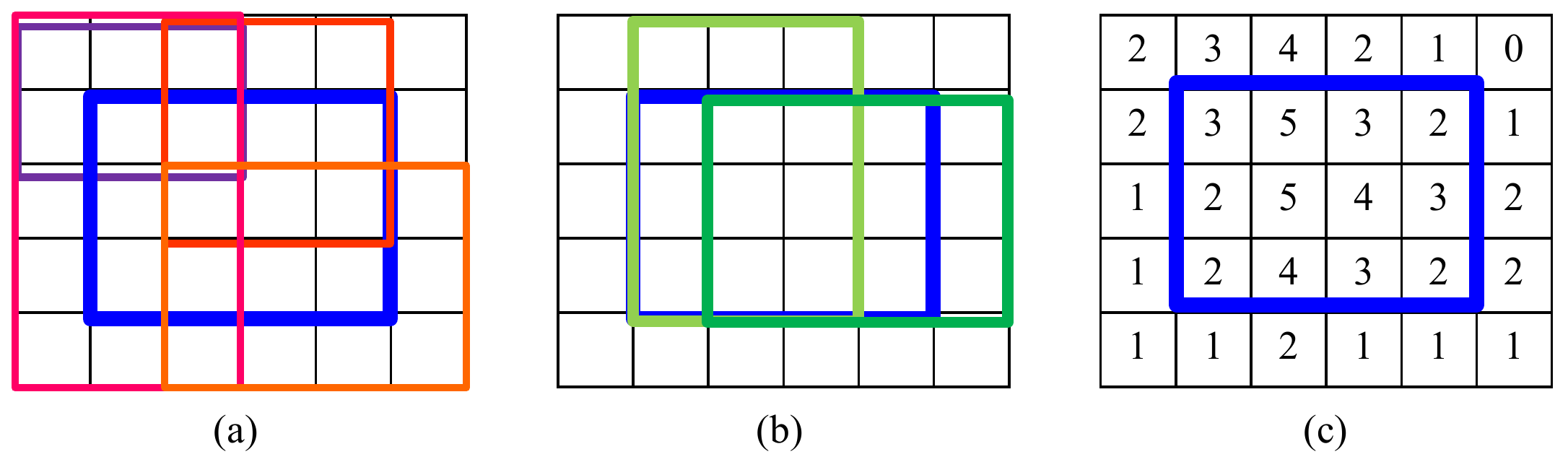}
\caption{An illustration of the imbalance in overlapping BBs. The grid represents the pixels of the image/feature map, and blue bounding box is the ground-truth. \textbf{(a)} Four negative input bounding boxes. \textbf{(b)} Two positive input bounding boxes. \textbf{(c)} Per-pixel number-of-overlaps of the input bounding boxes. Throughout the image, the number of sampling frequencies for the pixels vary due to the variation in the overlapping number of bounding boxes.}
\label{fig:BBSpatImb}
\end{figure}

\subsubsection{Imbalance in Overlapping BBs}
\label{subsub:SamplingImb}
{\noindent}\textit{Open Issue.} Due to the dynamic nature of bounding box sampling methods (Section \ref{subsec:fgbgClassImb}), some regions in the input image may be over-sampled (i.e. regions coinciding with many overlapping boxes) and some regions may be under-sampled (or not even sampled at all). The effect of this imbalance caused by BB sampling methods has not been explored. 


{\noindent}\textit{Explanation.}  Imbalance in overlapping BBs is illustrated in Figure \ref{fig:BBSpatImb}(a-c) on an example grid representing the image and six input BBs (four negative and two positive). The number of overlapping BBs for each pixel is shown in Figure \ref{fig:BBSpatImb}(c); in this example, this number ranges from $0$ to $5$. 

This imbalance may affect the performance for two reasons: (i) The number of highly sampled regions will play more role in the final loss functions, which can lead the method to overfit for specific features. (ii) The fact that some regions are over-sampled and some are under-sampled might have adverse effects on learning, as the size of sample (i.e. batch size) is known to be related to the optimal learning rate \cite{MegDet}. 

\subsubsection{Analysis of the Orientation Imbalance }
\label{subsub:OrientationImb}

{\noindent}\textit{Open Issue.} The effects of imbalance in the orientation distribution of objects need to be investigated. 

{\noindent}\textit{Explanation.}The  distribution of the orientation of object instances might have an effect on the final performance. If there is a typical orientation for the object, then the detector will likely overfit to this orientation and will make errors for the other orientations. To the best of our knowledge, this problem has not yet been explored.

\section{Imbalance 4: Objective Imbalance} 
\label{sec:ObjImb}
\begin{figure}
        \captionsetup[subfigure]{}
        \centering
        \begin{subfigure}[b]{0.21\textwidth}
                \includegraphics[width=\textwidth]{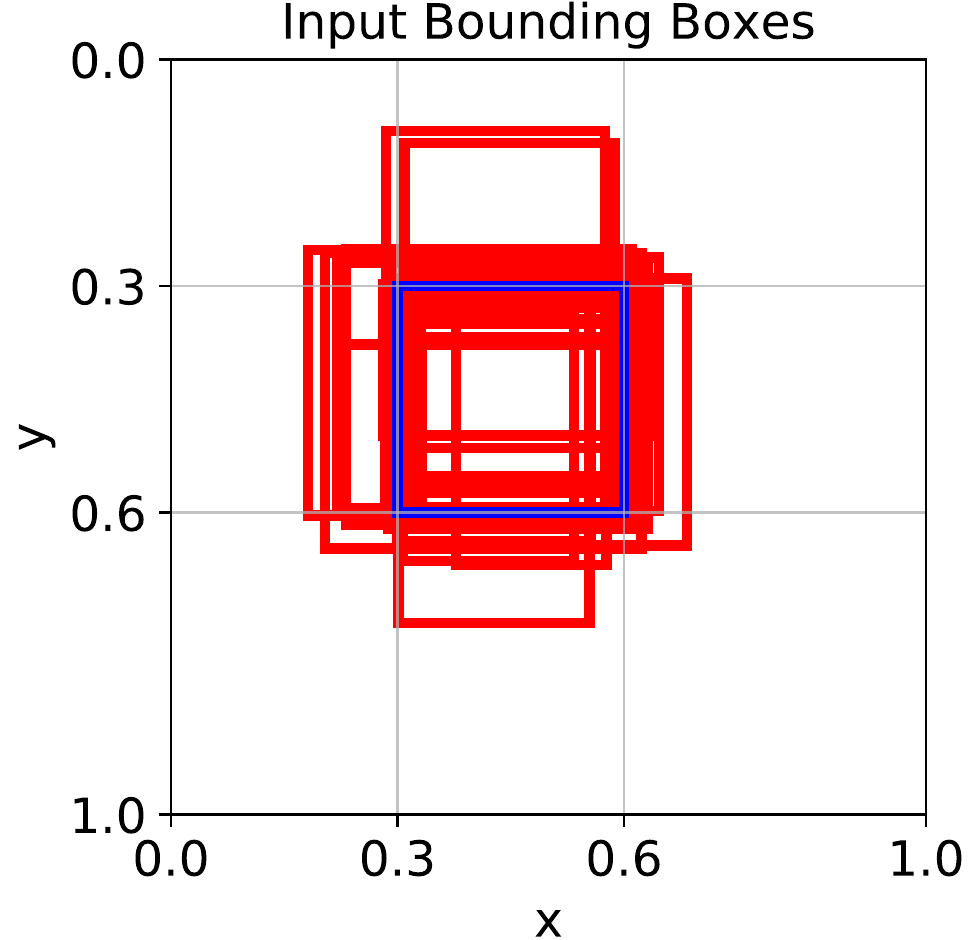}
                \caption{}
        \end{subfigure}       
        \begin{subfigure}[b]{0.20\textwidth}
                \includegraphics[width=\textwidth]{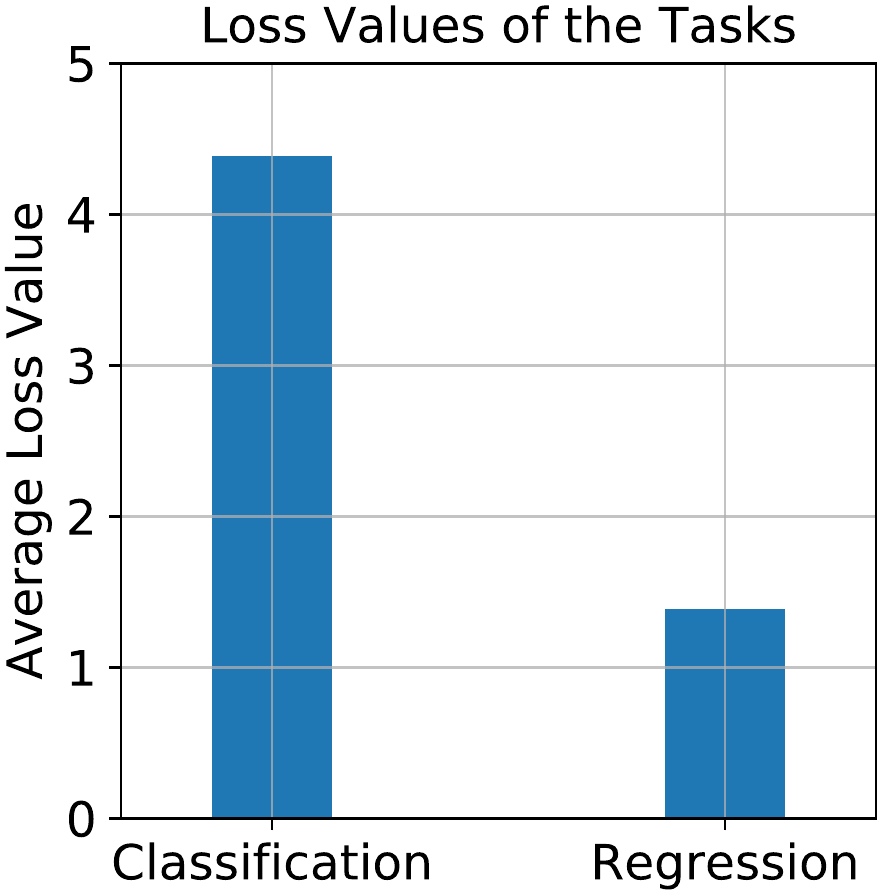}
                \caption{}
        \end{subfigure}
        \caption{\textbf{(a)} Randomly sampled 32 positive RoIs using the pRoI Generator \cite{BBSampling}. \textbf{(b)} Average classification and regression losses of these RoIs at the initialization of the object detector for MS COCO dataset \cite{COCO} with 80 classes. We use cross entropy for the classification task assuming that initially each class has the same confidence score, and smooth L1 loss for regression task. Note that right after initialization, the classification loss has more effect on the total loss.}
        \label{fig:ObjImbalance}
\end{figure}

\noindent\textit{Definition.} Objective imbalance pertains to the objective (loss) function that is minimized during training. By definition, object detection requires a multi-task loss in order to solve classification and regression tasks simultaneously. However, different tasks can lead to imbalance because of the following differences: (i) The norms of the gradients can be different for the tasks, and one task can dominate the training (see Figure \ref{fig:ObjImbalance}). (ii) The ranges of the loss functions from different tasks can be different, which hampers the consistent and balanced optimization of the tasks. (iii) The difficulties of the tasks can be different, which affects the pace at which the tasks are learned, and hence hinders the training process \cite{DynamicTaskPri}. 

Figure \ref{fig:ObjImbalance} illustrates a case where the loss of the classification dominates the overall gradient.

{\noindent}\textit{Solutions.} The most common solution is \vurgula{Task Weighting} which balances the loss terms by an additional hyper-parameter as the weighting factor. The hyper-parameter is selected using a validation set. Naturally, increasing the number of tasks, as in the case of two-stage detectors, will increase the number of weighting factors and the dimensions of the search space (note that there are four tasks in two-stage detectors and two tasks in one-stage detectors). 

An issue arising from the multi-task nature is the possible range inconsistencies among different loss functions. For example, in \vurgula{AP Loss},  smooth L1, which is in the logarithmic range (since the input to the loss is conventionally provided after applying a logarithmic transformation) with $[0, \infty)$, is used for regression while $\mathcal{L}_{AP} \in [0,1]$. Another example is the GIoU Loss \cite{GIoULoss}, which is in the $[-1,1]$ range and used together with  cross entropy loss. The authors set the weighting factor of GIoU Loss to $10$ and regularization is exploited to balance this range difference and ensure balanced training. 

Since it is more challenging to balance terms with different ranges, it is a better strategy to first make the ranges comparable.

A more prominent approach to combine classification and regression tasks is 
\vurgula{Classification-Aware Regression Loss} (CARL) \cite{PrimeSample}, which assumes that classification and regression tasks are correlated. To combine the loss terms, the regression loss is scaled by a coefficient determined by the (classification) confidence score of the bounding box:
\begin{align}
    L_{CARL}(x) = c_i'  L1_{smooth}(x), 
\end{align}
where $c_i'$ is a factor based on $p_i$, i.e., an estimation from the classification task. In this way, regression loss provides gradient signals to the classification branch as well, and therefore, this formulation improves localization  of high-quality (prime) examples.

An important contribution of CARL is to employ the correlation between the classification and regression tasks. However, as we discuss in Section \ref{subsec:OpenProblemsObjImb}, this correlation should be investigated and exploited more extensively.

Recently, Chen et al. \cite{SamplingHeuristics} showed that cumulative loss originating from cross entropy needs to be dynamically weighted since, when cross entropy loss is used, the contribution rate of individual loss component at each epoch can be different. To prevent this imbalance, the authors proposed \vurgula{Guided Loss} which simply weights the classification component by considering the total magnitude of the losses as:
\begin{align}
    \frac{w_{reg} L_{Reg}}{L_{cls}}.
\end{align}
The motivation of the method is that regression loss consists of only foreground examples and is normalized only by number of foreground classes, therefore it can be used as a normalizer for the classification loss.

\subsection{Comparative Summary} 
Currently, except for linear task weighting, there is no  method suitable for all  architectures alleviating objective imbalance, and unless the weights of the tasks are set accordingly, training diverges \cite{SamplingHeuristics}. While many studies use equal weights for the regression and classification tasks \cite{SSD,FocalLoss}, it is also shown that appropriate linear weighting can lead to small improvements on the performance \cite{mmdetection}. However, an in-depth analysis on objective imbalance is missing in the literature. 

CARL \cite{PrimeSample} is a method to promote examples with higher IoUs, and in such a way, $1.6 \%$ relative improvement is achieved compared to prime sampling without CARL (i.e. from $37.9$ to $38.5$). Another method, Guided Loss \cite{SamplingHeuristics}, weights the classification loss dynamically to ensure balanced training. This method achieves a similar performance with the baseline RetinaNet, without using Focal Loss. However, their method does not discard the linear weighting, and they search for the optimal weight for different architectures. Moreover, the effect of Guided Loss is not clear for the two-stage object detectors, which conventionally employ cross entropy loss.



\subsection{Open Issues}
\label{subsec:OpenProblemsObjImb}


{\noindent}\textit{Open Issue.} Currently, the most common approach is to linearly combine the loss functions of different tasks to obtain the overall loss function (except for classification-aware regression loss in \cite{PrimeSample}). However, as shown in Figure \ref{fig:ImbalanceUnified}(a-c), as the input bounding box is slid over the image, both classification and regression losses are affected, implying their dependence. This suggests that current linear weighting strategy may not be able to address the imbalance of the tasks that is related to (i) the loss values and their gradients, and (ii) the paces of the tasks.

{\noindent}\textit{Explanation.} The loss function of one  task (i.e. classification) can affect the other task (i.e. regression). To illustrate, \vurgula{AP Loss} \cite{APLoss} did not modify the regression branch; however, COCO style $AP@0.75$ increased around $3\%$. This example shows that the loss functions for different branches (tasks) are not independent (see also Figure \ref{fig:ImbalanceUnified}). This interdependence of tasks has been explored  in classification-aware regression loss by Cao et al. \cite{PrimeSample} (as discussed in Section \ref{sec:ObjImb}) to a certain extent. Further research is needed for a more detailed analysis of this interdependence and fully exploiting  it for object detection. 

\begin{figure}
\centering
\includegraphics[width=0.35\textwidth]{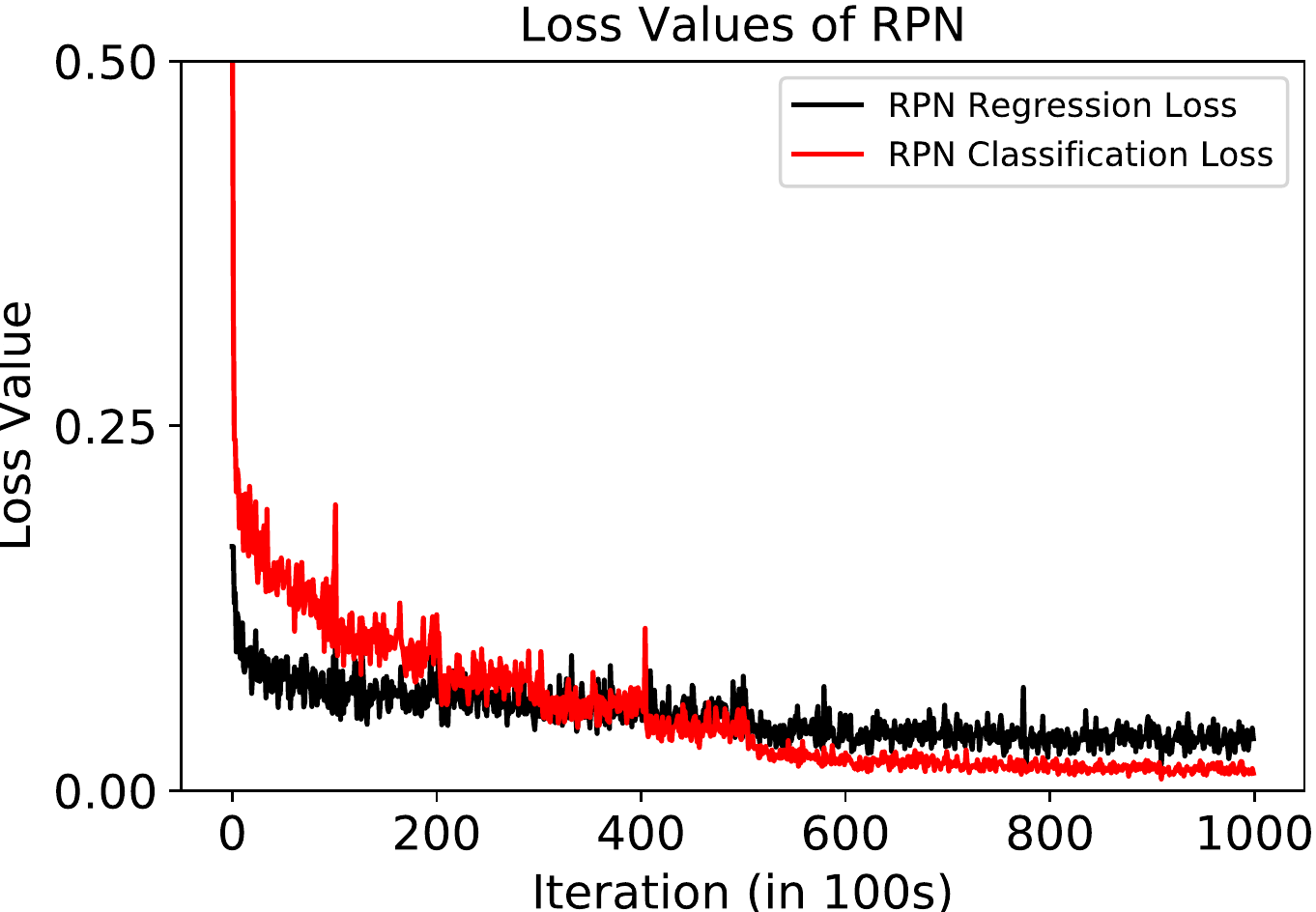}
\caption{Paces of the tasks in RPN (with Resnet-101 \cite{ResNet} backbone) \cite{FasterRCNN} on Pascal VOC 2007 training set \cite{PASCAL}.} 
\label{fig:Paces}
\end{figure}

Some studies in multi-task learning \cite{DynamicTaskPri} pointed out that learning pace affects performance. With this in mind, we plotted the regression and classification losses of the RPN \cite{FasterRCNN} during training on the Pascal VOC dataset \cite{PASCAL} in Figure \ref{fig:Paces}. We observe that the classification loss decreases faster than the regression loss.  Analyzing and balancing the learning paces of different tasks involved in the object detection problem might be another fruitful research direction. 

\section{Imbalance Problems in Other Domains}
\label{sec:OtherTasksImb}

In this section, we cover imbalance problems in other related domains in order to motivate the adaptation of  methods from related domains to the object detection problem. We identify two problems that are closely related to object detection: image classification and metric learning, discussed in individual subsections. In addition, the methods pertaining to the other domains are discussed in the last subsection.

\subsection{Image Classification}
\label{subsec:ImgClass}
Image classification is the problem of assigning a category label for an image. This problem is closely related to the object detection problem since it is one of two tasks in object detection. For image classification problem, class imbalance has been extensively studied from very different perspectives (compared to other imbalance problems), and in this section, we will focus only on class imbalance.

A common approach is \textbf{Resampling the dataset} \cite{Smote,UndeVersusOver,Adasyn,SystematicCNNImbalance,relay2016}, including oversampling and undersampling to balance the dataset. While oversampling adds more samples from the under-represented classes, undersampling balances over the classes by ignoring some of the data from the over-represented classes. When employed naively, oversampling may suffer from overfitting because duplication of samples from the under-represented classes can introduce bias. Therefore, despite it means ignoring a portion of the training data, undersampling was preferable for non-deep-learning approaches \cite{UndeVersusOver}. On the other hand, Buda et al. \cite{SystematicCNNImbalance} showed that deep neural networks do not suffer from overfitting under oversampling and in fact, better performance can be achieved compared to undersampling. 

Over the years, methods more complicated than just duplicating examples have been developed. For example, Chawla et al. \cite{Smote} proposed \vurgula{Smote} as a new way of oversampling by producing new samples as an interpolation of neighboring samples. \vurgula{Adasyn} \cite{Adasyn}, an extension of Smote to generate harder examples, aims to synthesize samples from the underrepresented classes. Li et al. \cite{relay2016} sample a mini-batch as uniform as possible from all classes also by restricting the same example and class to appear in the same order, which implies promoting the minority classes. 

Another approach addressing class imbalance in image classification is \textbf{transfer learning}  \cite{LearnToModelTail,LargeScaleFineGrained}. For example, Wang et al. \cite{LearnToModelTail} design a model (i.e. meta-learner) to learn how a model evolves when the size of the training set increases. The meta-learner model is trained gradually by increasing the provided number of examples from the classes with a large number of examples. The resulting meta-model is able to transform another model trained with less examples to a model trained with more examples, which makes it useful to be exploited by an underrepresented class. Another study \cite{LargeScaleFineGrained} adopted a different strategy during transfer learning: Firstly, the network is trained with the entire imbalanced dataset, and then the resulting network is fine-tuned by using a balanced subset of the dataset. 

\textbf{Weighting the loss function} \cite{LearningDeepRep,ClassBalancedLoss} is yet another way of balancing the classes (we called it soft sampling in the paper). Among these approaches, Huang et al. \cite{LearningDeepRep} use the inverse class frequency (i.e. $1/|C_i|$) to give more weight to under-represented classes. This is extended by class-balanced loss \cite{ClassBalancedLoss} to adopt inverse class frequency by employing a hyperparameter $\beta$ as  $w_i = (1-\beta)/(1-\beta^{|C_i|})$. When $\beta=0$, the weight degenerates to $1$; and  $\beta \rightarrow 1$ makes the weight approximate $1/|C_i|$, the inverse class frequency. 

Similar to object detection, giving more importance to ``useful'' examples is also common \cite{SelfPacedLearning,HideSeek,Dong2017ClassRH,NIPS2017_6701}. As done in object detection, hard examples have been utilized, e.g. by Dong et al. \cite{Dong2017ClassRH} who identify hard (positive and negative) samples in the batch level. The proposed method uses the hardness in two levels: (i) Class-level hard samples are identified based on the predicted confidence for the ground-truth class. (ii) Instance-level hard examples are the ones with larger $L2$ distance to a representative example in the feature space. An interesting approach that has not been utilized in object detection yet is to focus training on examples on which the classifier is uncertain (based on its prediction history) \cite{NIPS2017_6701}. Another approach is to generate hard examples by randomly cropping parts of an image as in Hide and Seek \cite{HideSeek}, or adopting a curricular learning approach by first training the classifier on easy examples and then on hard examples \cite{SelfPacedLearning}.

Another related set of methods addresses image classification problem from the \textbf{data redundancy} perspective \cite{Submodularity,SubmodularityImgCl,CoreSetApproach,AreAllEqual,10percent}. Birodkar et al. \cite{10percent} showed that around 10\% of the data in ImageNet \cite{ImageNet} and CIFAR-10 datasets is redundant during training.  This can be exploited not only for balancing the data but also to obtain faster convergence by ignoring useless data. The methods differ from each other on how  they mine for the redundant examples. Regardless of whether imbalance problem is targeted or not, a subset of these methods uses the active learning paradigm, where an oracle is used to find out the best set of training examples. The core set approach \cite{CoreSetApproach} uses the relative distances of the examples in the feature space to determine redundant samples whereas Vodrahalli et al. \cite{AreAllEqual} determine redundancies by looking at the magnitude of the gradients. 

Another mechanism to enrich a dataset is to use \textbf{weak supervision} for incorporating unlabelled examples. An example study is by Mahajan et al. \cite{Instagram}, who augment the dataset by systematically including Instagram images with the hashtags as the labels, which are rather noisy. In another example, Liu et al. \cite{SelectNet} selectively add unlabeled data to the training samples after labeling these examples using the classifier. 

Generative models (e.g. GANs) can also be used for extending the dataset to address imbalance. Many studies  \cite{OverSampleGAN,OverSampleGAN2,OverSampleGAN3} have successfuly used GANs to generate examples for under-represented classes for various image classification problems.

Special cases of image classification (e.g. face recognition) are also affected by imbalance  \cite{RangeLoss,Face1,Face2}. The general approaches are similar and therefore, due to the space constraint, we omit imbalance in specialized classification problems.

\noindent\textit{Comparative Summary.} Our analysis reveals that object detection community can benefit from the imbalance studies in image classification in many different aspects. The discussed methods for image classification are (by definition) the present solutions for the foreground-foreground class imbalance problem (see Section 4.2); however, they can possibly be extended to the foreground-background class imbalance problem. Foreground-background class imbalance is generally handled by under-sampling for the object detection problem, and other advanced resampling or transfer learning methods are not adopted yet for object detection from a class imbalance perspective. While there are loss functions (discussed in Section 4.1.2) that exploit a weighting scheme \cite{FocalLoss,gradientharmonizing}, Cui et al. \cite{ClassBalancedLoss} showed that class-balanced loss is complementary to the focal loss \cite{FocalLoss} in that focal loss aims to focus on hard examples while class-balanced loss implies balancing over all classes. However, since the number of background examples is not defined, the current definition does not fit into the object detection context. Similarly, the adoption of weakly supervised methods to balance under-represented classes or data redundancy by also decreasing the samples from an over-represented class can be used to alleviate the class imbalance problem. Finally, there are only a few generative approaches in object detection, much less than those proposed for addressing imbalance in image classification.

\subsection{Metric Learning}

Metric learning methods aim to find an embedding of the inputs, where the distance between the similar examples is smaller than the distance between dissimilar examples. In order to model such similarities, the methods generally employ pairs \cite{DDML,LearnVisSimilarity} or triplets \cite{FaceNet,TripletNetworkMetricLearning,HierarchiacalTriplet} during training. In the pair case, the loss function uses the information about whether both of the samples are from the same or different classes. In contrast, training using triplets require an anchor example, a positive example from the same class with the anchor and a negative example from a different class from the anchor. The triplet-wise training scheme introduces imbalance over positive and negative examples in favor of the latter one, and the methods also look for the hard examples as in object detection in both the pair-wise and triplet-wise training schemes. Accordingly, to present the imbalance and useful example mining requirement, note that there are approximately $O(n^2)$ pairs and $O(n^3)$ triplets assuming that the dataset size is $n$. This increase in the dataset size makes it impossible to mine for hard examples by processing the entire dataset to search for the most useful (i.e. hard) examples. For this reason, similar to object detection, a decision is to be made on which samples to be used during training. 

The metric learning methods use sampling, generative methods or novel loss functions to alleviate the problem. Note that this is very similar to our categorization of foreground-background class imbalance methods in Section 4.1, which also drove us to examine metric learning in detail. Owing to its own training, and resulting loss function configuration, here we only examine sampling strategies and generative methods.

\textbf{Sampling Methods} aim to find a useful set of training examples from a large training dataset. The usefulness criterion is considered to be highly relevant to the hardness of an example. Unlike the methods in object detection, some of the methods avoid using the hardest possible set of examples during traning. One example proposed by Schroff et al. \cite{FaceNet} use a rule based on Euclidean distance to define ``semi-hard'' triplets since selecting the hardest triplets can end up in local minima during the early stages of the training. This way, they decrease the effect of confusing triplets and avoid repeating the same hardest examples. Another approach that avoids considering the hardest possible set of examples is Hard-Aware Deeply Cascaded Embedding \cite{Yuan_2017_ICCV}, which proposes training a cascaded model such that the higher layers are trained with harder examples while the first layers are trained with the entire dataset. Similarly, Smart Mining \cite{SmartMetricLearning} also mines semi-hard examples exploiting the distance between nearest neighbor of anchor and the corresponding anchor and one novelty is that they increase the hardness of the negative examples in the latter epochs adaptively. Note that neither semi-hardness nor adaptive setting of the hardness level is considered by object detectors.

As a different approach, Cui et al. \cite{HumanLabelingMetricLearning} consider to exploit humans to label false positives during training, which are identified as hard examples and be added to the mini-batch for the next iteration. Song et al. \cite{songCVPR16} mine hard negatives not only for the anchor but also for all of the positives. Note that no object detection method considers the relation between all positives and the negative example while assigning a hardness value to a negative example. One promising idea shown by Huang et al. \cite{SimilarityAwareMetricLearning} is that larger intra-class distances can confuse the hard example mining process while only inter-class distance is considered during mining. For this reason, a position-dependent deep metric unit is proposed to take into account the intra-class variations.  


Similar to the generative methods for object detection (Section 4.1.4), \textbf{Generative Methods} have been used for generating examples or features for metric learning as well. Deep Adversarial Metric Learning \cite{Duan_2018_CVPR} simultaneously learns and embedding and a generator. Here, the generator outputs a synthetic hard negative example given the original triplet. Similar to Tripathi et al. \cite{GenerateSynthetic}, Zhao et al. \cite{AdversarialTriplet} also use a GAN \cite{GAN} in order to generate not only hard negatives but also hard positives. The idea to consider inter-class similarity have proven well as in the work by Huang et al. \cite{SimilarityAwareMetricLearning}. Finally, a different approach from the previous generative models, Hardness Aware Metric Learning \cite{HardnessAware}, aims to learn an autoencoder in the feature space. The idea is as follows: The authors first manipulate the features after the backbone such that the hardness of the example can be controlled by linearly interpolating the embedding towards the anchor by employing a coefficient relying on the loss values at the last epoch. Since it is not certain that the interpolated embedding preserves the original label, a label preserving mapping back to the feature space is employed using the autoencoder. Also, similar to Harwood et al. \cite{SmartMetricLearning}, the hardness of the examples in the latter epochs is increased.

\noindent\textit{Comparative Summary.} Looking at the studies presented above, we observe that the metric learning methods are able to learn an embedding of the data that preserves the desired similarity between data samples. Object detection literature have used different measures and metrics that have been designed by humans. However, as shown by the metric learning community, a metric that is directly learned from the data itself can yield better results and have interesting properties.  Moreover, the self-paced learning, where the hardness levels of the examples is increased adaptively, is definitely an important concept for addressing imbalance in object detection. Another idea that can be adopted by the object detectors is to label the examples by humans in an online manner (similar to the work by Yao \cite{InteractiveObjDet}) during training and to use the semi-hardness concept. 

\subsection{Multi-Task Learning}
\label{subsec:MultiTask}
Multi-task learning involves learning multiple tasks (with potentially conflicting objectives) simultaneously. A common approach is to weigh the objectives of the tasks to balance them. 

Many methods have been proposed for assigning the weights in a more systematic manner. For example, Li et al. \cite{SelfPacedMultitask} extended the self-paced learning paradigm to multi-task learning based on a novel regularizer. The hyperparameters in the proposed regularizer control the hardness of not only the instances but also the tasks, and accordingly, the hardness level is increased during training. In another work motivated by the self-paced learning approach, Guo et al. \cite{DynamicTaskPri} use more diverse set of tasks, including object detection. Their method weighs the losses dynamically based on the exponential moving average of a predefined key performance indicator (e.g. accuracy, average precision) for each task. Similar to image classification, one can also use the uncertainty of the estimations \cite{WeightUncert} or their loss values \cite{DynamicWeightAvg} to assign weights to the tasks.

In addition to the importance or hardness of tasks, Zhao Chen and Rabinovich \cite{GradNorm} identified the importance of the pace at which the tasks are learned. They suggested that the tasks are required to be trained in a similar pace. For this end, they proposed balancing the training pace of different tasks by adjusting their weights dynamically based on a normalization algorithm motivated by batch normalization \cite{inceptionv2}.

\noindent\textit{Comparative Summary.} Being a multi-task problem, object detection can benefit significantly from the multi-task learning approaches. However, this aspect of object detectors has not received attention from the community. 


\section{Open Issues for All Imbalance Problems}
\label{sec:OpenProblems}

In this section, we identify and  discuss the issues relevant to all imbalance problems. For open issues pertaining to a specific imbalance problem, please see its corresponding section in the text.

\label{sec:OpenProblemsGeneral}

\subsection{A Unified Approach to Addressing Imbalance}

\begin{figure}
\centering
\includegraphics[width=0.45\textwidth]{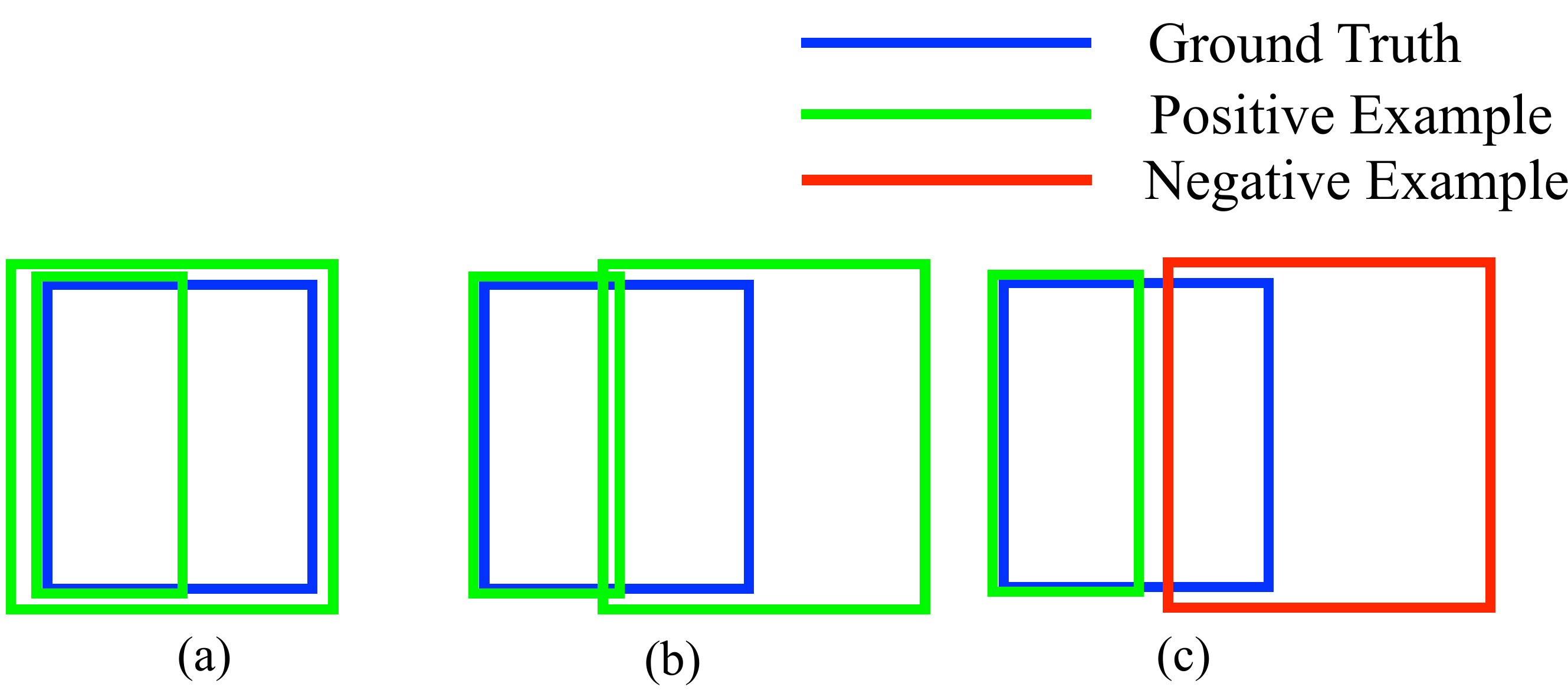}
\caption{An example suggesting the necessity of considering different imbalance problems together in a unified manner. Blue, green and red colors indicate ground-truth, positive example (prediction) and negative examples, respectively. The larger example (i.e. the one with higher IoU with the blue box) in (a) is shifted right. In (b), its IoU is significantly decreased and it eventually becomes a negative example in (c). This example shows the interplay between class imbalance, scale imbalance, spatial imbalance and objective imbalance by a change in BB positions.} 
\label{fig:ImbalanceUnified}
\end{figure}

\noindent\textit{Open Issue.} One of the main challenges is to come up with a unified approach that addresses all imbalance problems by considering the inter-dependence between them. 

\noindent\textit{Explanation.} We illustrate this inter-dependency using a toy example in Figure
\ref{fig:ImbalanceUnified}. In this figure, we shift an input bounding box with a high IoU (see Figure \ref{fig:ImbalanceUnified}(a)) to worse qualities in terms of IoU in two steps (see Figure \ref{fig:ImbalanceUnified}(b-c)) and observe how this shift affects the different imbalance problems. For the base case in Figure \ref{fig:ImbalanceUnified}(a), there are two positive bounding boxes (relevant for class imbalance) with different scales (relevant for scale imbalance), loss values (relevant for objective imbalance) and IoUs (relevant for BB imbalance). Shifting the box to the right, we observe the following:
\begin{itemize}
    \item In Figure \ref{fig:ImbalanceUnified}(b), we still have two positives, both of which now have less IoU with the ground truth (compared to (a)). 
    
    This leads to the following: (i) There are more hard examples (considering hard example mining \cite{OHEM,LibraRCNN}), and less prime samples \cite{PrimeSample}. For this reason, the methods for class imbalance are affected. (ii) The scales of the RoIs and ground truth do not change. Considering this, the scale imbalance seems not affected. (iii) Objective imbalance is affected in two ways: Firstly, the shifted BB will incur more loss (for regression, and possibly for classification) and thus, become more dominant in its own task. Secondly, since the cumulative individual loss values change, the contribution to the total loss of the individual loss values will also change, which implies its effect on objective imbalance. (iv) Finally, both BB IoU distribution and spatial distribution of the positive examples will be affected by this shift.
    
    \item In Figure \ref{fig:ImbalanceUnified}(c), by applying a small shift to the same BB, its label changes. 
    
    This leads to the following: (i) There are less positive examples and more negative examples. The ground truth class loses an example. Note that this example evolves from being a hard positive to a hard negative in terms hard example mining \cite{OHEM,LibraRCNN}, and the criterion that the prime sample attention \cite{PrimeSample} considers for the example changed from IoU to classification score. For this reason, in this case, the methods involving class imbalance and foreground-foreground class imbalance are affected. (ii) A positive RoI is removed from the set of RoIs with similar scales. Therefore, there will be less positive examples with the same scale, which affects scale imbalance. (iii) Objective imbalance is affected in two ways: Firstly, the now-negative BB is an additional hard example in terms of classification possibly with a larger loss value. Secondly, the shifted example is totally free from the regression branch, and moved to the set of hard examples in terms of classification. Hence, the contribution of the regression loss to the total loss  is expected to  decrease, and the contribution of classification would increase. (iv) Finally, the IoU distribution of the positive examples will  be affected by this shift since a positive example is lost.
\end{itemize}

Therefore, the aforementioned imbalance problems have an intertwined nature, which needs to be investigated and identified in detail to effectively address all imbalance problems. 

\subsection{Measuring and Identifying Imbalance}

\noindent\textit{Open Issue.} Another critical issue that has not been addressed yet is how to quantify or measure imbalance, and how to identify imbalance when there is one. We identify three questions that need to be studied:

\begin{enumerate}
\item  \textit{What is a balanced distribution for a property that is critical for a task?} This is likely to be uniform distribution for many properties like, e.g. class distribution. However, different modalities may imply a different concept of balance. For example, OHEM prefers a skewed distribution around $0.5$ for the IoU distribution; left-skewed for the positives and right-skewed for the negatives.

\item \textit{What is the desired distribution for the properties that are critical for a task?} Note that the desired distribution may be different from the balanced distribution since skewing the distribution in one way may be beneficial for faster convergence and better generalization. For example, online hard negative mining \cite{OHEM} favors a right-skewed IoU distribution towards 0.5 \cite{LibraRCNN}, whereas prime sample attention prefers the positive examples with larger IoUs \cite{PrimeSample} and the class imbalance methods aim to ensure a uniform distribution from the classes.

\item \textit{How can we quantify how imbalanced a distribution is?} A straightforward approach is to consider optimal transport measures such as the Wasserstein distance; however, such methods would neglect the effect of a unit change (imbalance) in the distribution on the overall performance, thereby jeopardizing a direct and effective consideration (and comparison) of the imbalance problems using the imbalance measurements.
\end{enumerate}


\subsection{Labeling a Bounding Box as Positive or Negative}
\label{subsec:OpenBBLabeling}

\noindent\textit{Open Issue.} Currently, object detectors use IoU-based thresholding (possibly with different values) for labeling an example as positive or negative and there is no consensus on this (i.e. Fast R-CNN \cite{FastRCNN}, Retina Net \cite{FocalLoss} and RPN \cite{FasterRCNN} label input BBs with IoUs $0.5$, $0.4$ and $0.3$ as negatives respectively.). However, a consensus on this is critical since labeling is very relevant to determining whether an example is a hard example.

\noindent\textit{Explanation.} Labeling bounding boxes is highly relevant to imbalance problems since this is the step where the set of all bounding boxes are split as positives and negatives in an online manner. Of special concern are the  bounding boxes around the decision boundary, which are typically  considered as  hard examples, and   a noisy labeling over them would result in large gradients in opposite directions. In other words, in order to define the hard negatives reliably, the number of outliers should be as small as possible. For this reason, consistent labeling of the input bounding boxes as positive or negative is a prerequisite of the imbalance problems in object detection. 

\begin{figure}
\centering
\includegraphics[width=0.4\textwidth]{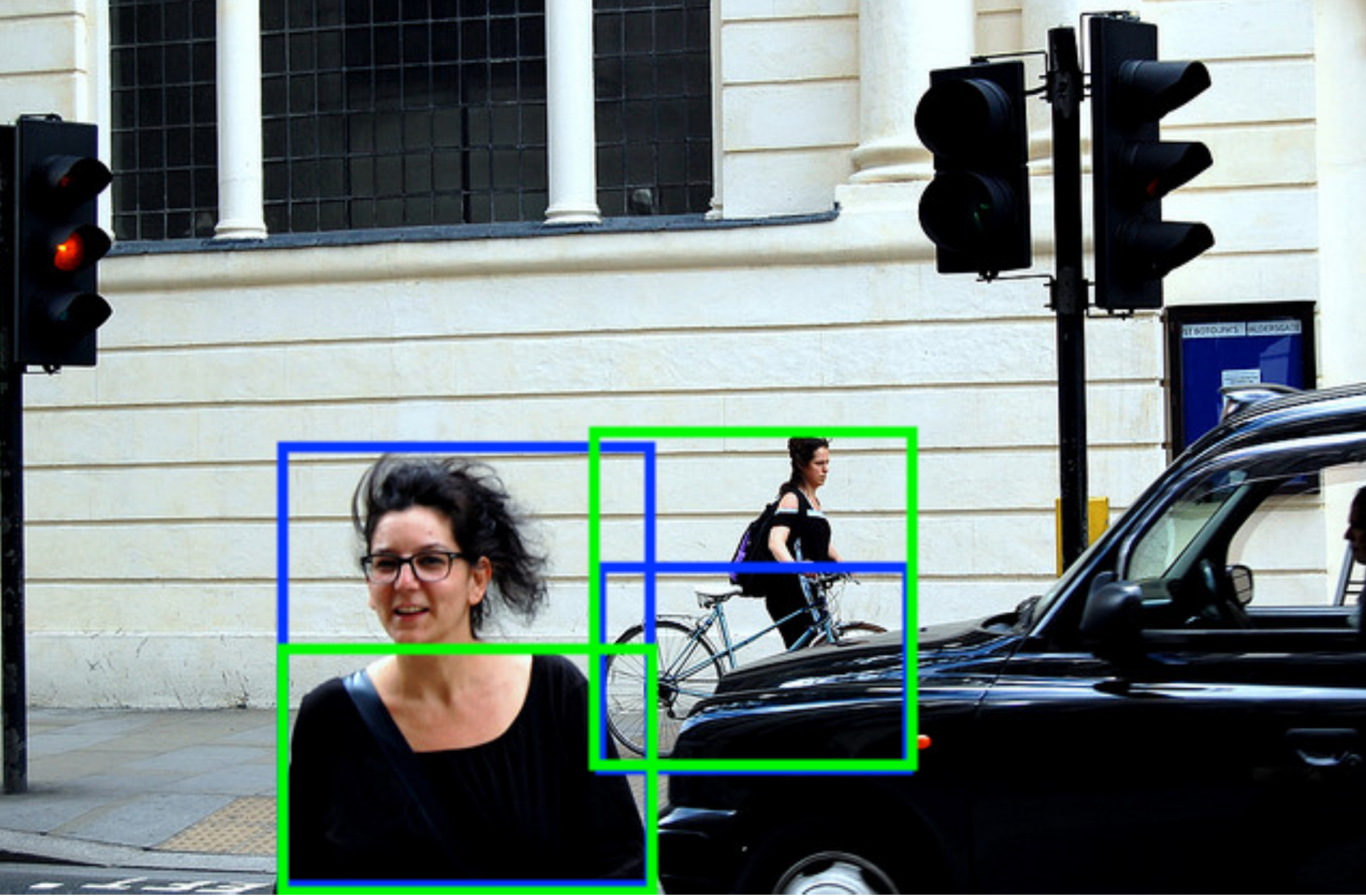} 
\caption{An illustration on ambiguities resulting from labeling examples. The blue boxes denote the ground truth. The green boxes are the estimated positive boxes with $IoU>0.5$. The input image is from the MS COCO \cite{COCO}.}
\label{fig:IoUOpen}
\end{figure}

Currently the methods use a hard IoU threshold (generally $0.5$) to split the examples; however, Li et al. \cite{gradientharmonizing} showed that this scheme results in a large number of noisy examples. In Figure \ref{fig:IoUOpen}, we illustrate two input bounding boxes that can be misleadingly labeled as positive; and once they are labeled as positive, it is likely that they will be sampled as hard positives: 
\begin{itemize}
    \item The estimated (positive) box for the bicycle (green) has two problems: It has occlusion (for the bicycle), and a big portion of it includes another object (a person). For this reason, during training,  this is  not only a hard example for the bicycle class but also a misleading example for the person class in that this specific example will try to suppress the probability of this box to be classified as person.
    \item The estimated (positive) box for the person class (green) consists of black pixels in most of it. In other words, the box does hardly include any visually descriptive part for a person. For this reason, this is a very hard example which is likely to fail in capturing the ground truth class  well.
\end{itemize}


\subsection{Imbalance in Bottom-Up Object Detectors}

{\noindent}\textit{Open Issue.} Bottom-up detectors \cite{CenterNet,CornerNet,BottomUp} adopt a completely different approach to object detection than the one-stage and the two-stage detectors (see Section \ref{sec:object_detection}). Bottom-up detectors might share many of the imbalance problems seen in the top-down detectors, and they may have their own imbalance issues as well. Further research needs to be conducted for (i) analyzing the known methods addressing imbalance problems in the context of bottom-up object detectors, and (ii) imbalance problems that are specific to bottom-up detectors. 

{\noindent}\textit{Explanation.} Addressing imbalance issues in bottom-up object detectors has received limited attention. CornerNet \cite{CornerNet} and ExtremeNet \cite{BottomUp} use focal loss \cite{FocalLoss} to address foreground-background class imbalance, and the hourglass network \cite{HourglassNet} to compensate for the scale imbalance. On the other hand, use of  hard sampling methods and the effects of other imbalance problems have not been investigated. For the top-down detectors, we can recap some of the findings: from the class imbalance perspective, Shrivastava et al. \cite{OHEM} show that the examples with larger losses are important; from the scale imbalance perspective, different architectures \cite{PANet,LibraRCNN,NASFPN} and training methods \cite{Snip,Sniper} involving feature and image pyramids are proven to be useful and finally from the objective imbalance perspective, Pang et al. \cite{LibraRCNN} showed that smooth $L1$ loss underestimates the effect of the inliers. Research is needed to come up with such findings for bottom-up object detectors.
\section{Conclusion}
\label{sec:Conclusion}

In this paper, we provided a thorough review of the imbalance problems in object detection. In order to provide a more complete and coherent perspective, we introduced a taxonomy of the problems as well as the solutions for addressing them. Following the taxonomy on problems, we discussed each problem separately in detail and presented the solutions with a unifying yet critical perspective.

In addition to the detailed discussions on the studied  problems and the solutions, we pinpointed and presented many  open issues and imbalance problems that are critical for object detection. In addition to the many open aspects that need further attention for the studied imbalance problems, we identified new imbalance issues that have not been addressed or discussed before.

With this review and our taxonomy functioning as a map, we, as the community, can identify where we are and the research directions to be followed to develop better solutions to the imbalance problems in object detection.

\section*{Acknowledgments}
This work was partially supported by the Scientific and Technological Research Council of Turkey (T\"UB\.ITAK) through the project titled ``Object Detection in Videos with Deep Neural Networks'' (grant number 117E054). Kemal \"Oks\"uz is supported by the T\"UB\.ITAK 2211-A National Scholarship Programme for Ph.D. students.

\bibliographystyle{IEEEtran}
\bibliography{proposalbibliography}

%

\begin{IEEEbiography}[{\includegraphics[width=1in,height=1.25in,clip,keepaspectratio]{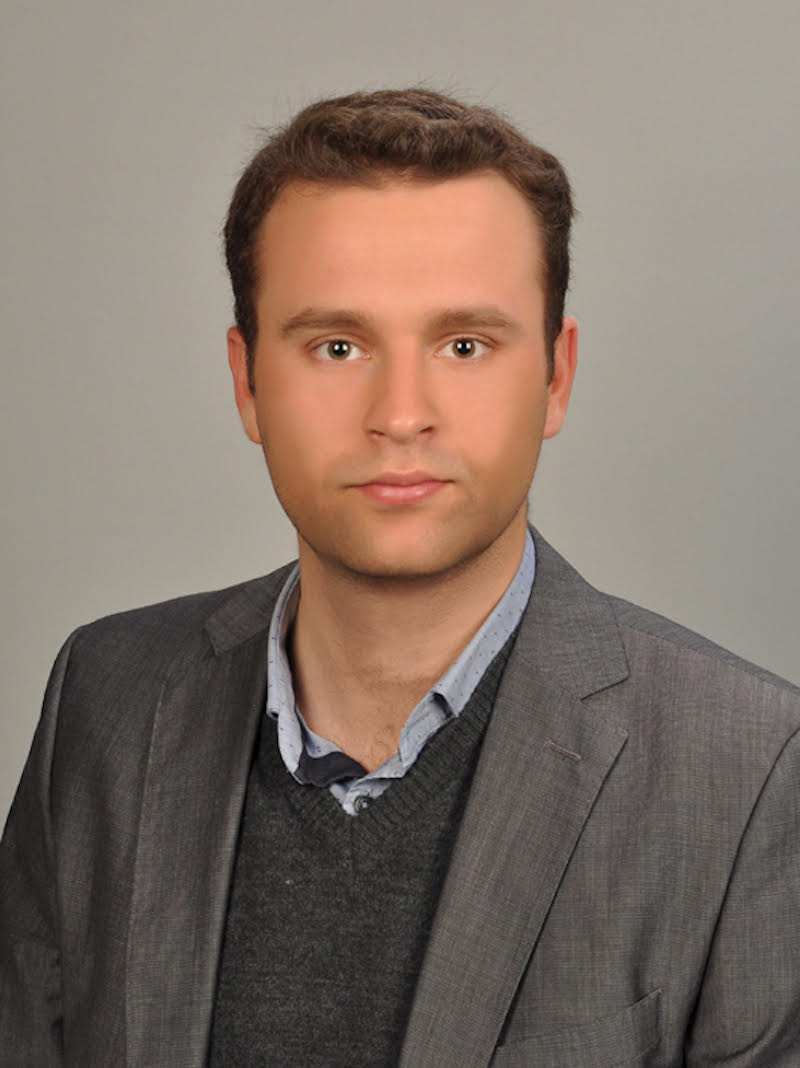}}]{Kemal Oksuz}
received B.Sc. in System Engineering from Land Forces Academy of Turkey in 2008 and his M.Sc. in Computer Engineering in 2016 from Bogazici University with high honor degree. He is currently pursuing his Ph.D. in Middle East Technical University, Ankara, Turkey. His research interests include computer vision with a focus on object detection.
\end{IEEEbiography}
\vspace{-1cm}
\begin{IEEEbiography}[{\includegraphics[width=1in,height=1.25in,clip,keepaspectratio]{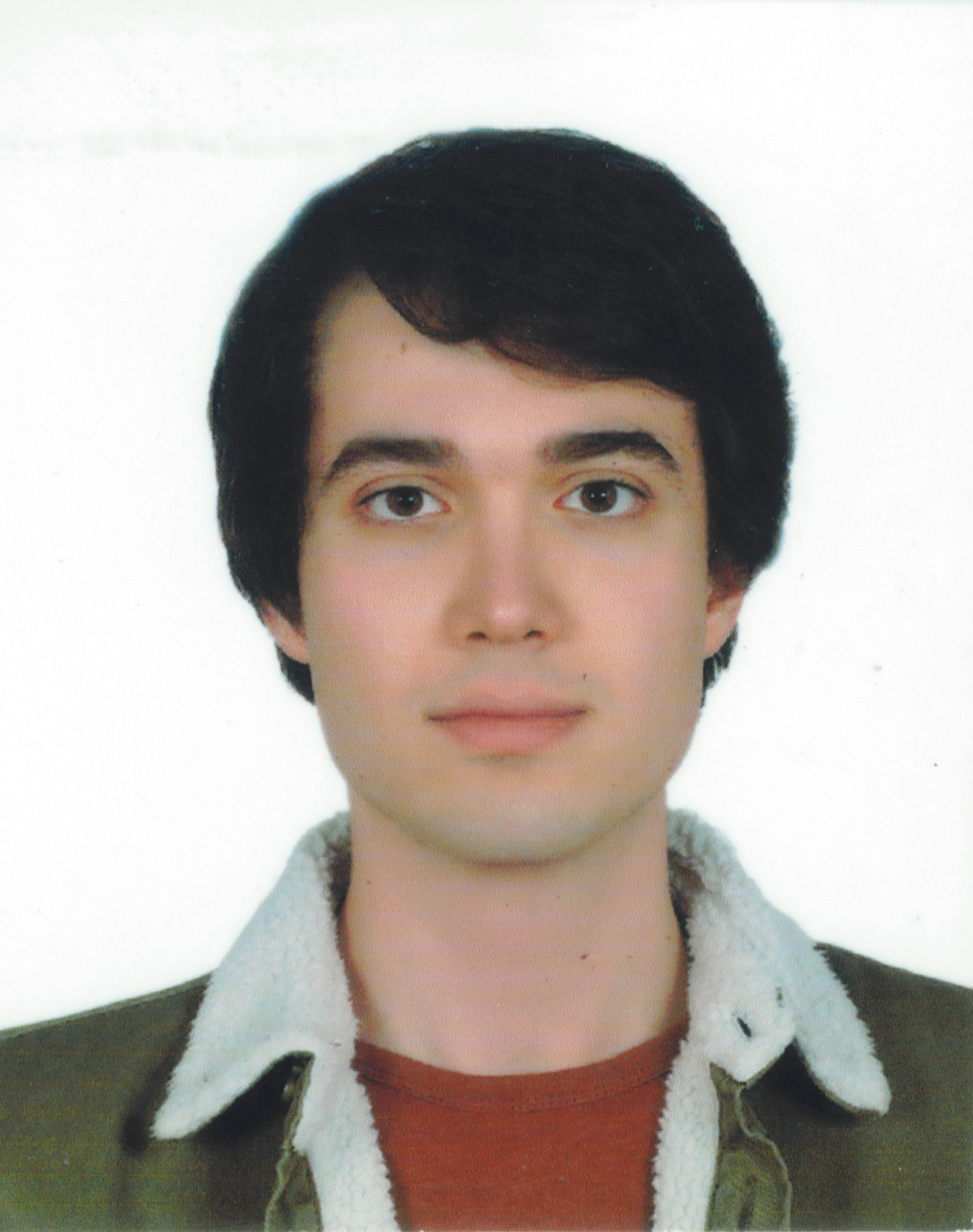}}]{Baris Can Cam}
received his B.Sc. degree in Electrical and Electronics Engineering from Eskisehir Osmangazi University, Turkey in 2016. He is currently pursuing his M.Sc. in Middle East Technical University, Ankara, Turkey. His research interests include computer vision with a focus on object detection.
\end{IEEEbiography}
\vspace{-1cm}

\begin{IEEEbiography}[{\includegraphics[width=1in,height=1.25in,clip,keepaspectratio]{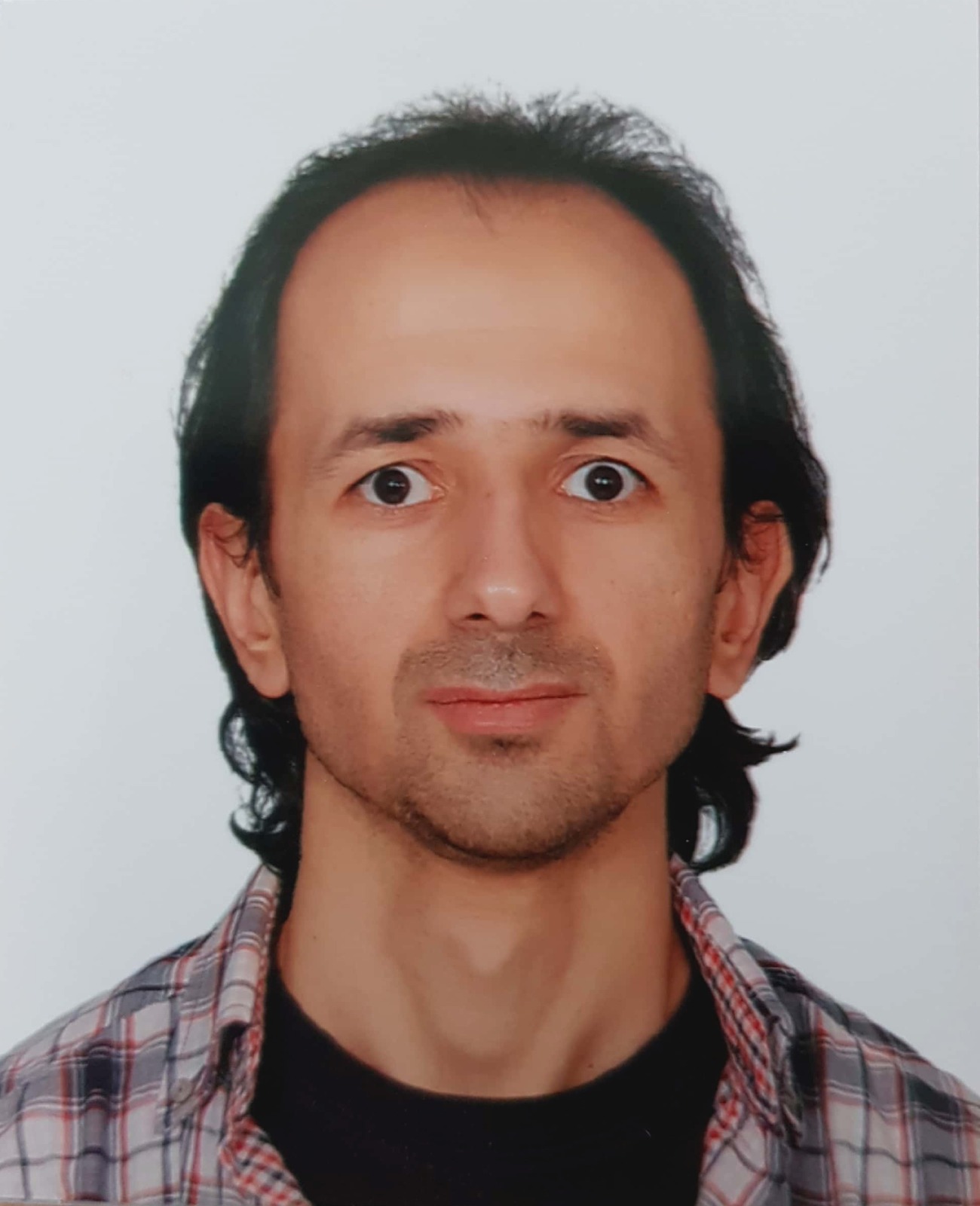}}]{Sinan Kalkan}
received his M.Sc. degree in Computer Engineering from Middle East Technical University (METU), Turkey in 2003, and his Ph.D. degree in Informatics from the University of Göttingen, Germany in 2008. After working as a postdoctoral researcher at the University of Gottingen and at METU, he became an assistant professor at METU in 2010. Since 2016, he has been working as an associate professor on problems within Computer Vision, and Developmental Robotics.
\end{IEEEbiography}
\vspace{-1cm}
\begin{IEEEbiography}[{\includegraphics[width=1in,height=1.25in,clip,keepaspectratio]{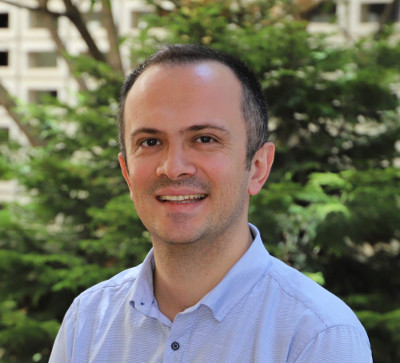}}]{Emre Akbas} is an assistant professor at the Department of Computer Engineering, Middle
    East Technical University (METU), Ankara, Turkey.
    He received his Ph.D. degree from the Department of Electrical and Computer
    Engineering, University of Illinois at Urbana-Champaign in 2011. His M.Sc. and
    B.Sc. degrees in computer science are both from METU.
    Prior to joining METU, he
    was a postdoctoral research associate at the Department of Pyschological and Brain
    Sciences, University of California Santa Barbara.
    His research interests
    are in computer vision and deep learning with a focus on object detection and human pose estimation. 
\end{IEEEbiography}




\end{document}